\documentclass[10pt,journal,compsoc]{IEEEtran}

\usepackage{times}
\usepackage{makecell}
\usepackage{multirow}
\usepackage{epsfig}
\usepackage{graphicx}
\usepackage{amsmath}
\usepackage{amssymb}
\usepackage{amsthm}

\ifCLASSOPTIONcompsoc
  \usepackage[nocompress]{cite}
\else
  \usepackage{cite}
\fi

\ifCLASSINFOpdf

\else

\fi

\usepackage{times}
\usepackage{epsfig}
\usepackage{graphicx}
\usepackage{amsmath}
\usepackage{amssymb}
\usepackage{amsthm}
\usepackage{bm}
\usepackage{subfigure}
\usepackage{epstopdf}
\usepackage{amsmath}
\usepackage{fancybox}
\usepackage{tikz}
\usepackage{environ}
\usepackage[linesnumbered,ruled,vlined]{algorithm2e}
\usepackage{cite}
\usepackage{threeparttable}
\usepackage{graphicx}
\usepackage[colorlinks,linkcolor=red,anchorcolor=blue,citecolor=green]{hyperref}

\newcommand{\eat}[1]{}
\newcommand{\etal}{{\em et al.~}} 
\newcommand{\ie}{{\em i.e.,~}} 

\begin{document}

\title{Hashing Learning with Hyper-Class Representation}

\author{Shichao~Zhang,~\IEEEmembership{Senior~Member,~IEEE, Jiaye Li}
\IEEEcompsocitemizethanks{\IEEEcompsocthanksitem School of Computer Science and Engineering,
	Central South University, Changsha 410083, PR China. \protect\\

E-mail: lijiaye@csu.edu.cn and zhangsc@csu.edu.cn

Corresponding author: ****.
}
\thanks{Manuscript received August **, 2022; revised ** **, 2022.}}

\markboth{}%
{Shell \MakeLowercase{\textit{et al.}}: Bare Demo of IEEEtran.cls for Computer Society Journals}

\IEEEtitleabstractindextext{%
\begin{abstract}
Existing unsupervised hash learning is a kind of attribute-centered calculation. It may not accurately preserve the similarity between data. This leads to low down the performance of hash function learning. In this paper, a hash algorithm is proposed with a hyper-class representation. It is a two-steps approach. The first step finds potential decision features and establish hyper-class. The second step constructs hash learning based on the hyper-class information in the first step, so that the hash codes of the data within the hyper-class are as similar as possible, as well as the hash codes of the data between the hyper-classes are as different as possible. To evaluate the efficiency, a series of experiments are conducted on four public datasets. The experimental results show that the proposed hash algorithm is more efficient than the compared algorithms, in terms of mean average precision (MAP), average precision (AP) and Hamming radius 2 (HAM2).
\end{abstract}

}

\maketitle

\IEEEdisplaynontitleabstractindextext

\IEEEpeerreviewmaketitle

\IEEEraisesectionheading{\section{Introduction}\label{sec:introduction}}

\IEEEPARstart{O}{n} the road to artificial intelligence, a major open problem is learning representation of data, which can be more efficiently applied to various data mining algorithms.
 A good data representation can make the follow-up learning task simple and efficient\cite{ray2022teaching}\cite{al2022fuzzy}. Traditional data representation only retains the information of data value. In order for AI to understand our world, it must be able to distinguish and separate the potential information hidden under the observed data. Therefore, how to construct an appropriate and efficient data representation is very important.

\begin{figure}[!ht]
	\begin{center}
		\vspace{-1mm}
		\subfigure{\scalebox{0.5}{\includegraphics{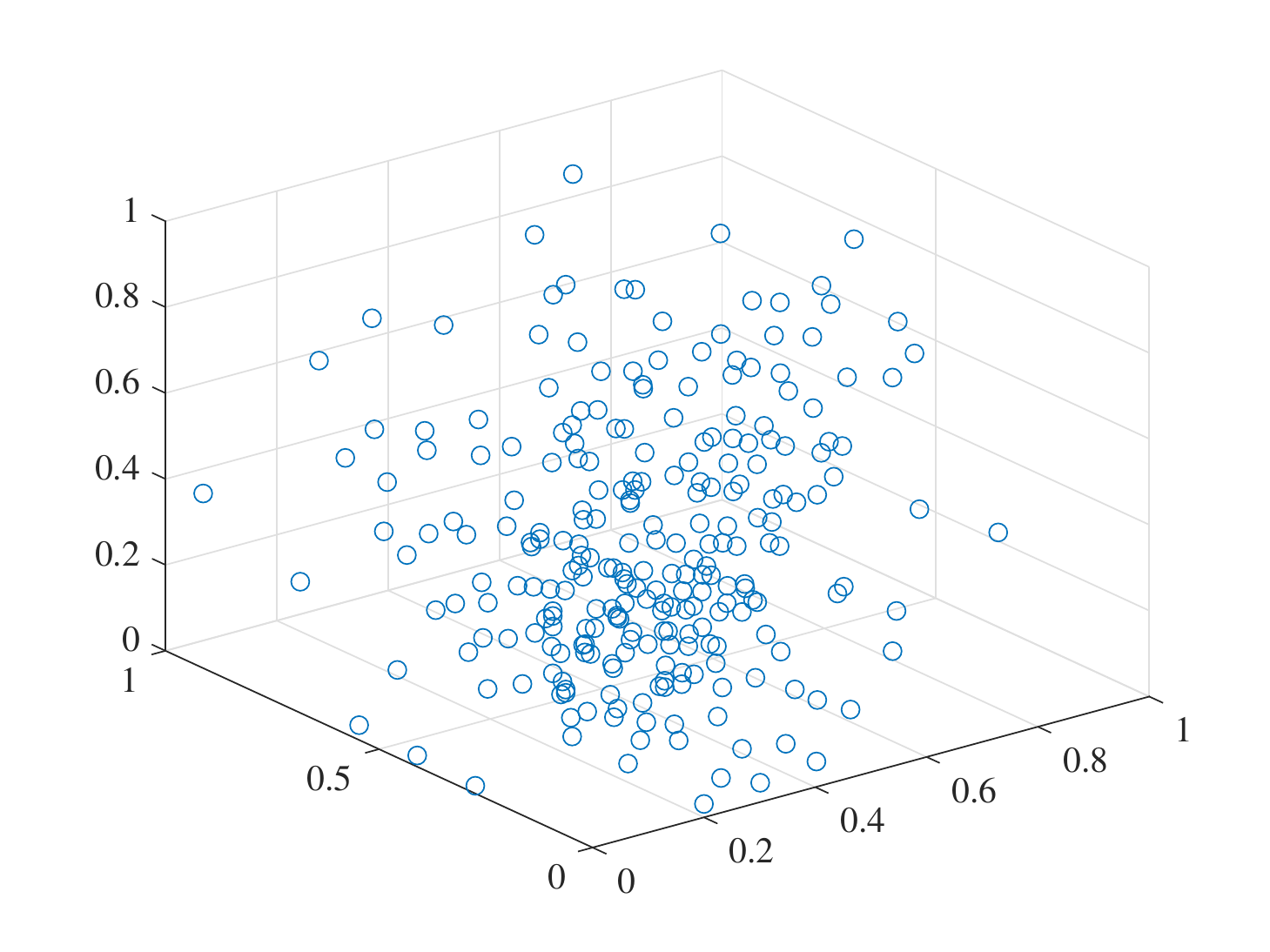}}}							
		\caption{An unlabeled dataset.}
		\label{fig1}
	\end{center}
\end{figure}

\begin{figure}[!ht]
	\begin{center}
		\vspace{-1mm}
		\subfigure{\scalebox{0.5}{\includegraphics{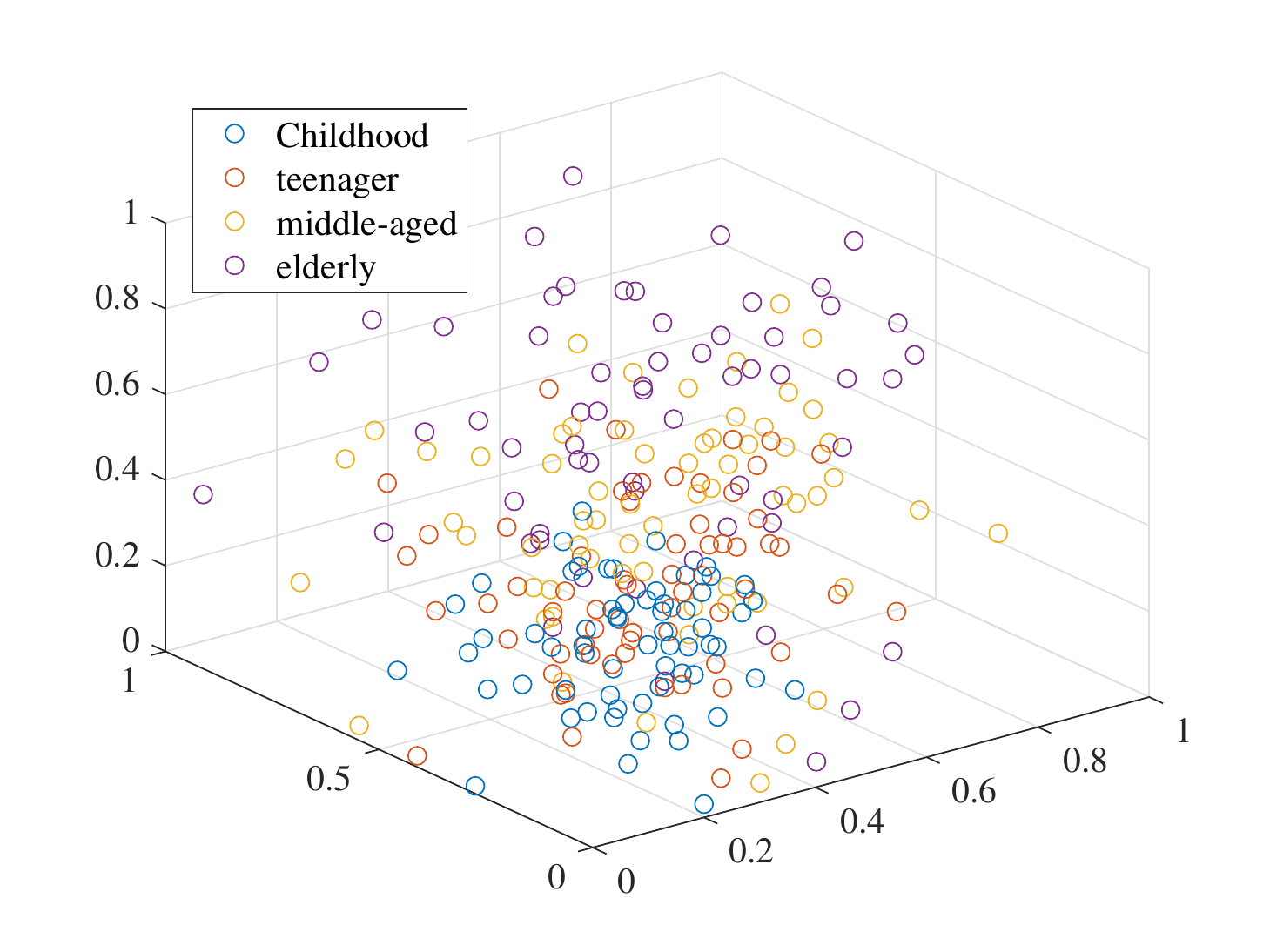}}}							
		\caption{Dataset under hyper-class representation}
		\label{fig2}
	\end{center}
\end{figure}

Most previous data representations convert text and image data into tables or matrices, which store discrete or continuous values of features\cite{xiao2022survey}\cite{nebli2022quantifying}. This often loses some potential information in the data. For example, in Fig. \ref{fig1}, we show an unlabeled data set (it has only three-dimensional features). The x-axis, y-axis and z-axis represent its three features, namely, height, weight and age. Obviously, according to their age, they can be divided into four classes, namely, childhood, teenager, middle-aged and elderly. This part of information is lost in Fig. \ref{fig1}. In this paper, we call this part of information hyper-class. As shown in Fig. \ref{fig2}, the information of hyper-classes 1-4 is added to the data. It is different from clustering. Clustering generates the class information of the data by calculating the similarity between all the features of the data. Hyper-class is the similarity calculation for a certain feature of all data. The hyper-class representation of data can add some potential information to the data, \ie hyper-class. In addition, it also simplifies the representation of data to a certain extent. For example, the original value of age may be a continuous value of 1-100, but now there are only four values: childhood, teenager, middle-aged and elderly.

Hash learning is an effective measure to solve the approximate nearest neighbor retrieval of large-scale data. Its core idea is to keep the similarity between data as short as possible. In unsupervised hash algorithm, the learned hash code can not perform approximate nearest neighbor retrieval well due to the lack of data labels. In addition, the missing label information can not guide the generation of hash code, which will lead to the high similarity of hash code in data of different classes, thus reducing the retrieval accuracy.

To solve the above problems, this paper proposes a novel hash algorithm under hyper-class representation. Specifically, we first find the potential decision features from the unlabeled data and establish the hyper-class representation of the data (the search decision features and feature selection algorithms are different. Feature selection is to find the feature subset that can best represent the overall information of the whole data, while we are looking for the feature that can best be used as the decision feature of the data). Then we construct a hash method according to the hyper-class representation, which can make the hash codes of the data within the hyper-class as similar as possible and the hash codes of the data between the hyper-classes as different as possible. On this basis, we can learn a more suitable hash code for approximate nearest neighbor retrieval.

The main contributions of this paper are as follows:
\begin{itemize}
	\item  We improve the hyper-class representation of data and apply it to hash learning. It can represent the potential hyper-class information in the data and simplify the representation of the data.
	\item  Based on the hyper-class representation, an effective hash algorithm is proposed, which can learn the more appropriate hash code for approximate nearest neighbor retrieval. At the same time, an alternating iterative algorithm is proposed to optimize it.
		
\end{itemize}

In addition, we conducted a series of experiments to show that the proposed algorithm exceeds the state-of-the-art methods in terms of MAP, AP and HAM2.

The rest of this paper is organized as follows. Section \ref{Preliminary} briefly reviews previous related work. Section \ref{Method} describes our proposed method in detail, the optimization process and time complexity analysis. Section \ref{experiments} shows the results of all algorithms on real datasets. Section \ref{conclusion} summarizes the full paper.

\section{Related work}\label{Preliminary}

In this section, we briefly introduce some data representation methods and hash methods.

\subsection{Data representation}

Data representation is the basis of data mining algorithm\cite{xiao2022survey}\cite{jiang2021novel}. An appropriate data representation can make the performance of data mining algorithm get twice the result with half the effort \cite{zhang2022hyper}. The traditional data representation methods include list method\cite{choi2021relative}, drawing method\cite{you2022novel} and equation method\cite{rasool2022introduction}. List method is the most commonly used data representation method. It puts the values of data into a table. Usually, each row of the table represents a sample and each column represents a feature. It only saves the value information of the data. Drawing method is to represent the data in the form of image. This kind of representation is generally not directly applicable to the data mining task of computer\cite{zhu2020unsupervised}. It also needs to further represent the data in a form that is easy to be processed by computer\cite{zhu2019efficient}. Equation method is to express the data with function according to some linear or nonlinear relationship in the data\cite{tang2018robust}. This method is rarely used because it is difficult to represent most data with only one equation. 

In addition to the traditional data representation methods, with the rapid development of artificial intelligence, many data representation methods have been proposed. For example, entity relationship representation \cite{al2021conceptual}. It uses entities to represent each sample or event, each entity has its corresponding features, and the relationship between entities is represented by edges. It is widely used in the field of natural language processing. On this basis, it is often used together with one-hot encoding\cite{yu2022missing}. For example, one-hot encoding is used to represent some concepts in medical data. It represents medical concepts as a binary vector, and the number of elements in the vector is the number of classes of medical concepts. The position corresponding to medical concepts in this vector is 1, otherwise it is 0. Although one-hot encoding is simple, it still has some shortcomings. For example, it will increase the dimension of data. High dimensional data will bring trouble to the algorithm\cite{zhu2018low}\cite{zhu2018one}. At this time, the sparse representation of data just solves this problem. Sparse representation is usually carried out by using sparse constraints, which can be expressed as the following convex optimization problem:
\begin{eqnarray}
\label{eq1}
\begin{array}{l}
\mathop {\min }\limits_{\mathbf W,\mathbf E} {\left\| \mathbf W \right\|_1} + \alpha {\left\| \mathbf E \right\|_{2,1}}\\
s.t.,\mathbf X = \mathbf X\mathbf W,diag(\mathbf W) = 0
\end{array}
\end{eqnarray}
where ${\left\| \mathbf W \right\|_1}$ is $l_1-$norm of $\mathbf W$ matrix and ${\left\| \mathbf E \right\|_{2,1}}$ is $l_{2,1}-$norm of noise matrix $\mathbf E$. $\mathbf X$ is the data matrix. In order to avoid obtaining the trivial solution $\mathbf W = \mathbf I$, the restriction of $diag(\mathbf W) = 0$ is added. Sparse representation can make the values in the data more sparse without affecting the information contained in the data itself. Based on the sparse representation of data, it is further proposed to use it for feature selection of data. For example, $l_1-$norm restriction is applied to the feature weight vector, which can remove the redundant features, so as to retain only the features that can better represent the data information, and further obtain the low dimensional representation of the data. Or we can obtain the low dimensional representation of the data by limiting the $l_{2,1}-$norm of the feature relation matrix. Of course, some other subspace learning algorithms also belong to the low-dimensional representation of data, such as principal component analysis (PCA)\cite{hasan2021review}, linear discriminant analysis (LDA)\cite{guo2021reverse} and local preserving projects (LPP)\cite{qiang2021robust}. 

In recent years, with the advent of alphago, artificial intelligence has ushered in a new upsurge\cite{9566737}. Many data representation methods based on deep learning have been proposed. Zhou \etal used deep learning and sparse representation to diagnose weak fault of bearing \cite{zhou2022hybrid}. Liu \etal proposed a data representation method based on knowledge map \cite{liu2022knowledge}. Specifically, it first constructs a multi-layer knowledge map for industrial Internet of things data. Then it uses the cognitive driven knowledge map to realize the automatic data fusion. Finally, it embeds the graph representation into the knowledge map and makes reasoning. Ahmadian \etal proposed a reliable depth representation method \cite{ahmadian2022reliable}. Specifically, it first proposes a probability model and reliability measure to score users implicitly. Then these implicit scores are input into deep sparse coding to generate a new representation of user characteristics. Finally, it proposes a new similarity measure function to calculate the similarity between users and recommend projects to target users. Prusa \etal proposed a new data representation method for deep neural network and text classification \cite{prusa2016designing}. It can reduce the dimension of data, \ie from \emph m characters to ${\log _2}(m)$ characters. Based on this, it can greatly reduce the memory consumption of the computer and reduce the amount of calculation. In addition, this method can reduce the memory use of the computer by 16 times, so as to reduce the number of neurons in deep learning. Bethge \etal proposed a representation learning method based on private encoder \cite{bethge2022domain}. Specifically, it first uses a neural network architecture for private coding from the far end of the data. Then the learned feature representation is shared to the global classifier. Finally, it uses domain alignment to learn the data representation independent of the data source. Jokanovic \etal studied the impact of data representation on deep learning \cite{jokanovic2016effect}. It mainly aims at the impact of falls related injuries on the elderly. Specifically, it studies the influence of different time-frequency representations on the performance of deep learning detector. Finally, they found that the appropriate time-frequency representation method is very important to the performance of the depth detector. Yi \etal studied the method, trend and application of graph representation \cite{yi2022graph}. They analyzed the application of graph representation in biological information research in detail. Rebuffi \etal proposed incremental data representation, \ie it is not a fixed data representation, but a representation that changes continuously with the increase of data \cite{rebuffi2017icarl}. Prabhakar and Lee proposed an improved sparse representation method \cite{prabhakar2022improved}. Specifically, it first preprocesses the data, and then thins the preprocessed data. Finally, it develops six different sparse representation optimization combinations. Wang \etal proposed a data representation method and applied it to network intrusion detection system \cite{wang2022representation}. Specifically, it first learns the explicit and implicit representation of data from the two spaces of samples and features, so as to establish the model of network behavior. Then, it establishes an unsupervised eigenvalue representation module to learn the relationship between features. Finally, it establishes a supervised neural network module to represent the object and learn the potential implicit relationship in the data.

\subsection{Hashing learning}

Hash learning is one of the effective measures for approximate nearest neighbor retrieval of large-scale high-dimensional data \cite{zhu2017graph}. Hash learning can be divided into data independent hash and data dependent hash according to whether there is a training process\cite{shen2022learning}. According to the label information, the data dependent hash can be divided into supervised hash and unsupervised hash\cite{wang2017survey}. Next, we will introduce them.

\textbf{Data independent hash}: Data independent hash has no training process. Its core is to preset the hash function and learn the distribution information in the data. The traditional data independent hash includes random hash, local sensitive hash and structure projection hash \cite{hayashi2016more} \cite{liu2015structure}. The core idea of random hash is first to reduce the dimension of data, and then randomly select the hash function in a specific function set. The prediction time complexity of random hash can be constant. Local sensitive hash (LSH) makes the hash codes of two similar points in the original space similar by maintaining the local structure of the data \cite{andoni2017optimal}. Although it ensures the structure information of data, its efficiency is relatively low and requires a long hash code. The core of structure projection hash is to divide or map the data space through some data structure (such as tree) or some projection (such as Hilbert curve).

Data independent hashes do not need training time\cite{kafai2014discrete}. They are widely used in the fields of approximate duplicate image detection and large-scale retrieval. However, they often need a long hash code to ensure the performance of the algorithm, which will lead to the memory burden of the computer and reduce the scope of their application.

\textbf{Data dependent hash}: The data dependent hash function obtains the hash function according to the training data\cite{jose2022deep}. Its core idea is to learn hash function by mining the information of linear relationship, nonlinear relationship, local structure and global structure in data. Unlike the data independent hash, the hash codes it learns are relatively compact, which can reduce the internal consumption of computer storage. Because of its high efficiency, data dependent hash is widely used in approximate nearest neighbor retrieval of large-scale image data\cite{yang2021deep}. 

According to whether the data has labels, the data dependent hash can be divided into supervised hash and unsupervised hash. In supervised hash learning, it usually uses the relationship between label information and data to establish hash function. Lin \etal proposed a supervised hash algorithm (\ie FastH)\cite{lin2014fast}. Specifically, it uses the block search method of graph cutting to solve large-scale reasoning, and uses the training enhanced decision tree to learn the nonlinear relationship in the data and construct the hash function. In unsupervised hash learning, it usually does not use data label information. For example, the iterative quantization method (ITQ)\cite{gong2012iterative} proposed by Gong \etal It uses the rotation problem of zero center data to solve the binary code, and uses the alternating minimization algorithm to calculate the quantization error of data mapping to the vertex of zero center binary hypercube. In addition, according to the form of data, data dependent hash can be divided into single-mode hash and cross mode hash. Monomodal hash only works on one form of data, such as images or text, and only constructs hash codes in monomodal data to approximate neighbor search. For example, k-nearest neighbors hashing (KNNH)\cite{he2019k} and concatenation hashing (CH)\cite{weng2020concatenation}. Transmembrane hash refers to establishing the association between multiple modes, so as to learn their hash code\cite{li2021adaptive}. It uses the hash code of one mode to retrieve the data of another mode, such as the hash code of text to retrieve the image or the hash code of image to retrieve the text.

\section{Our Approach}\label{Method}

In this section, we will introduce the constructed hyper-class representation, the proposed hash algorithm and the optimization process of the algorithm in detail.

\subsection{Notation}

In this paper, we use uppercase bold letters to represent matrices and lowercase bold letters to represent vectors. Given a matrix $\mathbf X = [{x_{ij}}]$, its \emph i-th row and \emph j-th column are represented as $\mathbf X_i$ and $\mathbf X^j$ respectively. Frobenius norm, $l_{2,1}-$norm and $l_1-$norm of matrix $\mathbf X$ are ${\left\| \mathbf X \right\|_F} = {(\sum\nolimits_i {\sum\nolimits_j {x_{ij}^2} } )^{{1 \mathord{\left/
				{\vphantom {1 2}} \right.
				\kern-\nulldelimiterspace} 2}}}$, ${\left\| \mathbf X \right\|_{2,1}} = \sum\nolimits_i {{{(\sum\nolimits_j {x_{ij}^2} )}^{{1 \mathord{\left/
					{\vphantom {1 2}} \right.
					\kern-\nulldelimiterspace} 2}}}} $ and ${\left\| \mathbf X \right\|_1} = \sum\nolimits_i {\sum\nolimits_j {\left| {{x_{ij}}} \right|} } $ respectively. The transpose, inverse and trace of matrix $\mathbf X$ are expressed as ${\mathbf X^T}$, ${\mathbf X^{ - 1}}$ and $tr(\mathbf X)$, respectively.

We  summarize these notations used in our paper in Table \ref{tab1}.
\begin{table}[!tb]
	\centering
	\caption{\footnotesize The detail of the notations used in this paper.}
	\centering
	{\footnotesize
		\begin{tabular}[c]{|c|c|} \hline
			$\mathbf X$             & training data 	  \\ \hline
			$\mathbf y$             & test data  \\ 	\hline
			$\mathbf X_i$           & the \emph i-th row of $\mathbf X$ \\ \hline
			$\mathbf X^j$           & the \emph j-th column of $\mathbf X$ \\ \hline	
			$\mathbf X_{^\neg i}$           & All remaining rows after \emph i-th row is removed from $\mathbf X$ \\ \hline								
			${\left\| \mathbf X \right\|_F}$	  & the frobenius norm of $\mathbf{X}$, \ie $||\mathbf{X}|{|_F} =  \sqrt {\sum\nolimits_{i,j} \mathbf{x}_{i,j} ^2}$ \\ \hline
			${\left\|\mathbf x \right\|_1}$    &the $l_1$ -norm of $\mathbf X$, \ie ${\left\|\mathbf x \right\|_1} = \sum\nolimits_{i = 1}^n {\left| {{x_i}} \right|}$   \\  \hline
			${\left\| \mathbf X \right\|_{2,1}}$    &$l_{2,1}-$norm of matrix \ie ${\left\| \mathbf X \right\|_{2,1}}  = \sum\nolimits_{i = 1}^n {{{\left\| {{\mathbf x_i}} \right\|}_2}} $   \\  \hline
			$\mathbf X^T$            & the transpose of $\mathbf X$  \\  \hline
			$\mathbf X^{-1}$            & the inverse of $\mathbf X$ \\ \hline		
			$tr(\mathbf X)$         & the trace of $\mathbf X$  \\ \hline	
			$\mathbf H$         & coding matrix  \\ \hline
            $\mathbf E$         & identity matrix  \\ \hline		
		\end{tabular}}
		\label{tab1}
	\end{table}

\subsection{Find decision features and establish hyper-class}

Hyper-class representation is a form of data representation. In this paper, the core process of establishing hyper-class representation is as follows: 1. Find potential decision feature. 2. Use the found decision features to classify the data into hyper-classes, so as to generate the hyper-class information of the data. The first step is very important, which is different from the traditional unsupervised feature selection algorithm. Unsupervised feature selection is to select the feature subset that can best represent the overall information of the data. The first step here is to find the most likely decision features as class labels. Next, we first carry out the first step. Specifically, we use each feature to fit all other features, and use the least square loss to get the relationship between each feature and other features. In order to maintain the nonlinear relationship in the data, we use the kernel function to map the samples. In addition, in order to obtain the closed form solution of the method, we introduce the $l_{2,1}-$norm limit of the relationship matrix, as shown in the following formula:
\begin{eqnarray}
\label{eq2}
\begin{array}{l}
\mathop {\min }\limits_\mathbf W \sum\limits_{i = 1}^d {\left\| {{\mathbf X_i} - k({\mathbf X_{{}^\neg i}})\mathbf W} \right\|_F^2}  + \varsigma {\left\| \mathbf W \right\|_{2,1}}
\end{array}
\end{eqnarray}
where ${\mathbf X_i} \in {\mathbb R^{1 \times n}}$ represents the \emph i-th feature and ${\mathbf X_{{}^\neg i}} \in {\mathbb R^{(d - 1) \times n}}$ represents all features except the \emph i-th feature. $\mathbf W \in {\mathbb R^{n \times n}}$ is the relation matrix. $k({\mathbf X_{{}^\neg i}}) \in {\mathbb R^{1 \times n}}$ is the vector after ${\mathbf X_{{}^\neg i}}$ is mapped to the kernel space. The specific mapping process is as follows:
\begin{eqnarray}
\label{eq3}
\begin{array}{l}
k({\mathbf X_{{}^\neg i}}) = [\varphi ({({\mathbf X_{{}^\neg i}})^{{1^T}}}{({\mathbf X_{{}^\neg i}})^1}),\varphi ({({\mathbf X_{{}^\neg i}})^{{2^T}}}{({\mathbf X_{{}^\neg i}})^2})\\
, \ldots ,\varphi ({({\mathbf X_{{}^\neg i}})^{{n^T}}}{({\mathbf X_{{}^\neg i}})^n})]
\end{array}
\end{eqnarray}

In addition, in order to obtain the possibility of each feature as a decision feature, we apply a weight $v_i$ to each feature (there are \emph d features in total). The greater the weight, the more it can be used as a decision feature. As follows:
\begin{eqnarray}
\label{eq4}
\begin{array}{l}
\mathop {\min }\limits_{\mathbf v, \mathbf W} \sum\limits_{i = 1}^d {{v_i}\left\| {{\mathbf X_i} - k({\mathbf X_{{}^\neg i}})\mathbf W} \right\|_F^2}  + \varsigma {\left\| \mathbf W \right\|_{2,1}}\\ - \sigma (\frac{1}{2}\left\| \mathbf v \right\|_2^2 - {\left\| \mathbf v \right\|_1})
\end{array}
\end{eqnarray}
where $\mathbf v \in {\mathbb R^{d \times 1}}$ represents the weight that each feature can be used as a decision feature. $\sigma $ and $\varsigma $ are adjustable parameters. $\frac{1}{2}\left\| \mathbf v \right\|_2^2 - {\left\| \mathbf v \right\|_1}$ is the ``soft" weight regular term, because it can obtain the actual value of the weight of each feature. Next, we solve Eq. (\ref{eq4}) by alternating iteration.

(1) Fix $\mathbf v$ and solve $\mathbf W$.

When $\mathbf v$ is fixed, Eq. (\ref{eq4}) can be written as follows:
\begin{eqnarray}
\label{eq5}
\begin{array}{l}
\mathop {\min }\limits_\mathbf W \sum\limits_{i = 1}^d {{v_i}\left\| {{\mathbf X_i} - k({\mathbf X_{{}^\neg i}})\mathbf W} \right\|_F^2}  + \varsigma {\left\| \mathbf W \right\|_{2,1}}
\end{array}
\end{eqnarray}

To facilitate the solution, Eq. (\ref{eq5}) can be further written as:
\begin{eqnarray}
\label{eq6}
\begin{array}{l}
\mathop {\min }\limits_\mathbf W \left\| {\mathbf Q - \mathbf G\mathbf W} \right\|_F^2 + \varsigma {\left\| \mathbf W \right\|_{2,1}}
\end{array}
\end{eqnarray}
where $\mathbf Q = \mathbf V\mathbf X$, $\mathbf G = \mathbf V \tilde {\mathbf X}$, $\mathbf V = diag(\sqrt \mathbf v )$ and $\tilde {\mathbf X} = [k({\mathbf X_{{}^\neg 1}});k({\mathbf X_{{}^\neg 2}}); \ldots ;k({\mathbf X_{{}^\neg d}})]$. Further, we can get that Eq. (\ref{eq6}) is equivalent to the following equation:
\begin{eqnarray}
\label{eq7}
\begin{array}{l}
\mathop {\min }\limits_\mathbf W \left\| {\mathbf Q - \mathbf G\mathbf W} \right\|_F^2 + \varsigma tr({\mathbf W^T}\mathbf F\mathbf W)
\end{array}
\end{eqnarray}

We use Eq. (\ref{eq7}) to derive $\mathbf W$, and we can get:
\begin{eqnarray}
\label{eq8}
\begin{array}{l}
 - {\mathbf Q^T}\mathbf G + {\mathbf G^T}\mathbf G\mathbf W + \varsigma \mathbf F\mathbf W
\end{array}
\end{eqnarray}
where
\begin{eqnarray}
\label{eq9}
\begin{array}{l}
{F_{ii}} = \frac{1}{{2\sqrt {\left\| {{\mathbf W^i}} \right\|_2^2 + \varepsilon } }}(\varepsilon  \to 0,i = 1,2, \ldots ,n)
\end{array}
\end{eqnarray}

If we make Eq. (\ref{eq8}) equal to zero, we can get the solution of $\mathbf W$ as follows:
\begin{eqnarray}
\label{eq10}
\begin{array}{l}
\mathbf W = {({\mathbf G^T}\mathbf G + \varsigma F)^{ - 1}}{\mathbf G^T}\mathbf Q
\end{array}
\end{eqnarray}

(2) Fix $\mathbf W$ and solve $\mathbf v$.

When $\mathbf W$ is fixed, Eq. (\ref{eq4}) can be written as follows:
\begin{eqnarray}
\label{eq11}
\begin{array}{l}
\mathop {\min }\limits_\mathbf v \sum\limits_{i = 1}^d {{v_i}\left\| {{\mathbf X_i} - k({\mathbf X_{{}^\neg i}})\mathbf W} \right\|_F^2}  - \sigma (\frac{1}{2}\left\| \mathbf v \right\|_2^2 - {\left\| \mathbf v \right\|_1})
\end{array}
\end{eqnarray}

We let $L({\mathbf X_i},{\mathbf X_{{}^\neg i}}, \mathbf W) = \left\| {{\mathbf X_i} - k({\mathbf X_{{}^\neg i}})\mathbf W} \right\|_F^2$, then Eq. (\ref{eq11}) can be further written as follows:
\begin{eqnarray}
\label{eq12}
\begin{array}{l}
\mathop {\min }\limits_\mathbf v \sum\limits_{j = 1}^d {{v_i}L({\mathbf X_i},{\mathbf X_{{}^\neg i}},\mathbf W)}  - \frac{1}{2}\sigma \sum\limits_{i = 1}^d {(v_i^2 - 2{v_i})} 
\end{array}
\end{eqnarray}

We use Eq. (\ref{eq12}) to derive $v_i$ and make the derivative zero, and we can get the solution of $\mathbf v$ as follows:
\begin{eqnarray}
\label{eq13}
\begin{array}{l}
{v_i} = \left\{ {\begin{array}{*{20}{c}}
	{1 - \frac{{L({\mathbf X_i},{\mathbf X_{{}^\neg i}},\mathbf W)}}{\sigma },}&{L({\mathbf X_i},{\mathbf X_{{}^\neg i}},\mathbf W) < \sigma }\\
	{0,}&{L({\mathbf X_i},{\mathbf X_{{}^\neg i}},\mathbf W) \ge \sigma }
	\end{array}} \right.
\end{array}
\end{eqnarray}

After the optimal solution of $\mathbf v$ is obtained, the weight of each feature as a decision feature is obtained. We can obtain the most likely decision feature by the following formula:
\begin{eqnarray}
\label{eq14}
\begin{array}{l}
df = \max \{ {v_1},{v_2}, \ldots ,{v_d}\} 
\end{array}
\end{eqnarray}

Through Eq. (\ref{eq14}), after obtaining the most likely decision feature $df$, we use the method in literature \cite{makarychev2022performance} to divide the data under decision feature $df$.

\subsection{Hash method based on hyper-class representation}

After the data is represented by hyper-class, its original data still exists. Hyper-class representation only adds some potential hyper-class information (\ie class $\{ 1,2,3, \ldots ,c\}$, assuming that the established hyper-class has \emph c classes) on the basis of the original data. When the hyper-class has only one class, \ie  all data are of the same class, and their class labels are the same ($c=1$). At this time, our idea is to make the similarity between the hash codes corresponding to each sample as high as possible, because they belong to the same class. 

When $c=1$, \ie there is only one class. At this time, the objective function is as follows:
\begin{eqnarray}
\label{eq15}
\begin{array}{l}
\mathop {\min }\limits_{\mathbf H,\mathbf U,\mathbf S} \sum\nolimits_{i,j}^n {\left\| {{\mathbf H^{(i)}}\mathbf U - {\mathbf H^{(j)}}\mathbf U} \right\|_2^2{s_{ij}}}  + \alpha \sum\nolimits_i^n {\left\| {{\mathbf s_i}} \right\|_2^2} \\
+ \beta \sum\nolimits_{i = 1}^n {\left\| {{\mathbf H^{(i)}} - \bm \mu } \right\|_2^2} \\
s.t.{\rm{ }}\mathbf s_i^T1 = 1,~{s_{ii}} = 0,\\
{s_{ij}} \ge 0, ~if~{\rm{ }}j \in N(i),~{\rm{ }}otherwise{\rm{ }}~0
\end{array}
\end{eqnarray}
where $\mathbf S \in {\mathbb R^{n \times n}}$ is the similarity matrix of hash codes, which records the similarity between the hash codes of each sample, where ${s_{ii}} = 0$, because the similarity between the hash code of each sample and itself is 0. ${\mathbf H^{(i)}}$ is the hash code of the \emph i-th sample, and ${\mathbf H^{(j)}}$ is the hash code of the \emph j-th sample. $U$ is the mapping matrix. $\bm \mu$ is the average of all sample hash codes. In Eq. (\ref{eq15}), $\sum\nolimits_{i,j}^n {\left\| {{\mathbf H^{(i)}}\mathbf U - {\mathbf H^{(j)}}\mathbf U} \right\|_2^2{s_{ij}}} $ calculates the similarity between the hash codes of all samples, and $\sum\nolimits_i^n {\left\| {{\mathbf s_i}} \right\|_2^2} $ is the regular term limit of variable $\mathbf s$. The main function of $\sum\nolimits_{i = 1}^n {\left\| {{\mathbf H^{(i)}} - \bm \mu } \right\|}$ is to reduce the deviation between all hash codes and the mean, so as to reduce the influence of outliers. Eq. (\ref{eq15}) only considers the relationship between hash codes and does not consider the relationship between hash codes and samples. Therefore, we further obtain the following formula:
\begin{eqnarray}
\label{eq16}
\begin{array}{l}
\mathop {\min }\limits_{\mathbf H,\mathbf U,\mathbf S} \sum\nolimits_{i,j}^n {\left\| {{\mathbf H^{(i)}}\mathbf U - {\mathbf H^{(j)}}\mathbf U} \right\|_2^2{s_{ij}}}  + \alpha \sum\nolimits_i^n {\left\| {\mathbf s_i} \right\|_2^2} \\
+ \beta \sum\nolimits_{i = 1}^n {\left\| {{\mathbf H^{(i)}} - \bm \mu } \right\|_2^2}  + \lambda \left\| {\mathbf H - {\mathbf X^T}{\mathbf U^T}} \right\|_F^2 + \eta \left\| \mathbf U \right\|_2^2\\
s.t.~{\rm{ }}\mathbf s_i^T1 = 1,~{s_{ii}} = 0,\\
{s_{ij}} \ge 0,~if{\rm{ }}j \in N(i),~{\rm{ }}otherwise{\rm{ }}0,~{\rm{ }}\mathbf U{\mathbf U^T} = \mathbf I
\end{array}
\end{eqnarray}
where $\left\| {\mathbf H - {\mathbf X^T}{\mathbf U^T}} \right\|_F^2$ considers the relationship between the hash code and the sample, and $\left\| \mathbf U \right\|_2^2$ is the regular term containing $\mathbf U$, which can maintain the closed solution of the function. $\mathbf U{\mathbf U^T} = \mathbf I$ is the orthogonal limit of $\mathbf U$, which can make the hyperplane of hash function irrelevant (\ie orthogonal to each other). In practice, it is almost impossible for the hyper-class representation we established to have only one class (\ie in most cases, $c > 1$). We further construct the objective function in the following two cases:

When $c=2$, \ie there are only two hyper-class to be constructed. At this time, the objective function is as follows:
\begin{eqnarray}
\label{eq17}
\begin{array}{l}
\mathop {\min }\limits_{\mathbf H,\mathbf U,\mathbf S} \sum\nolimits_{i,j}^{{n_1}} {\left\| {\mathbf H_1^{(i)}\mathbf U - \mathbf H_1^{(j)}\mathbf U} \right\|_2^2{s_1^{(ij)}}}  + \alpha \sum\nolimits_i^{{n_1}} {\left\| {{s_1^{(i)}}} \right\|_2^2} \\
- \beta \sum\nolimits_{i = 1}^{{n_1}} {\left\| {\mathbf H_1^{(i)} - {\bm \mu _2}} \right\|_2^2}  + \sum\nolimits_{i,j}^{{n_2}} {\left\| {\mathbf H_2^{(i)}\mathbf U - \mathbf H_2^{(j)}\mathbf U} \right\|_2^2{s_2^{(ij)}}} \\
+ \alpha \sum\nolimits_i^{{n_2}} {\left\| {{\mathbf s_2^{(i)}}} \right\|_2^2}  - \beta \sum\nolimits_{i = 1}^{{n_2}} {\left\| {\mathbf H_2^{(i)} - {\bm \mu _1}} \right\|_2^2} \\
+ \lambda \left\| {\mathbf H - {\mathbf X^T}{\mathbf U^T}} \right\|_F^2 + \eta \left\| \mathbf U \right\|_2^2\\
s.t.~{\rm{ }}\mathbf s_1^T1 = \mathbf 1,~{s_1^{ii}} = 0, \mathbf s_2^T1 = \mathbf 1,~{s_2^{ii}} = 0,\\
{s_{ij}} \ge 0,~if{\rm{ }}j \in N(i),~{\rm{ }}otherwise{\rm{ }}~0,~{\rm{ }}\mathbf U{\mathbf U^T} = \mathbf I
\end{array}
\end{eqnarray}
where $n_1$ represents the number of samples in the first hyper-class and $n_2$ represents the number of samples in the second hyper-class. $\bm \mu_1$ is the center of the hash code of the first hyper-class data, and $\bm \mu_2$ is the center of the hash code of the second hyper-class data. In Eq. (\ref{eq17}), it should be noted that we use $\sum\nolimits_{i,j}^{{n_1}} {\left\| {\mathbf H_1^{(i)}\mathbf U - \mathbf H_1^{(j)}\mathbf U} \right\|_2^2{s_{ij}}} $ and $\sum\nolimits_{i,j}^{{n_2}} {\left\| {\mathbf H_2^{(i)}\mathbf U - \mathbf H_2^{(j)}\mathbf U} \right\|_2^2{s_{ij}}} $ to calculate the similarity of the internal hash codes of the first hyper-class and the second hyper-class respectively. $\sum\nolimits_{i = 1}^{{n_1}} {\left\| {\mathbf H_1^{(i)} - {\bm \mu _2}} \right\|}$ is used to calculate the similarity between the hash code of the internal data of the first hyper-class and the hash code of the second hyper-class. $\sum\nolimits_{i = 1}^{{n_2}} {\left\| {\mathbf H_2^{(i)} - {\bm \mu _1}} \right\|} $ is used to calculate the similarity between the hash code of the internal data of the second hyper-class and the hash code of the first hyper-class. Different from the case of $c=1$, we use a minus sign for the three terms of the objective function, because $\sum\nolimits_{i = 1}^{{n_1}} {\left\| {\mathbf H_1^{(i)} - {\bm \mu _2}} \right\|}$ and $\sum\nolimits_{i = 1}^{{n_2}} {\left\| {\mathbf H_2^{(i)} - {\bm \mu _1}} \right\|}$ represent the similarity between different hyper-classes hash codes, which can reduce the similarity between data hash codes between hyper-classes. \ie it can make the hash code similarity of the internal data of each hyper-class high and the hash code similarity between different hyper-class data low. Similarly, further, we can get the case of $c > 2$.

When $c > 2$, the objective function is as follows:
\begin{equation}
\label{eq18}
\begin{array}{l}
\mathop {\min }\limits_{\mathbf H,\mathbf U,\mathbf S} \sum\nolimits_{i,j}^{{n_1}} {\left\| {\mathbf H_1^{(i)}\mathbf U - \mathbf H_1^{(j)}\mathbf U} \right\|_2^2s_1^{(ij)}}  + \alpha \sum\nolimits_i^{{n_1}} {\left\| {\mathbf s_1^{(i)}} \right\|_2^2} \\
- \beta \sum\nolimits_{i = 1}^{{n_1}} {\left\| {{\mathbf I_{{}^\neg 1}}\mathbf H_1^{(i)} - {\mathbf V_{{}^\neg 1}}} \right\|_2^2}  + \sum\nolimits_{i,j}^{{n_2}} {\left\| {\mathbf H_2^{(i)}\mathbf U - \mathbf H_2^{(j)}\mathbf U} \right\|_2^2s_2^{(ij)}} \\
+ \alpha \sum\nolimits_i^{{n_2}} {\left\| {\mathbf s_2^{(i)}} \right\|_2^2}  - \beta \sum\nolimits_{i = 1}^{{n_2}} {\left\| {{\mathbf I_{{}^\neg 2}}\mathbf H_2^{(i)} - {\mathbf V_{{}^\neg 2}}} \right\|_2^2} \\
+  \cdots  + \sum\nolimits_{i,j}^{{n_c}} {\left\| {\mathbf H_c^{(i)}\mathbf U - \mathbf H_c^{(j)}\mathbf U} \right\|_2^2s_c^{(ij)}}  + \alpha \sum\nolimits_i^{{n_c}} {\left\| {\mathbf s_c^{(i)}} \right\|_2^2} \\
- \beta \sum\nolimits_{i = 1}^{{n_c}} {\left\| {{\mathbf I_{{}^\neg c}}\mathbf H_c^{(i)} - {\mathbf V_{{}^\neg c}}} \right\|_2^2}  + \lambda \left\| {\mathbf H - {\mathbf X^T}{\mathbf U^T}} \right\|_F^2 + \eta \left\| \mathbf U \right\|_2^2\\
s.t.{\rm{ }}\mathbf s_1^{{{(i)}^T}}1 = 1,\mathbf s_2^{{{(i)}^T}}1 = 1, \ldots ,\mathbf s_c^{{{(i)}^T}}1 = 1,\\
s_1^{(ii)} = 0,s_2^{(ii)} = 0, \ldots ,s_c^{(ii)} = 0\\
s_1^{(ij)},s_2^{(ij)}, \ldots s_c^{(ij)} \ge 0,if{\rm{ }}j \in N(i),{\rm{ }}otherwise{\rm{ }}~0,{\rm{ }}\mathbf U{\mathbf U^T} = \mathbf I
\end{array}
\end{equation}

In Eq. (\ref{eq18}), $n_c$ represents the number of samples in the \emph c-th hyper-class. ${\mathbf I_{{}^\neg 1}}{\mathbf I_{{}^\neg 2}} \ldots {\mathbf I_{{}^\neg c}}$ is all 1 vectors of $(c - 1) \times 1$. ${\mathbf V_{{}^\neg 1}} \in {\mathbb R^{(c - 1) \times l}}$ represents the mean matrix composed of the mean values of hash codes in each hyper-class except the first hyper-class. Similarly, ${\mathbf V_{{}^\neg 2}} \in {\mathbb R^{(c - 1) \times l}}$ represents the mean matrix composed of the mean values of hash codes in each hyper-class except the second hyper-class. ${\mathbf V_{{}^\neg c}} \in {\mathbb R^{(c - 1) \times l}}$ represents the mean matrix composed of the mean values of hash codes in each hyper-class except the \emph c hyper-class. ${\mathbf H_1} \in {\mathbb R^{{n_1} \times l}}$, ${\mathbf H_2} \in {\mathbb R^{{n_2} \times l}}$ and ${\mathbf H_c} \in {\mathbb R^{{n_c} \times l}}$ represent the hash code matrix of the samples in the first, second and \emph c hyper-class, respectively. Its idea is the same as in the case of $c=2$. Based on the hyper-class representation, the similarity of hash codes of data within hyper-class is as large as possible, and the similarity of hash codes between hyper-classes is as small as possible.

\subsection{Hash method optimization process}\label{Hash method optimization process}

In this section, we optimize the proposed objective function in two cases (\ie $c=1$ and $c>1$). We still use the optimization method of alternating iteration (\ie solving one variable and fixing other variables) to optimize them.

When $c=1$, the optimization process of the objective function (\ie Eq. (\ref{eq16}) is as follows:

When $\mathbf U$ and $\mathbf S$ are fixed to solve $\mathbf H$, Eq. (\ref{eq16}) can be written as follows:
\begin{eqnarray}
\label{eq19}
\begin{array}{l}
\mathop {\min }\limits_\mathbf H \sum\nolimits_{i,j}^n {\left\| {{\mathbf H^{(i)}}\mathbf U - {\mathbf H^{(j)}}\mathbf U} \right\|_2^2{s_{ij}}}  + \beta \sum\nolimits_{i = 1}^n {\left\| {{\mathbf H^{(i)}} - \bm \mu } \right\|} _2^2\\
+ \lambda \left\| {\mathbf H - {\mathbf X^T}{\mathbf U^T}} \right\|_F^2
\end{array}
\end{eqnarray}

Further, Eq. (\ref{eq19}) can be written as follows:
\begin{eqnarray}
\label{eq20}
\begin{array}{l}
\mathop {\min }\limits_\mathbf H \beta \sum\nolimits_{i = 1}^n {\left( {\left\langle {{\mathbf H^{(i)}},{\mathbf H^{(i)}}} \right\rangle  - 2\left\langle {{\mathbf H^{(i)}},\bm \mu } \right\rangle  + \left\langle {\bm \mu ,\bm \mu } \right\rangle } \right)} \\
+ \lambda \left\| {\mathbf H - {\mathbf X^T}{\mathbf U^T}} \right\|_F^2 + tr({\mathbf U^T}{\mathbf H^T}\mathbf L\mathbf H\mathbf U)
\end{array}
\end{eqnarray}

Because $tr({\mathbf U^T}{\mathbf H^T}\mathbf L\mathbf H\mathbf U) = tr(\mathbf U{\mathbf U^T}{\mathbf H^T}\mathbf L\mathbf H)$ and $\mathbf U{\mathbf U^T} = \mathbf I$. Therefore, the first term in Eq. (\ref{eq20}) can be written as $tr({\mathbf H^T}\mathbf L\mathbf H)$. We use Eq. (\ref{eq20}) to find the derivative of ${\mathbf H^{(i)}}$, and let the derivative is 0, so we can get the following formula:
\begin{eqnarray}
\label{eq21}
\begin{array}{l}
{\mathbf L^{(i)}}[{({\mathbf E^{(i)}})^T}{\mathbf H^{(i)}} + \sum\limits_{j \ne i}^{n - 1} {{{({\mathbf E^{(i)}})}^T}{\mathbf H^{(j)}}} ]\\
+ (\beta  + \lambda ){\mathbf H^{(i)}} - \lambda {({\mathbf X^T}{\mathbf U^T})^i} = \beta \frac{1}{n}\sum\limits_{i = 1}^n {{{({\mathbf X^T}{\mathbf U^T})}^i}} 
\end{array}
\end{eqnarray}
where $\mathbf E$ is the identity matrix, the solution of ${\mathbf H^{(i)}}$ can be further obtained as follows:
\begin{equation}
\label{eq22}
\begin{array}{l}
{\mathbf H^{(i)}} = {({\mathbf L^{(i)}}{({\mathbf E^{(i)}})^T} + \beta  + \lambda )^{ - 1}}\\
(\lambda {({\mathbf X^T}{\mathbf U^T})^i} - {\mathbf L^{(i)}}\sum\limits_{j \ne i}^{n - 1} {{{({\mathbf E^{(i)}})}^T}{\mathbf H^{(j)}}}  + \beta \frac{1}{n}\sum\limits_{i = 1}^n {{{({\mathbf X^T}{\mathbf U^T})}^i}} )
\end{array}
\end{equation}

When $\mathbf H$ and $\mathbf S$ are fixed to solve $\mathbf U$, Eq. (\ref{eq16}) can be written as follows:
\begin{eqnarray}
\label{eq23}
\begin{array}{l}
\mathop {\min }\limits_\mathbf U \sum\nolimits_{i,j}^n {\left\| {{\mathbf H^{(i)}}\mathbf U - {\mathbf H^{(j)}}\mathbf U} \right\|_2^2{s_{ij}}} \\
+ \lambda \left\| {\mathbf H - {\mathbf X^T}{\mathbf U^T}} \right\|_F^2 + \eta \left\| \mathbf U \right\|_2^2\\
s.t.~{\rm{ }}\mathbf U{\mathbf U^T} = \mathbf I
\end{array}
\end{eqnarray}

Further, Eq. (\ref{eq23}) can be rewritten as follows:
\begin{eqnarray}
\label{eq24}
\begin{array}{l}
\mathop {\min }\limits_\mathbf H tr({\mathbf U^T}{\mathbf H^T}\mathbf L\mathbf H\mathbf U)\\
+ \lambda (({\mathbf H^T} - \mathbf U\mathbf X)(\mathbf H - {\mathbf X^T}{\mathbf U^T})) + \eta \left\| \mathbf U \right\|_2^2\\
s.t.{\rm{ }}\mathbf U{\mathbf U^T} = \mathbf I
\end{array}
\end{eqnarray}

Let's take $f(\mathbf U) = tr({\mathbf U^T}{\mathbf H^T}\mathbf L\mathbf H\mathbf U) + \lambda (({\mathbf H^T} - \mathbf U\mathbf X)(\mathbf H - {\mathbf X^T}{\mathbf U^T})) + \eta \left\| \mathbf U \right\|_2^2$ and take the derivative of $\mathbf U$, and we can get:
\begin{eqnarray}
\label{eq25}
\begin{array}{l}
\frac{{\partial f(\mathbf U)}}{{\partial \mathbf U}} = 2{\mathbf H^T}\mathbf L\mathbf H\mathbf U\\
+ \lambda (2\mathbf U\mathbf X{\mathbf X^T} - 2{\mathbf H^T}{\mathbf X^T}) + 2\eta \mathbf I\mathbf U
\end{array}
\end{eqnarray}

Because Eq. (\ref{eq24}) has an orthogonal restriction on $\mathbf U$, \ie $\mathbf U{\mathbf U^T} = \mathbf I$, we use the method in literature \cite{wen2013feasible} to solve it.

When $\mathbf H$ and $\mathbf U$ are fixed to solve $\mathbf S$, Eq. (\ref{eq16}) can be written as follows:
\begin{eqnarray}
\label{eq26}
\begin{array}{l}
\mathop {\min }\limits_\mathbf U \sum\nolimits_{i,j}^n {\left\| {{\mathbf H^{(i)}}\mathbf U - {\mathbf H^{(j)}}\mathbf U} \right\|_2^2{s_{ij}}}  + \alpha \sum\nolimits_i^n {\left\| {{\mathbf s_i}} \right\|_2^2} \\
s.t.~{\rm{ }}\mathbf s_i^T\mathbf 1 = 1,~{s_{ii}} = 0,\\
{s_{ij}} \ge 0,~if{\rm{ }}j \in N(i),~{\rm{ }}otherwise{\rm{ }}0
\end{array}
\end{eqnarray}

Optimizing $\mathbf s$ is equivalent to optimizing each ${\mathbf s_i}(i = 1,...,n)$ separately, so we further convert the optimization problem into the following formula:
\begin{eqnarray}
\label{eq27}
\begin{array}{l}
\mathop {\min }\limits_{\mathbf s_i^T\mathbf 1 = 1,{s_{i,i}} = 0,{s_{i,j}} \ge 0} \sum\nolimits_{i,j}^n {(\left\| {{\mathbf H^{(i)}}\mathbf U - {\mathbf H^{(j)}}\mathbf U} \right\|_2^2{s_{i,j}}} \\ + \alpha s_{i,j}^2)
\end{array}
\end{eqnarray}

Here, let $\mathbf Z \in {\mathbb R^{n \times n}}$, where ${Z_{i,j}} = \left\| {{\mathbf H^{(i)}}\mathbf U - {\mathbf H^{(j)}}\mathbf U} \right\|_2^2$, so equation (26) further becomes:
\begin{eqnarray}
\label{eq28}
\begin{array}{l}
\mathop {\min }\limits_{\mathbf s_i^T\mathbf 1 = 1,~{s_{i,i}} = 0,~{s_{i,j}} \ge 0} \left\| {{\mathbf s_i} + \frac{1}{{2\alpha }}{\mathbf Z_i}} \right\|_2^2
\end{array}
\end{eqnarray}

Under KKT conditions, we can get the following results:
\begin{eqnarray}
\label{eq29}
\begin{array}{l}
{s_{i,j}} = {( - \frac{1}{{2\alpha }}{Z_{i,j}} + \tau )_ + }
\end{array}
\end{eqnarray}

Since each hash code has close neighbors, we arrange each ${\mathbf Z_i}(i = 1,...,n)$ in descending order, \ie ${\hat Z_i} = \{ {\hat Z_{i,1}},...,{\hat Z_{i,n}}\}$, then we can know that ${s_{i,k + 1}} = 0$ and ${s_{i,k}} > 0$. Available:
\begin{eqnarray}
\label{eq30}
\begin{array}{l}
 - \frac{1}{{2\alpha }}{\hat Z_{i,k + 1}} + \tau  \le 0
\end{array}
\end{eqnarray}

Under the condition ${\mathbf s_i}^T\mathbf 1 = 1$, we can get:
\begin{eqnarray}
\label{eq31}
\begin{array}{l}
\sum\nolimits_{j = 1}^k {(\frac{1}{{2\alpha }}{{\hat Z}_{i,k}} + \tau )}  = 1 \Rightarrow \tau  = \frac{1}{k} + \frac{1}{{2k\alpha }}\sum\nolimits_{j = 1}^k {{{\hat Z}_{i,k}}} 
\end{array}
\end{eqnarray}

When $c>1$, the optimization process of the objective function (\ie Eq. (\ref{eq18}) is as follows:

When $\mathbf U$ and $\mathbf S$ are fixed to solve $\mathbf H$, Eq. (\ref{eq18}) can be written as follows:
\begin{eqnarray}
\label{eq32}
\begin{array}{l}
\mathop {\min }\limits_\mathbf H \sum\nolimits_{i,j}^{{n_1}} {\left\| {\mathbf H_1^{(i)}\mathbf U - \mathbf H_1^{(j)}\mathbf U} \right\|_2^2s_1^{(ij)}} \\
- \beta \sum\nolimits_{i = 1}^{{n_1}} {\left\| {{\mathbf I_{{}^\neg 1}}\mathbf H_1^{(i)} - {\mathbf V_{{}^\neg 1}}} \right\|_2^2} \\
+ \sum\nolimits_{i,j}^{{n_2}} {\left\| {\mathbf H_2^{(i)}\mathbf U - \mathbf H_2^{(j)}\mathbf U} \right\|_2^2s_2^{(ij)}} \\
- \beta \sum\nolimits_{i = 1}^{{n_2}} {\left\| {{\mathbf I_{{}^\neg 2}}\mathbf H_2^{(i)} - {\mathbf V_{{}^\neg 2}}} \right\|_2^2} \\
+  \cdots  + \sum\nolimits_{i,j}^{{n_c}} {\left\| {\mathbf H_c^{(i)}\mathbf U - \mathbf H_c^{(j)}\mathbf U} \right\|_2^2s_c^{(ij)}} \\
- \beta \sum\nolimits_{i = 1}^{{n_c}} {\left\| {{\mathbf I_{{}^\neg c}}\mathbf H_c^{(i)} - {\mathbf V_{{}^\neg c}}} \right\|_2^2}  + \lambda \left\| {\mathbf H - {\mathbf X^T}{\mathbf U^T}} \right\|_F^2
\end{array}
\end{eqnarray}

In Eq. (\ref{eq32}), we can find that $\mathbf H = [{\mathbf H_1};{\mathbf H_2}; \ldots ;{\mathbf H_c}] \in {\mathbb R^{n \times l}}$. Therefore, we still use the alternating iteration method to solve it. \ie when fixing ${\mathbf H_2}, \ldots ,{\mathbf H_c}$ to solve $\mathbf H_1$, Eq. (\ref{eq32}) can be written as follows:
\begin{eqnarray}
\label{eq33}
\begin{array}{l}
\mathop {\min }\limits_\mathbf H \sum\nolimits_{i,j}^{{n_1}} {\left\| {\mathbf H_1^{(i)}\mathbf U - \mathbf H_1^{(j)}\mathbf U} \right\|_2^2s_1^{(ij)}} \\
- \beta \sum\nolimits_{i = 1}^{{n_1}} {\left\| {{\mathbf I_{{}^\neg 1}}\mathbf H_1^{(i)} - {\mathbf V_{{}^\neg 1}}} \right\|_2^2}  + \lambda \left\| {{\mathbf H_1} - {\mathbf X_1}^T{\mathbf U^T}} \right\|_F^2
\end{array}
\end{eqnarray}
where ${\mathbf V_{{}^\neg 1}} \in {\mathbb R^{(c - 1) \times l}}$ represents the mean matrix composed of the mean values of hash codes in each hyper-class except the first hyper-class. Eq. (\ref{eq33}) can be further written as follows:
\begin{eqnarray}
\label{eq34}
\begin{array}{l}
\mathop {\min }\limits_\mathbf H  - \beta \sum\nolimits_{i = 1}^{{n_1}} {\left( \begin{array}{l}
	\left\langle {{\mathbf I_{{}^\neg 1}}\mathbf H_1^{(i)},{\mathbf I_{{}^\neg 1}}\mathbf H_1^{(i)}} \right\rangle  - 2\left\langle {{\mathbf I_{{}^\neg 1}}\mathbf H_1^{(i)},{\mathbf V_{\neg 1}}} \right\rangle \\
	+ \left\langle {{\mathbf V_{\neg 1}},{\mathbf V_{\neg 1}}} \right\rangle 
	\end{array} \right)} \\
+ \lambda \left\| {{\mathbf H_1} - {\mathbf X_1}^T{\mathbf U^T}} \right\|_F^2 + tr({\mathbf U^T}{\mathbf H_1}^T{\mathbf L_1}{\mathbf H_1}\mathbf U)
\end{array}
\end{eqnarray}

Because $tr({\mathbf U^T}{\mathbf H_1}^T{\mathbf L_1}{\mathbf H_1}\mathbf U) = tr(\mathbf U{\mathbf U^T}{\mathbf H_1}^T{\mathbf L_1}{\mathbf H_1})$ and $\mathbf U{\mathbf U^T} = \mathbf I$. Therefore, the first term in Eq. (\ref{eq34}) can be written as $tr({\mathbf H_1}^T{\mathbf L_1}{\mathbf H_1})$. We use Eq. (\ref{eq34}) to find the derivative of $\mathbf H_1^{(i)}$, and make derivative is 0, so we can get the following formula:
\begin{eqnarray}
\label{eq35}
\begin{array}{l}
\mathbf L_1^{(i)}[{({\mathbf E^{(i)}})^T}\mathbf H_1^{(i)} + \sum\limits_{j \ne i}^{{n_1} - 1} {{{({\mathbf E^{(j)}})}^T}\mathbf H_1^{(j)}} ] \\+ (\beta  + \lambda )\hat H_1^{(i)} - \lambda {({\mathbf X_1}^T{\mathbf U^T})^i} =  - \beta \mathbf I_{{}^\neg 1}^T{\mathbf V_{{}^\neg 1}}
\end{array}
\end{eqnarray}

According to Eq. (\ref{eq35}), we can get the solution of $\mathbf H_1$ as follows:
\begin{eqnarray}
\label{eq36}
\begin{array}{l}
\hat H_1^{(i)} = {(\mathbf L_1^{(i)}{({\mathbf E^{(i)}})^T} - \beta  + \lambda )^{ - 1}}\\(\lambda {({\mathbf X_1}^T{\mathbf U^T})^i} - \mathbf L_1^{(i)}\sum\limits_{j \ne i}^{{n_1} - 1} {{{({\mathbf E^{(j)}})}^T}\mathbf H_1^{(j)}}  - \beta \mathbf I_{{}^\neg 1}^T{\mathbf V_{{}^\neg 1}})
\end{array}
\end{eqnarray}

Similarly, the solution of ${\mathbf H_2},{\mathbf H_3}, \ldots ,{\mathbf H_c}$ can be obtained as follows:
\begin{eqnarray}
\label{eq37}
\begin{array}{l}
\left\{ {\begin{array}{*{20}{c}}
	\begin{array}{l}
	\hat H_2^{(i)} = {(\mathbf L_2^{(i)}{({\mathbf E^{(i)}})^T} - \beta  + \lambda )^{ - 1}}\\
	(\lambda {({\mathbf X_2}^T{\mathbf U^T})^i} - \mathbf L_2^{(i)}\sum\limits_{j \ne i}^{{n_2} - 1} {{{({\mathbf E^{(j)}})}^T}\mathbf H_2^{(j)}}  - \beta \mathbf I_{{}^\neg 2}^T{\mathbf V_{{}^\neg 2}})
	\end{array}\\
	\begin{array}{l}
	\hat H_3^{(i)} = {(\mathbf L_3^{(i)}{({\mathbf E^{(i)}})^T} - \beta  + \lambda )^{ - 1}}\\
	(\lambda {({\mathbf X_3}^T{\mathbf U^T})^i} - \mathbf L_3^{(i)}\sum\limits_{j \ne i}^{{n_3} - 1} {{{({\mathbf E^{(j)}})}^T}\mathbf H_3^{(j)}}  - \beta \mathbf I_{{}^\neg 3}^T{\mathbf V_{{}^\neg 3}})
	\end{array}\\
	\vdots \\
	\begin{array}{l}
	\hat H_c^{(i)} = {(\mathbf L_c^{(i)}{({\mathbf E^{(i)}})^T} - \beta  + \lambda )^{ - 1}}\\
	(\lambda {({\mathbf X_c}^T{\mathbf U^T})^i} - \mathbf L_c^{(i)}\sum\limits_{j \ne i}^{{n_c} - 1} {{{({\mathbf E^{(j)}})}^T}\mathbf H_c^{(j)}}  - \beta \mathbf I_{{}^\neg c}^T{\mathbf V_{{}^\neg c}})
	\end{array}
	\end{array}} \right.
\end{array}
\end{eqnarray}

After we get the solution of ${\mathbf H_1},{\mathbf H_2}, \ldots ,{\mathbf H_c}$, we can splice them together and get the solution of $\mathbf H$.

When $\mathbf H$ and $\mathbf S$ are fixed to solve $\mathbf U$, Eq. (\ref{eq18}) can be written as follows:
\begin{eqnarray}
\label{eq38}
\begin{array}{l}
\mathop {\min }\limits_U \sum\nolimits_{i,j}^{{n_1}} {\left\| {\mathbf H_1^{(i)}\mathbf U - \mathbf H_1^{(j)}\mathbf U} \right\|_2^2s_1^{(ij)}} \\
+ \sum\nolimits_{i,j}^{{n_2}} {\left\| {\mathbf H_2^{(i)}\mathbf U - \mathbf H_2^{(j)}\mathbf U} \right\|_2^2s_2^{(ij)}} \\
+  \cdots  + \sum\nolimits_{i,j}^{{n_c}} {\left\| {\mathbf H_c^{(i)}\mathbf U - \mathbf H_c^{(j)}\mathbf U} \right\|_2^2s_c^{(ij)}} \\
+ \lambda \left\| {\mathbf H - {\mathbf X^T}{\mathbf U^T}} \right\|_F^2 + \eta \left\| \mathbf U \right\|_2^2\\
s.t.~{\rm{ }}\mathbf U{\mathbf U^T} = \mathbf I
\end{array}
\end{eqnarray}

Further, Eq. (\ref{eq38}) can be written as follows:
\begin{eqnarray}
\label{eq39}
\begin{array}{l}
\mathop {\min }\limits_\mathbf H tr({\mathbf U^T}{\mathbf H_1}^T{\mathbf L_1}{\mathbf H_1}\mathbf U) + tr({\mathbf U^T}{\mathbf H_2}^T{\mathbf L_2}{\mathbf H_2}\mathbf U)\\
+  \cdots  + tr({\mathbf U^T}{\mathbf H_c}^T{\mathbf L_c}{\mathbf H_c}\mathbf U)\\
+ \lambda (({\mathbf H^T} - \mathbf U\mathbf X)(\mathbf H - {\mathbf X^T}{\mathbf U^T})) + \eta \left\| \mathbf U \right\|_2^2\\
s.t.~{\rm{ }}\mathbf U{\mathbf U^T} = \mathbf I
\end{array}
\end{eqnarray}

We use Eq. (\ref{eq39}) to derive $\mathbf U$, and we can get:
\begin{eqnarray}
\label{eq40}
\begin{array}{l}
\frac{{\partial f(\mathbf U)}}{{\partial \mathbf U}} = 2{\mathbf H_1}^T{\mathbf L_1}{\mathbf H_1}\mathbf U + 2{\mathbf H_2}^T{\mathbf L_2}{\mathbf H_2}\mathbf U\\
+  \cdots  + 2{\mathbf H_c}^T{\mathbf L_c}{\mathbf H_c}\mathbf U\\
+ \lambda (2\mathbf U\mathbf X{\mathbf X^T} - 2{\mathbf H^T}{\mathbf X^T}) + 2\eta \mathbf I \mathbf U
\end{array}
\end{eqnarray}

Because Eq. (\ref{eq39}) has an orthogonal restriction on $\mathbf U$, \ie $\mathbf U{\mathbf U^T} = \mathbf I$, we still use the method in literature \cite{wen2013feasible} to solve it.

When $\mathbf H$ and $\mathbf U$ are fixed to solve $\mathbf S$, Eq. (\ref{eq18}) can be written as follows:
\begin{eqnarray}
\label{eq41}
\begin{array}{l}
\mathop {\min }\limits_{\mathbf H,\mathbf U,\mathbf S} \sum\nolimits_{i,j}^{{n_1}} {\left\| {\mathbf H_1^{(i)}\mathbf U - \mathbf H_1^{(j)}\mathbf U} \right\|_2^2s_1^{(ij)}}  + \alpha \sum\nolimits_i^{{n_1}} {\left\| {\mathbf s_1^{(i)}} \right\|_2^2} \\
+ \sum\nolimits_{i,j}^{{n_2}} {\left\| {\mathbf H_2^{(i)}\mathbf U - \mathbf H_2^{(j)}\mathbf U} \right\|_2^2s_2^{(ij)}}  + \alpha \sum\nolimits_i^{{n_2}} {\left\| {\mathbf s_2^{(i)}} \right\|_2^2} \\
+  \cdots  + \sum\nolimits_{i,j}^{{n_c}} {\left\| {\mathbf H_c^{(i)}\mathbf U - \mathbf H_c^{(j)}\mathbf U} \right\|_2^2s_c^{(ij)}}  + \alpha \sum\nolimits_i^{{n_c}} {\left\| {\mathbf s_c^{(i)}} \right\|_2^2} \\
s.t.~{\rm{ }}\mathbf s_1^{{{(i)}^T}}\mathbf 1 = 1,~\mathbf s_2^{{{(i)}^T}}\mathbf 1 = 1, \ldots ,~\mathbf s_c^{{{(i)}^T}}1 = 1,\\
s_1^{(ii)} = 0,~s_2^{(ii)} = 0, \ldots ,~s_c^{(ii)} = 0\\
s_1^{(ij)},~s_2^{(ij)}, \ldots s_c^{(ij)} \ge 0,if{\rm{ }}j \in N(i),~{\rm{ }}otherwise{\rm{ }}0
\end{array}
\end{eqnarray}

Since $\mathbf S = {[{\mathbf s_1},{\mathbf s_2}, \ldots ,{\mathbf s_c}]^T}$, we still adopt the alternating iterative method to solve it. \ie  when fixing ${\mathbf s_2}, \ldots ,{\mathbf s_c}$ to solve $\mathbf s_1$, Eq. (\ref{eq41}) can be written as follows:
\begin{eqnarray}
\label{eq42}
\begin{array}{l}
\mathop {\min }\limits_{\mathbf H,\mathbf U,\mathbf S} \sum\nolimits_{i,j}^{{n_1}} {\left\| {\mathbf H_1^{(i)}\mathbf U - \mathbf H_1^{(j)}\mathbf U} \right\|_2^2s_1^{(ij)}}  + \alpha \sum\nolimits_i^{{n_1}} {\left\| {\mathbf s_1^{(i)}} \right\|_2^2} \\
s.t.~{\rm{ }}\mathbf s_1^{{{(i)}^T}}1 = 1,~s_1^{(ii)} = 0\\
s_1^{(ij)} \ge 0,~if{\rm{ }}j \in N(i),~{\rm{ }}otherwise{\rm{ }}0
\end{array}
\end{eqnarray}

According to the method of solving $\mathbf S$ when $c=1$, \ie Eq. (\ref{eq29}), we can get the solution of $\mathbf s_1$ as follows:
\begin{eqnarray}
\label{eq43}
\begin{array}{l}
s_1^{(i,j)} = {( - \frac{1}{{2\alpha }}Z_1^{(i,j)} + {\tau _1})_ + }
\end{array}
\end{eqnarray}
where $\mathbf Z \in {R^{{n_1} \times {n_1}}}$, $Z_1^{(i,j)} = \left\| {\mathbf H_1^{(i)}\mathbf U - \mathbf H_1^{(j)}\mathbf U} \right\|_2^2$, ${\tau _1} = \frac{1}{k} + \frac{1}{{2k\alpha }}\sum\nolimits_{j = 1}^k {\hat Z_1^{i,k}}$. Similarly, we can get the solution of ${\mathbf s_2}, \ldots ,{\mathbf s_c}$ as follows:
\begin{eqnarray}
\label{eq44}
\begin{array}{l}
\left\{ {\begin{array}{*{20}{c}}
	{s_2^{(i,j)} = {{( - \frac{1}{{2\alpha }}Z_2^{(i,j)} + {\tau _2})}_ + }}\\
	{s_3^{(i,j)} = {{( - \frac{1}{{2\alpha }}Z_3^{(i,j)} + {\tau _3})}_ + }}\\
	\vdots \\
	{s_c^{(i,j)} = {{( - \frac{1}{{2\alpha }}Z_c^{(i,j)} + {\tau _c})}_ + }}
	\end{array}} \right.
\end{array}
\end{eqnarray}
where $Z_2^{(i,j)},Z_3^{(i,j)}, \ldots ,Z_c^{(i,j)}$ and ${\tau _2},{\tau _3}, \ldots ,{\tau _c}$ are respectively as follows:
\begin{eqnarray}
\label{eq45}
\begin{array}{l}
\left\{ {\begin{array}{*{20}{c}}
	{Z_2^{(i,j)} = \left\| {\mathbf H_2^{(i)}\mathbf U - \mathbf H_2^{(j)}\mathbf U} \right\|_2^2}\\
	{Z_3^{(i,j)} = \left\| {\mathbf H_3^{(i)}\mathbf U - \mathbf H_3^{(j)}\mathbf U} \right\|_2^2}\\
	\vdots \\
	{Z_c^{(i,j)} = \left\| {\mathbf H_c^{(i)}\mathbf U - \mathbf H_c^{(j)}\mathbf U} \right\|_2^2}
	\end{array}} \right.
\end{array}
\end{eqnarray}
\begin{eqnarray}
\label{eq46}
\begin{array}{l}
\left\{ {\begin{array}{*{20}{c}}
	{{\tau _2} = \frac{1}{k} + \frac{1}{{2k\alpha }}\sum\nolimits_{j = 1}^k {\hat Z_2^{i,k}} }\\
	{{\tau _3} = \frac{1}{k} + \frac{1}{{2k\alpha }}\sum\nolimits_{j = 1}^k {\hat Z_3^{i,k}} }\\
	\vdots \\
	{{\tau _c} = \frac{1}{k} + \frac{1}{{2k\alpha }}\sum\nolimits_{j = 1}^k {\hat Z_c^{i,k}} }
	\end{array}} \right.
\end{array}
\end{eqnarray}

So far, the optimization process of the proposed algorithm is completed.

\begin{algorithm}
	\setlength{\algomargin}{0.5em}
	\SetAlgoLined
	\caption{\label{alg1} Training stage of HCH.}
	\IncMargin{0.5em}
	\KwIn{Training set $\mathbf{X} \in \mathbb{R}^{d \times n}$, $c$ (the number of clusters), $l$ (the number of bits), adjustable parameters $\varsigma$, $\sigma$, $\alpha$, $\beta$, $\lambda$ and $\eta$ \;}
	\KwOut{$\mathbf v \in {\mathbb R^{d \times 1}}$, $\mathbf B \in {\mathbb R^{n \times l}}$, $
		\mathbf U \in {\mathbb R^{l \times d}}$;}
	
	Initialize \emph t=0\;
	Randomly initialize $\mathbf W^{(0)}$ and $\mathbf v^{(0)}$\;
	\For{$i=1 \to d $}
	{ 
		Calculate $k({\mathbf X_{{}^\neg i}})$ via Eq. (\ref{eq3})\;	
	}	
	\Repeat{converge}{
		Update $\mathbf F^{(t)}$ via 
		${F_{ii}} = \frac{1}{{2\sqrt {\left\| {{\mathbf W^i}} \right\|_2^2 + \varepsilon } }}$\;
		Update $\mathbf W^{(t+1)}$ via Eq. (\ref{eq10})\;
		Updata $\mathbf v^{(t+1)}$ via Eq. (\ref{eq13})\;
		\emph{t} = \emph{t}+1 \;
	}	
	Get the potential decision feature $df$ via Eq. (\ref{eq14})\;
	The method in reference \cite{makarychev2022performance} is used to cluster or divide the potential decision feature, so as to construct hyper-class\;
	Initialize \emph t=0, initialize $\mathbf S$ to all zero matrix, randomly initialize $\mathbf U^{(0)}$ and $\mathbf H^{(0)}$\;
	\Repeat{converge}{
		Update $\mathbf H^{(t+1)}$ via Eqs. (\ref{eq36}) and (\ref{eq37}) \;
		Update $\mathbf U^{(t+1)}$ via Eq. (\ref{eq40}) and Reference \cite{wen2013feasible}\;
		Updata $\mathbf S^{(t+1)}$ via Eqs. (\ref{eq43}) and (\ref{eq44})\;
		\emph{t} = \emph{t}+1 \;
	}
	Calculate $\mathbf B$ via $sgn(\mathbf X^T\mathbf U^T)$ for all the data points $\mathbf X$;

\end{algorithm}

\begin{algorithm}
	\setlength{\algomargin}{0.5em}
	\SetAlgoLined
	\caption{\label{alg2}  Testing stage of HCH.}
	\IncMargin{0.5em}
	\KwIn{$\mathbf y \in {\mathbb R^d}$ and $\mathbf U \in {\mathbb R^{l \times d}}$\;}
	\KwOut{${\mathbf b_y} \in {\mathbb R^l}$;}
	Calculate $\mathbf b_y$ via $sgn(\mathbf y^T\mathbf U^T)$\;
	Calculate the Hamming distance between $\mathbf b_y$ and $\mathbf B$;

\end{algorithm}

\subsection{Summary of algorithm}

As shown in algorithm \ref{alg1} and algorithm \ref{alg2}, the proposed HCH algorithm can be divided into four steps, \ie 1. Use the relationship between each feature and all other features to learn the potential decision features in the data, as shown in steps 1-12. 2. establish hyper-class according to the obtained decision features and clustering algorithm, as shown in step 13. 3. construct the hash learning objective function according to the hyper-class in the data, \ie Eq. (\ref{eq18}). And optimize the objective function, as shown in step 14-20. 4. perform binary hash coding on the data according to the learned $\mathbf U$ matrix.

In addition, in the proposed HCH algorithm, we use the alternative iterative optimization method to solve the objective function, which can make the algorithm converge quickly\cite{chen2019extended}. In section 4.4, we also verify its convergence.

\subsection{Time complexity analysis}

In the proposed HCH algorithm, in the training phase, the time complexity of finding decision features, clustering method, hash function learning and binarization are $O(t({n^2}d + d))$, $O(dcn)$, $O(t(n{l^2} + ndl + {n^2}l))$ and $O(nl)$ respectively. Where, $t$ represents the number of iterations, $n$ represents the number of samples, $d$ represents the number of features, $l$ represents the number of bits, and $c$ represents the number of clusters. In the test phase, we need to spend $O(dl)$ time to binarize the test data and $O(1)$ time to perform reverse lookup in the hash table. Since $l$ is usually less than $d$ and $t$ is a small number, the time complexity of HCH is $O({n^2}d + dcn)$ in the training phase and $O(dl)$ in the testing phase. In addition, we also list the time complexity comparison of some popular hash algorithms, such as ITQ, FastH, KNNH, CH, LSH, MDSH and ALECH, as shown in Table \ref{tab2}.

In the training phase, the spatial complexity of HCH is $O(d(c + n))$, and that in the test phase is $O((d + m)l)$.

\begin{table}[!tb]
	\centering
	\caption{\footnotesize Time complexity of comparison algorithm.}
	\centering
	{\footnotesize
		\begin{tabular}[c]{|c|c|c|} \hline
			            &Training phase 	&Test phase \\ \hline
			ITQ             &$O({d^2}n + {l^3})$ 	&$O(dl)$  \\ \hline
			FastH             &$O({d^2}n)$     &$O(dl)$  \\ 	\hline
			KNNH	         &$O({n^2}(d + \log n))$ 	&$O(1)$   \\ \hline
			CH           &$O(2t{d^2}n + tncd)$       & $O(dl)$   \\ \hline
			LSH           &$O(1)$        &$O(1)$     \\ \hline			
			MDSH	  &$O({d^2}n)$        &  $O(dl)$       \\ \hline
			ALECH    &$O(nld + {d^3} + n{d^2})$        &    $O(dl)$       \\  \hline
			HCH            &$O({n^2}d + dcn)$       & $O(dl)$       \\  \hline			
		\end{tabular}}
	 \begin{tablenotes}
	 	\footnotesize
	 	\item where $t$ represents the number of iterations, $n$ represents the number of samples, $d$ represents the number of features, $l$ represents the number of bits, and $c$ represents the number of clusters.
	 \end{tablenotes}

		\label{tab2}
	\end{table}

\section{Experiments} \label{experiments}

In this section, we compare the performance of the proposed algorithm with seven comparison algorithms on four real datasets.

\begin{figure*}[!ht]
	\begin{center}
		\vspace{-1mm}
		\subfigure[Mnist]{\scalebox{0.6}{\includegraphics{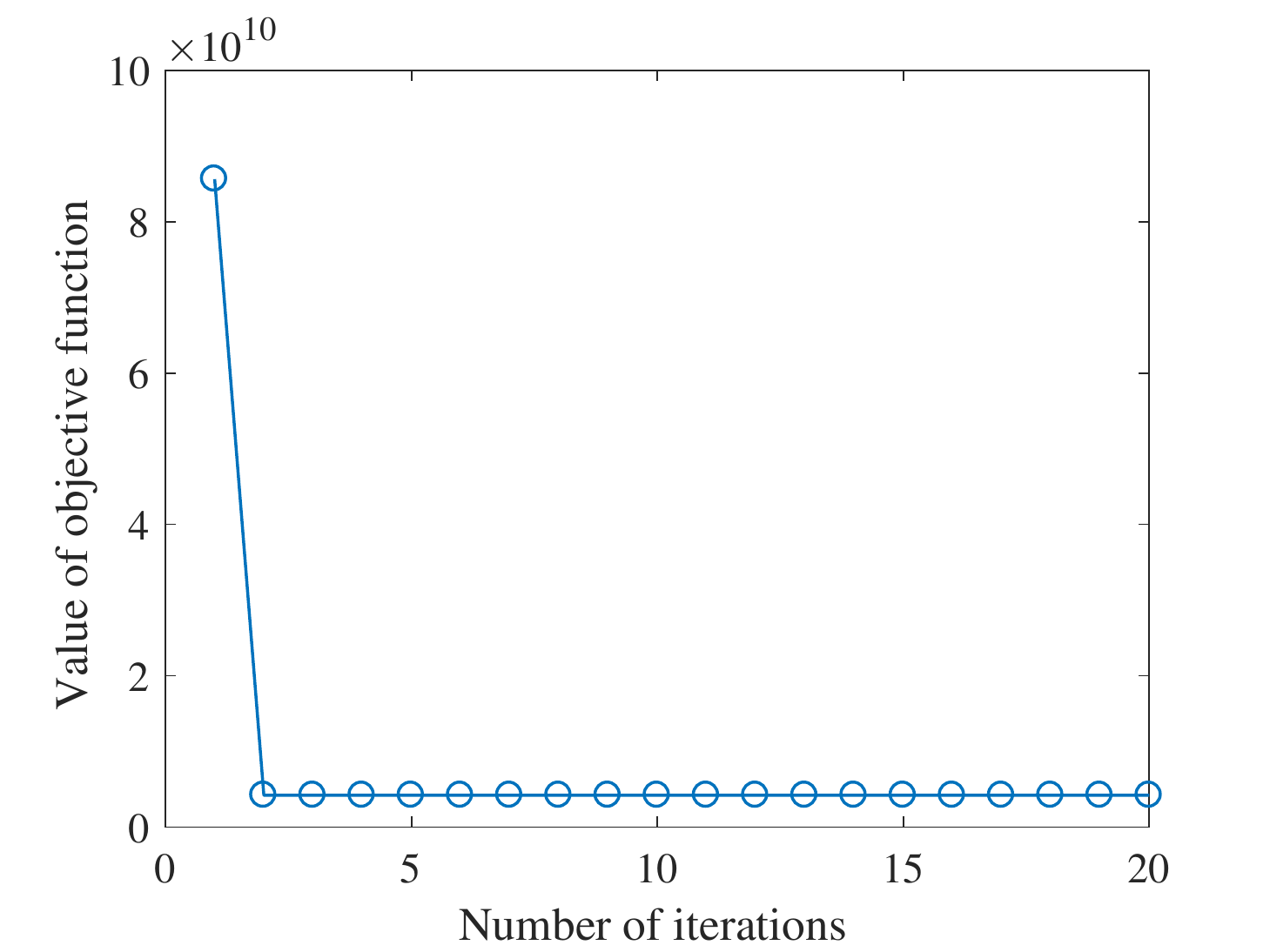}}}	
		\vspace{-1mm}
		\subfigure[Cifar]{\scalebox{0.6} {\includegraphics{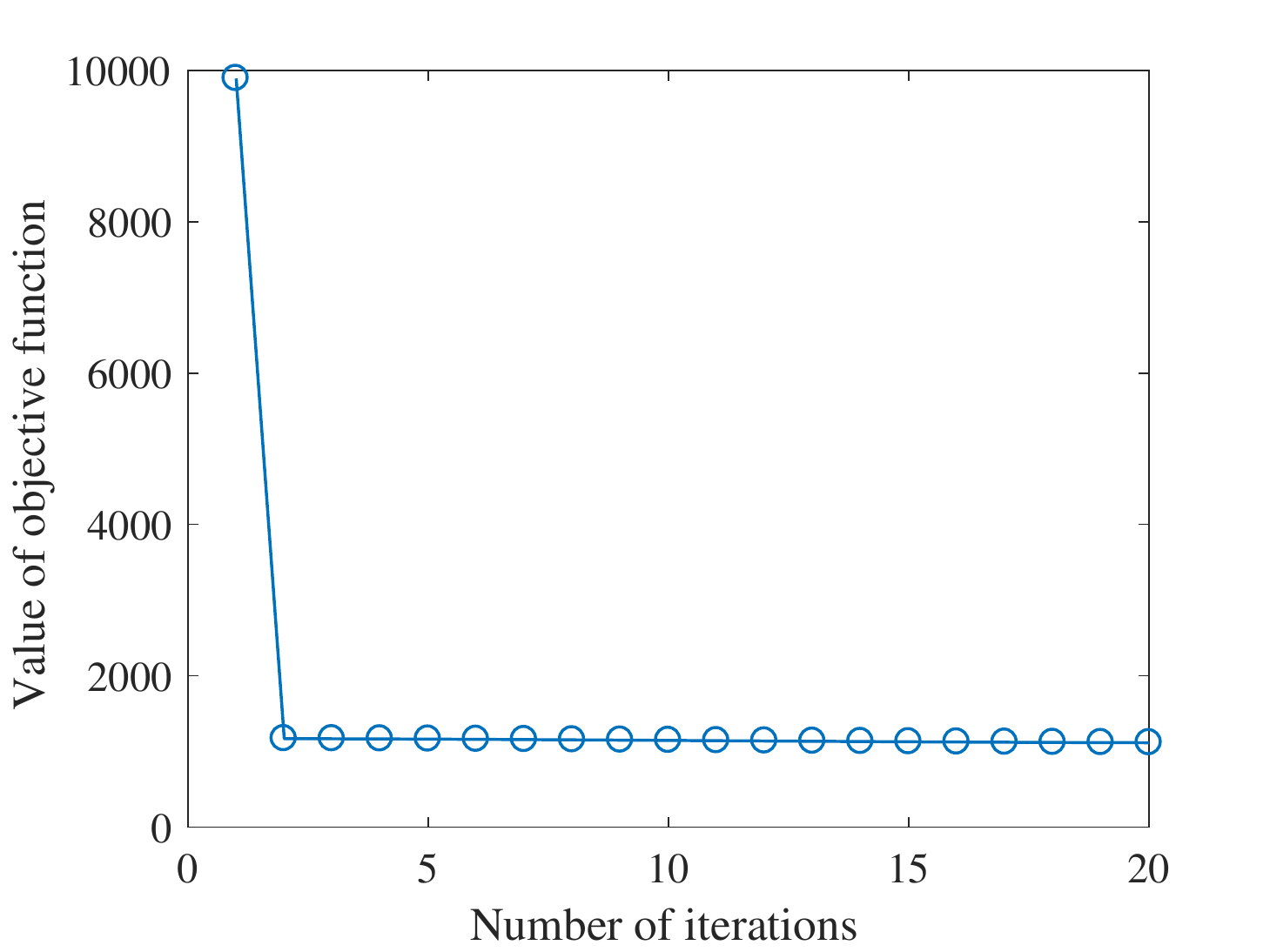}}}			
		\vspace{-1mm}
		\subfigure[Labelme]{\scalebox{0.6} {\includegraphics{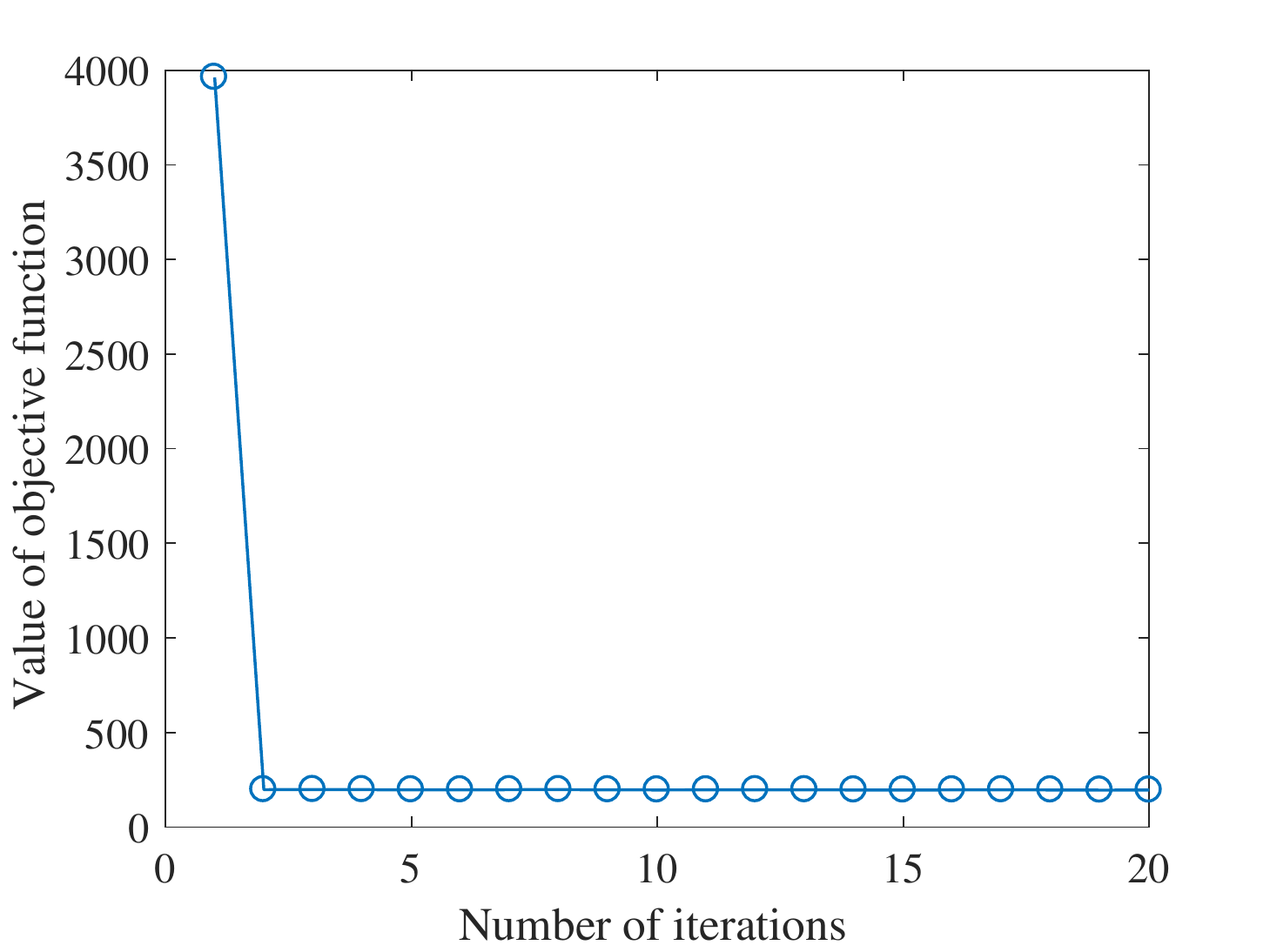}}}
		\vspace{-1mm}		
		\subfigure[Place]{\scalebox{0.6} {\includegraphics{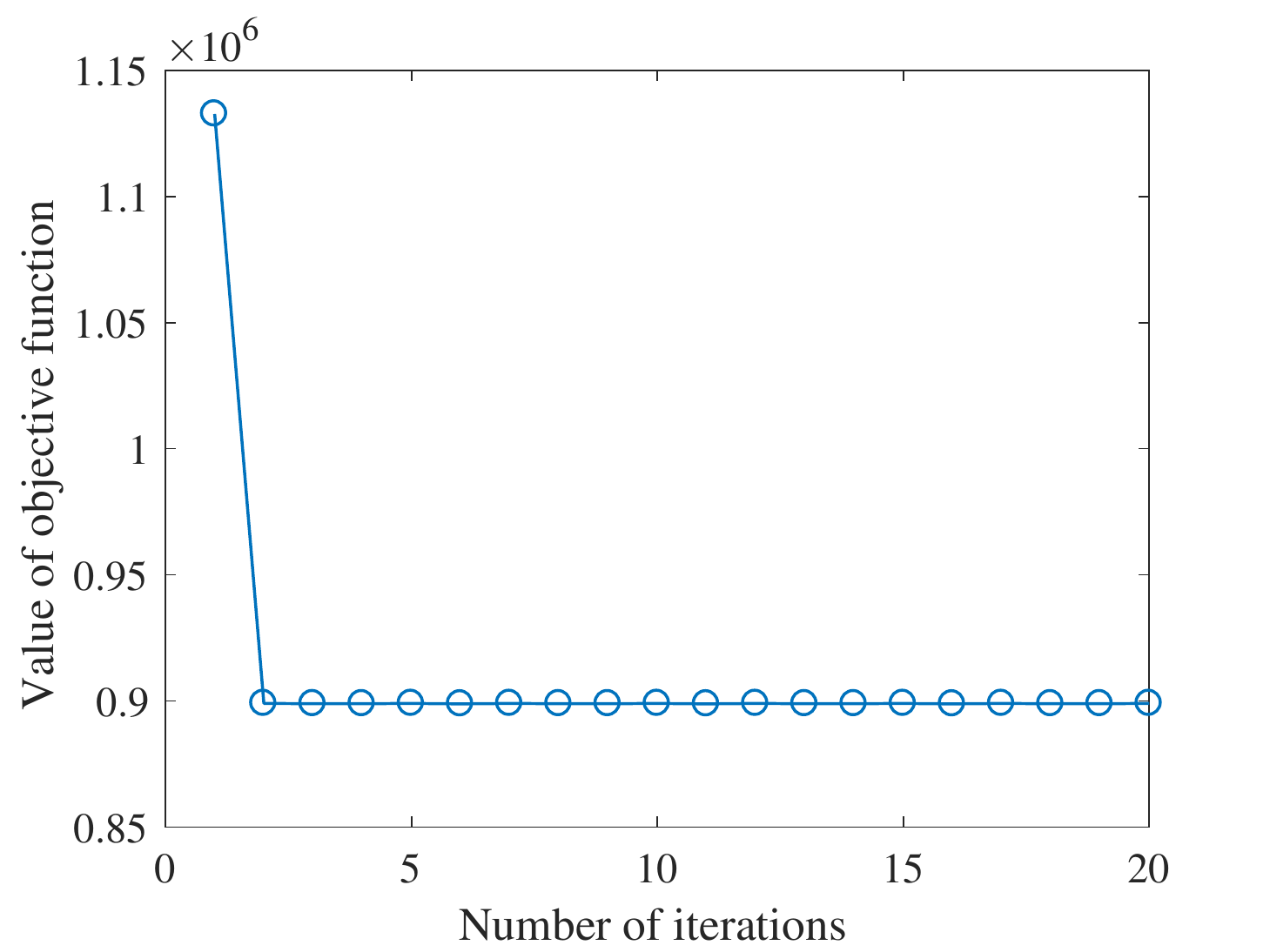}}}		
		\caption{The value of each iteration of the objective function on all datasets}
		\label{fig3}
	\end{center}
\end{figure*}

\subsection{Datasets}

In our experiment, we used four real datasets, namely, Mnist, Cifar, Labelme and Place. Their details are as follows:

The Mnist dataset contains 70000 images and their labels. Each image is represented by a 784 dimensional vector. 60000 of them are training sets and 10000 are testing sets. In the testing sets, the first 5000 data are more regular than the last 5000 data. The reason is that they come from different sources.

The Cifar dataset was collected by Alex krizhevsky, Vinod Nair, and Geoffrey Hinton. It contains a total of 60000 color images, each of which is 32$\times$32. It has a total of 10 classes, each containing 6000 images. In this paper, we use 50000 images as the training set and 10000 images as the testing set.

The Labelme dataset contains a total of 50000 images, and each image is a 256$\times$256 jpeg image. Among them, 45007 images are used as the training set and 4993 images are used as the testing set.

There are 2448872 images in the Place dataset. They can be divided into 205 classes, and each sample has 128 features.

\begin{figure*}[!ht]
	\begin{center}
		\vspace{-1mm}
		\subfigure[Mnist]{\scalebox{0.6}{\includegraphics{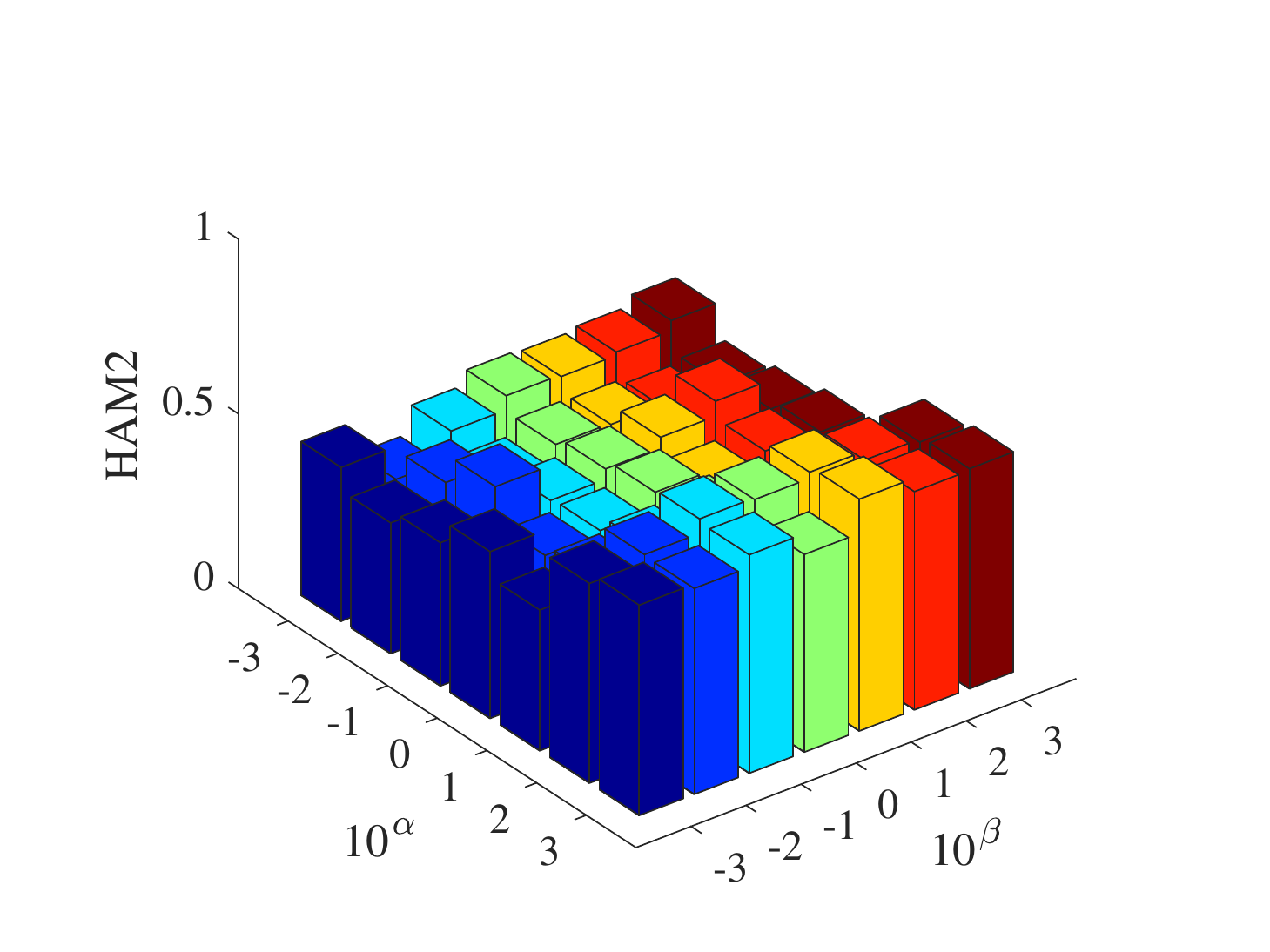}}}	
		\vspace{-1mm}
		\subfigure[Cifar]{\scalebox{0.6} {\includegraphics{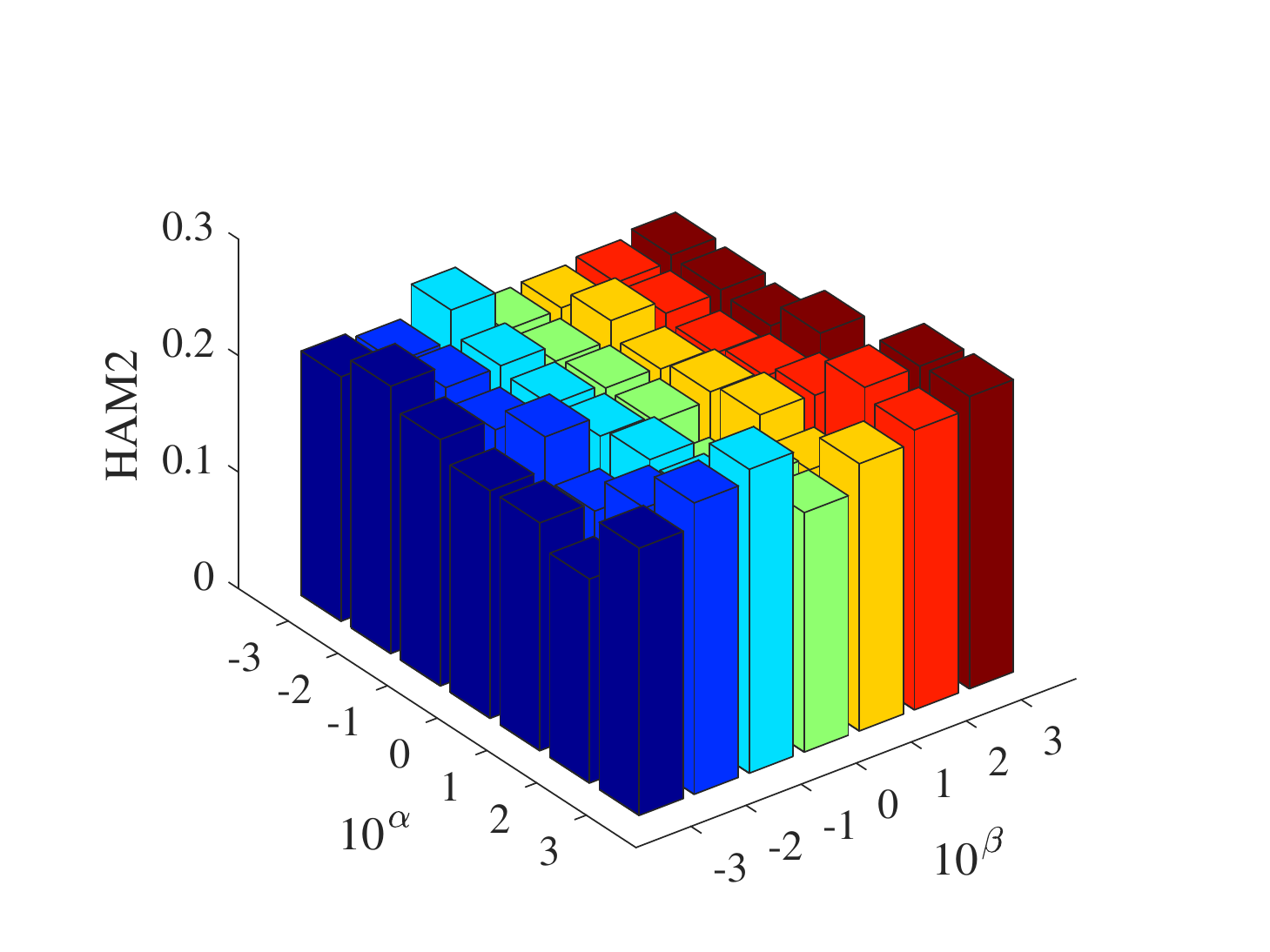}}}			
		\vspace{-1mm}
		\subfigure[Labelme]{\scalebox{0.6} {\includegraphics{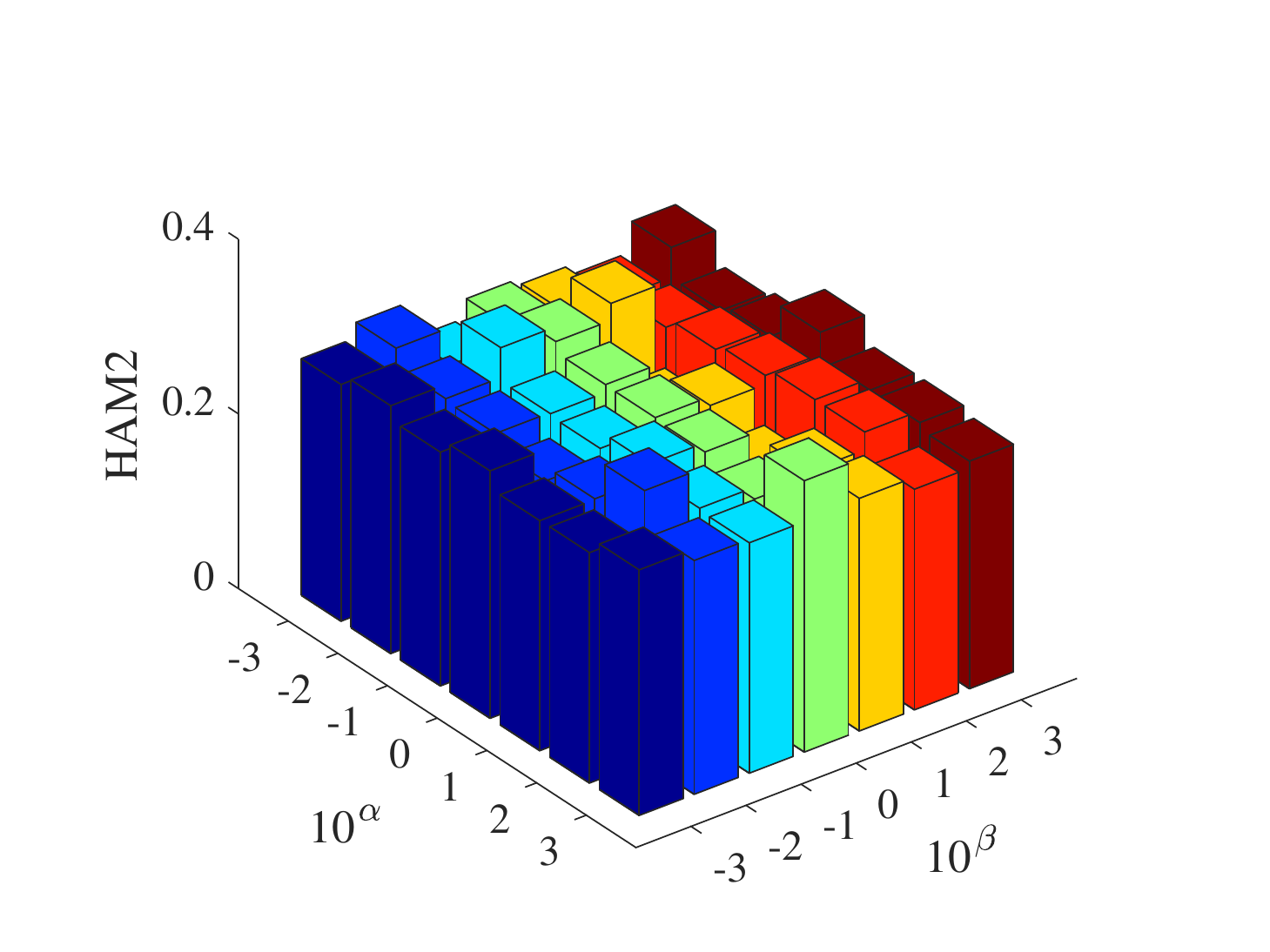}}}
		\vspace{-1mm}		
		\subfigure[Place]{\scalebox{0.6} {\includegraphics{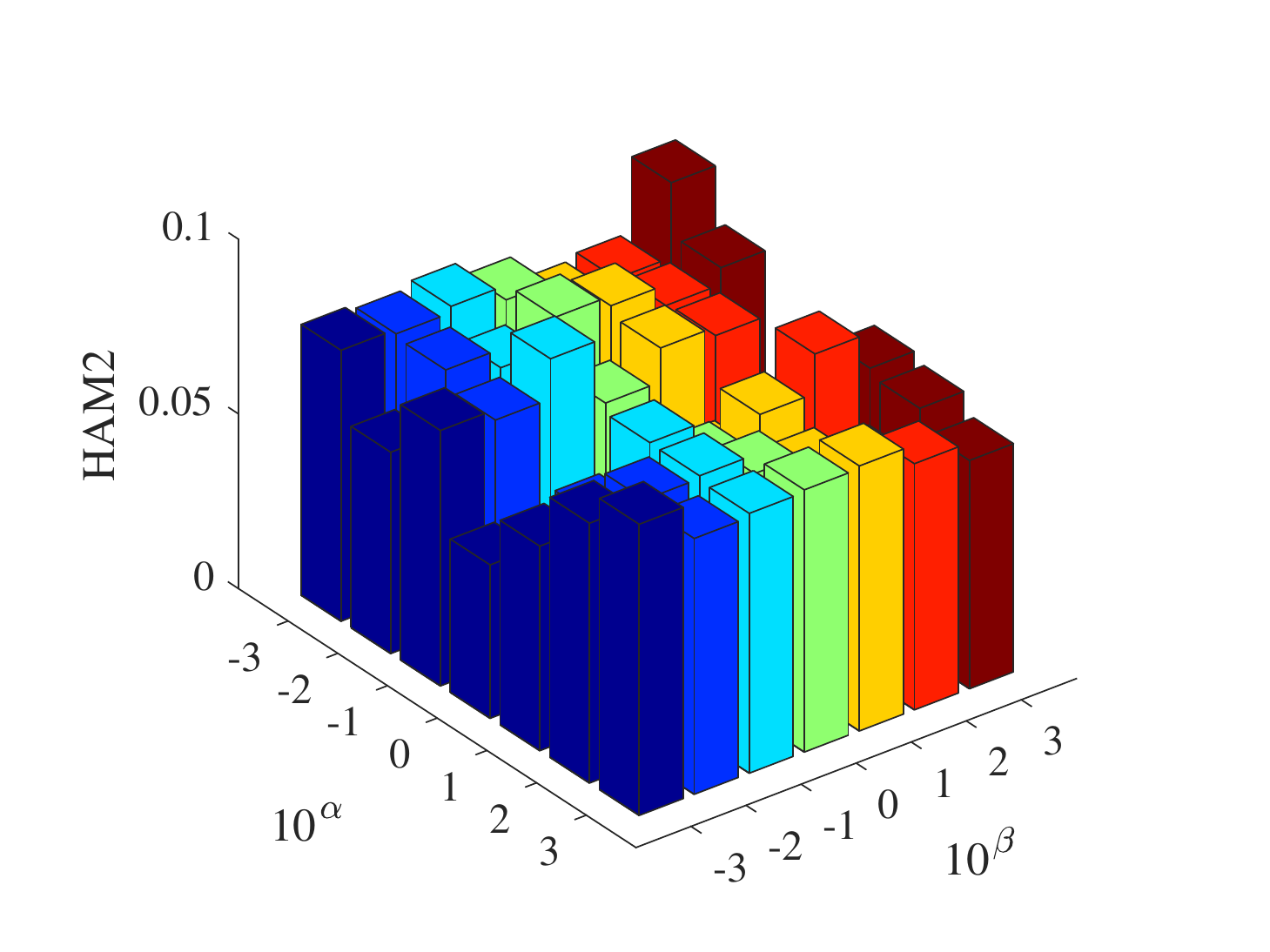}}}		
		\caption{The HAM2 of our proposed method with different parameters' setting \ie $\alpha$ and $\beta$}
		\label{fig4}
	\end{center}
\end{figure*}

\subsection{Comparied Algorithms}

ITQ\cite{gong2012iterative}: this method is a classic data dependent hash algorithm. It learns the binary hash code of the data by minimizing the quantization error between the data and the vertex of the zero center binary hypercube. In addition, it can be used for unsupervised PCA embedding and supervised canonical correlation analysis (CCA) embedding.

FastH\cite{lin2014fast}: it is a supervised hash algorithm. Different from other nonlinear hash algorithms, it uses decision tree to mine nonlinear relationships in data rather than kernel functions. Specifically, it first proposes a sub module formula for hash code binary reasoning. Then it trains the enhanced decision tree to adapt it to binary hash codes. Finally, it verifies the effect on high-dimensional data in experiments.

LSH\cite{andoni2017optimal}: this method is a locally sensitive hash algorithm. Its core idea is to divide the random space of data. Given a query sample, it first searches in the same divided space as the query sample. In other words, it improves query speed by shortening the retrieval range of data.

MDSH\cite{weiss2012multidimensional}: it is a multi-dimensional hash algorithm. It aims to establish the affinity relationship between data rather than the distance between data. Specifically, it first establishes a formula for learning binary coding to obtain the affinity matrix, and then it uses the threshold eigenvector of the affinity matrix to obtain bits. Finally, in the experiment, it also shows some results with the increase of the number of bits.

KNNH\cite{he2019k}: it is a KNN hash algorithm. Specifically, it first uses PCA algorithm to reduce the dimension of the original data. Then KNN algorithm is used to find k nearest neighbors of each data. Finally, it learns the binary representation in each subspace of the data.

CH\cite{weng2020concatenation}: this method is a hash algorithm based on clustering technology. Different from other hash algorithms based on clustering technology, it mainly focuses on reducing the influence of clustering boundary samples. It uses alternating iterative algorithm to carry out hash learning and clustering at the same time, so as to maintain the relative position of each data to the cluster center. In addition, it also splices the hash function learning in each cluster to obtain the binary code corresponding to all data.

ALECH\cite{li2021adaptive}: this method is a hash algorithm based on adaptive label correlation. Specifically, it first uses the least square loss to obtain the relationship between semantic labels and hash codes. Then it uses the alternating iterative optimization method to optimize and solve the proposed objective function. Finally, it maps the data to the kernel space to obtain the hash function. This method is a supervised hash algorithm.

\begin{figure*}[!ht]
	\begin{center}
		\vspace{-1mm}
		\subfigure[Mnist]{\scalebox{0.6}{\includegraphics{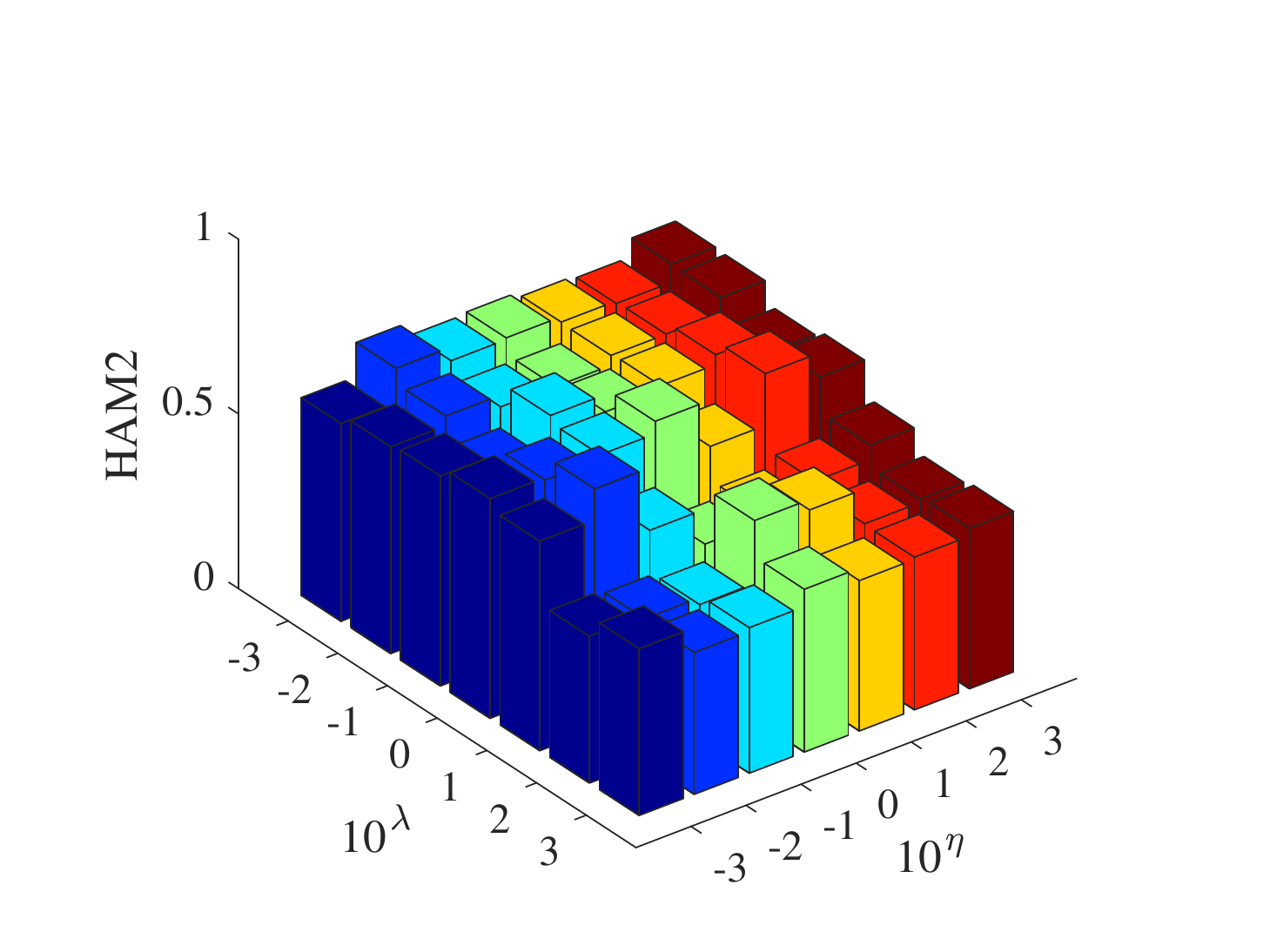}}}	
		\vspace{-1mm}
		\subfigure[Cifar]{\scalebox{0.6} {\includegraphics{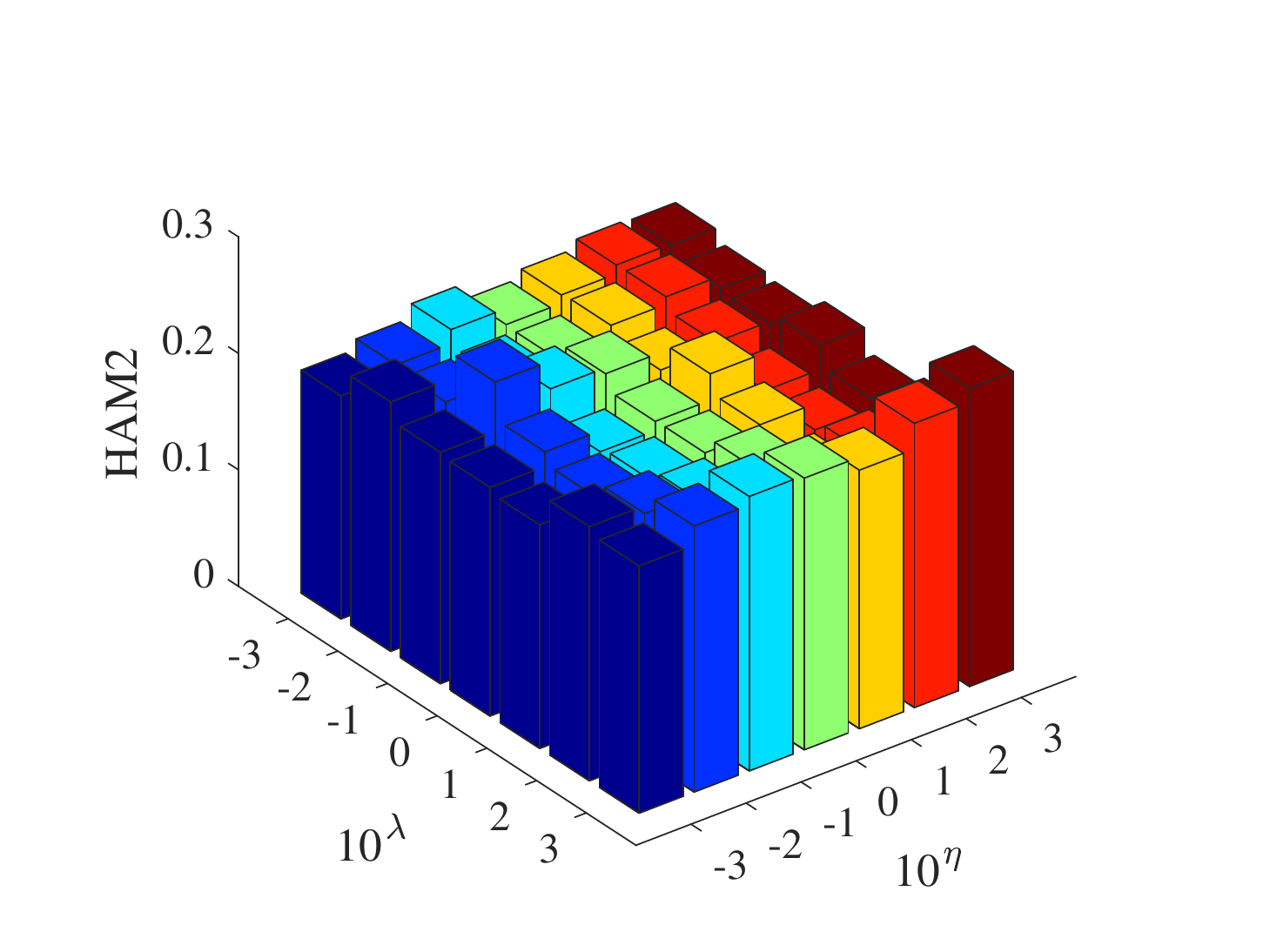}}}			
		\vspace{-1mm}
		\subfigure[Labelme]{\scalebox{0.6} {\includegraphics{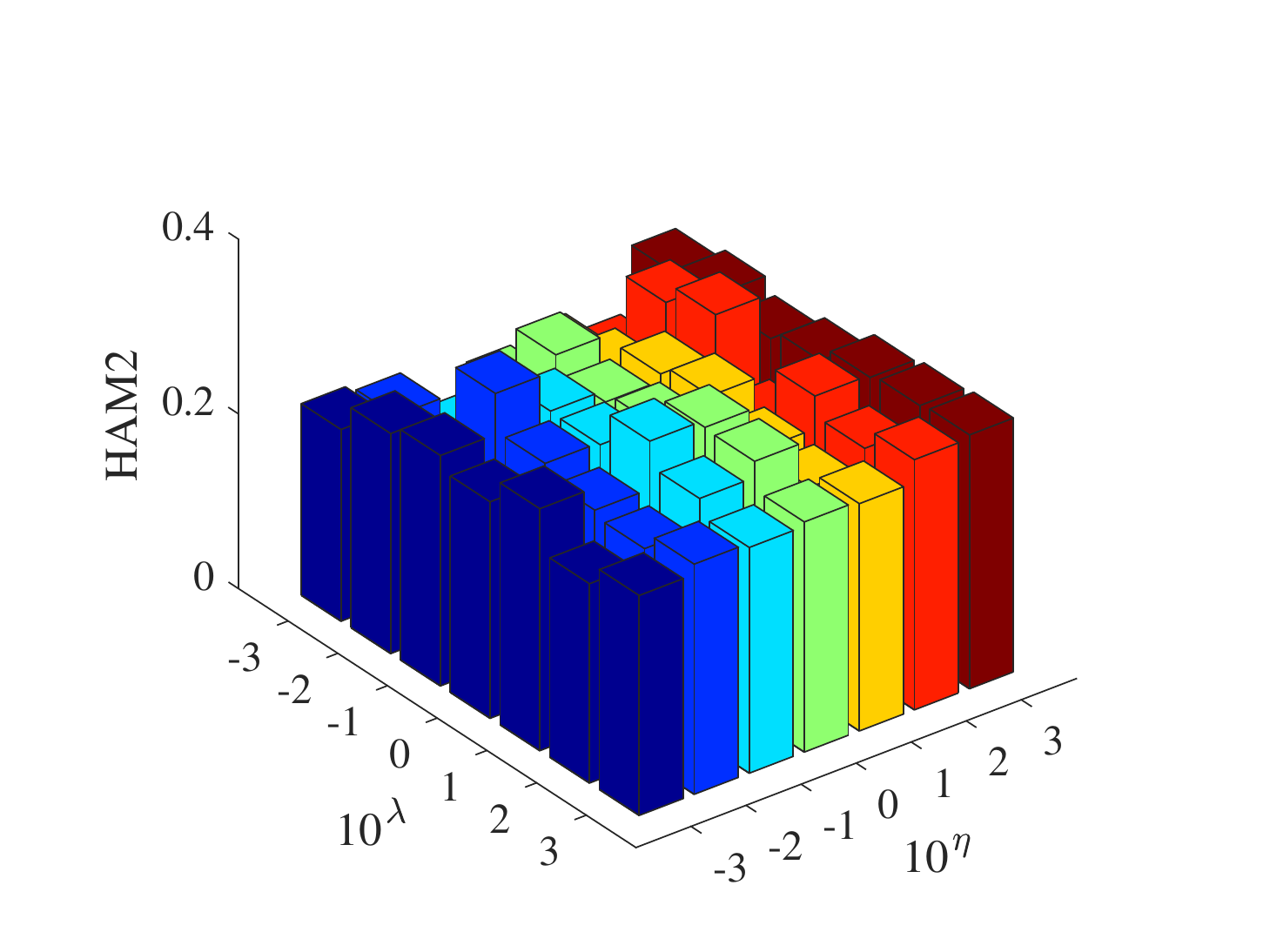}}}
		\vspace{-1mm}		
		\subfigure[Place]{\scalebox{0.6} {\includegraphics{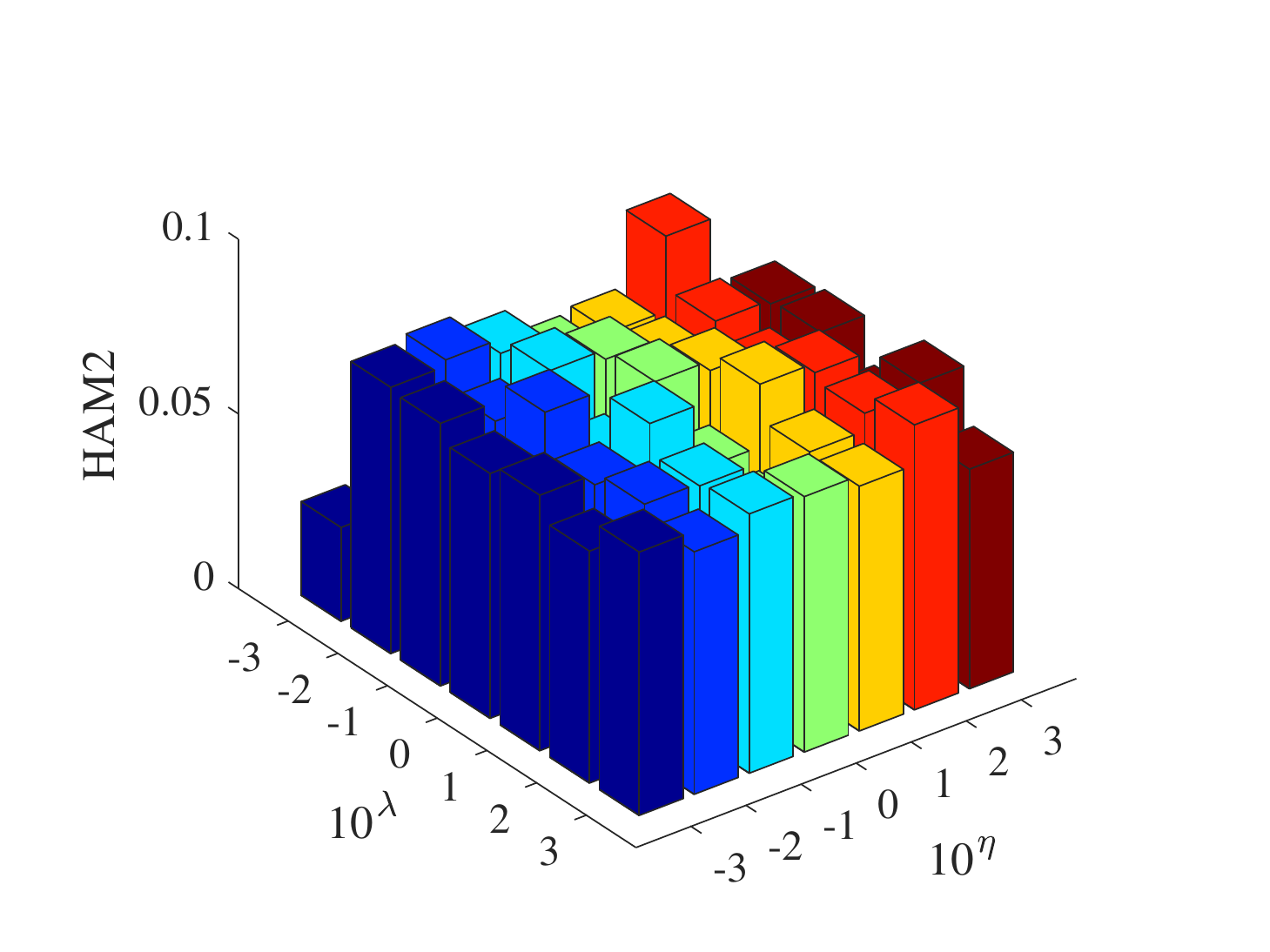}}}		
		\caption{The HAM2 results of our proposed method with different parameters' setting \ie $\lambda$ and $\eta$}
		\label{fig5}
	\end{center}
\end{figure*}

\begin{figure*}[!ht]
	\begin{center}
		\vspace{-1mm}
		\subfigure[Mnist]{\scalebox{0.3}{\includegraphics{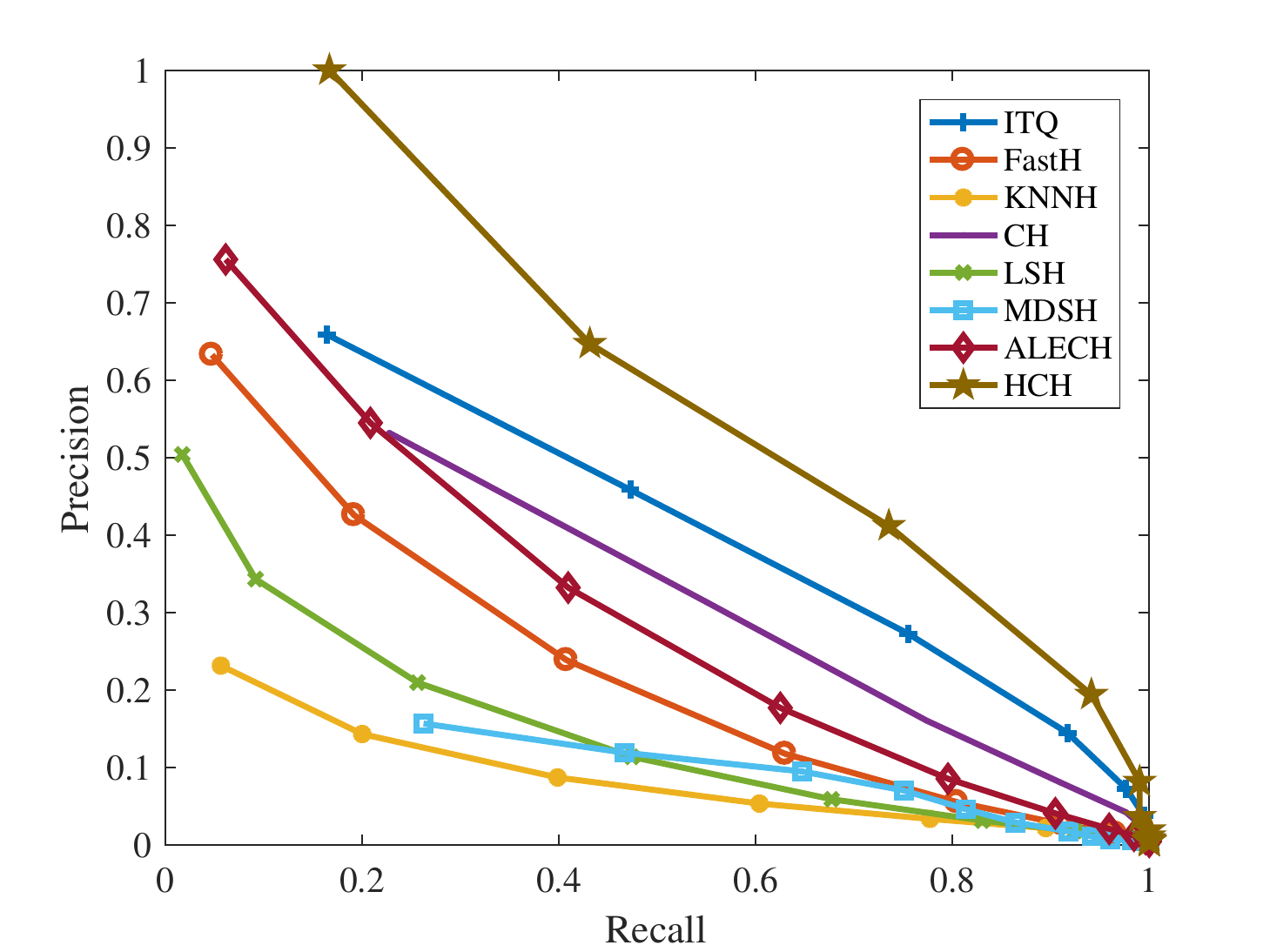}}}	
		\vspace{-1mm}
		\subfigure[Cifar]{\scalebox{0.3} {\includegraphics{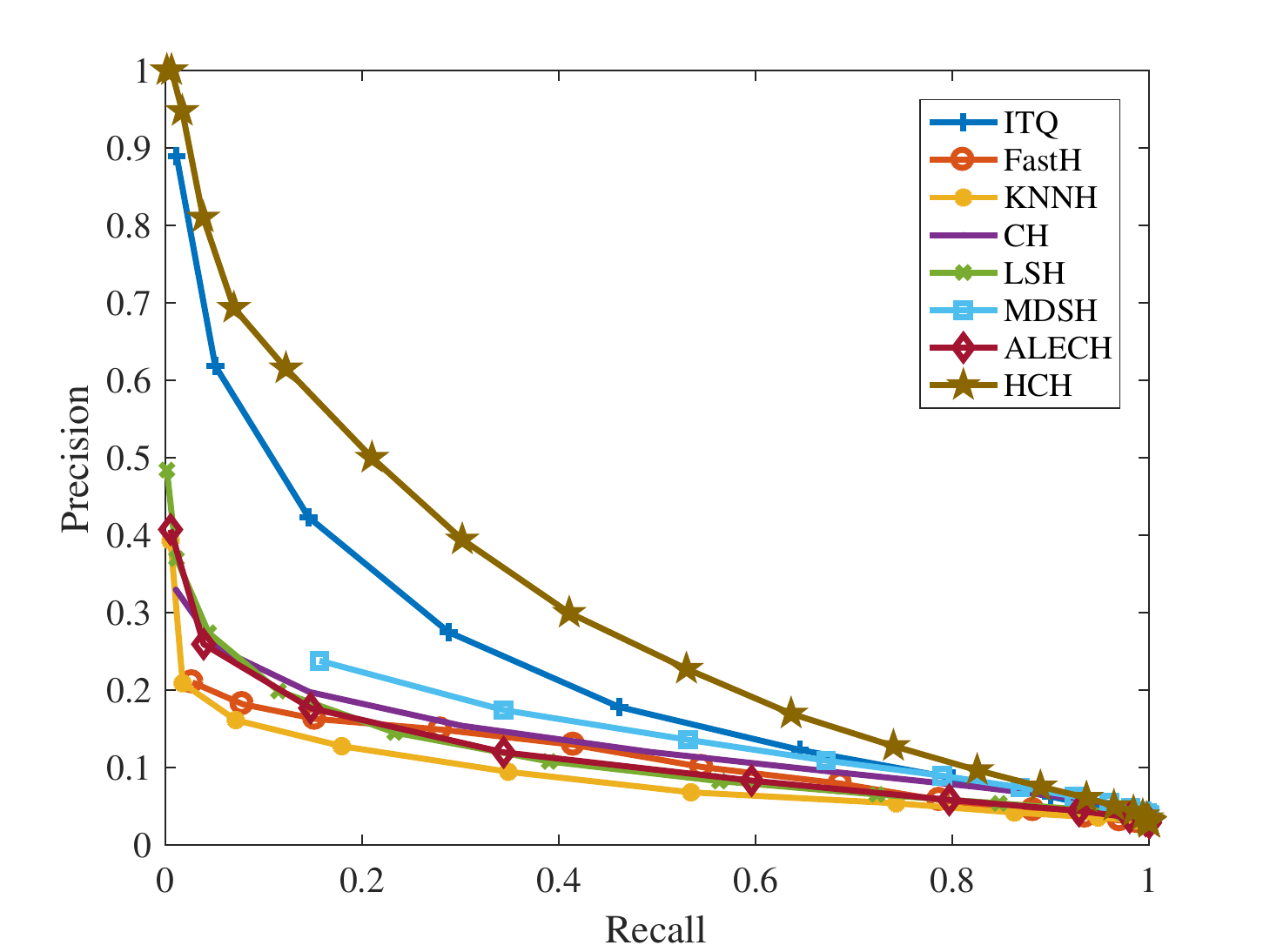}}}			
		\vspace{-1mm}
		\subfigure[Labelme]{\scalebox{0.3} {\includegraphics{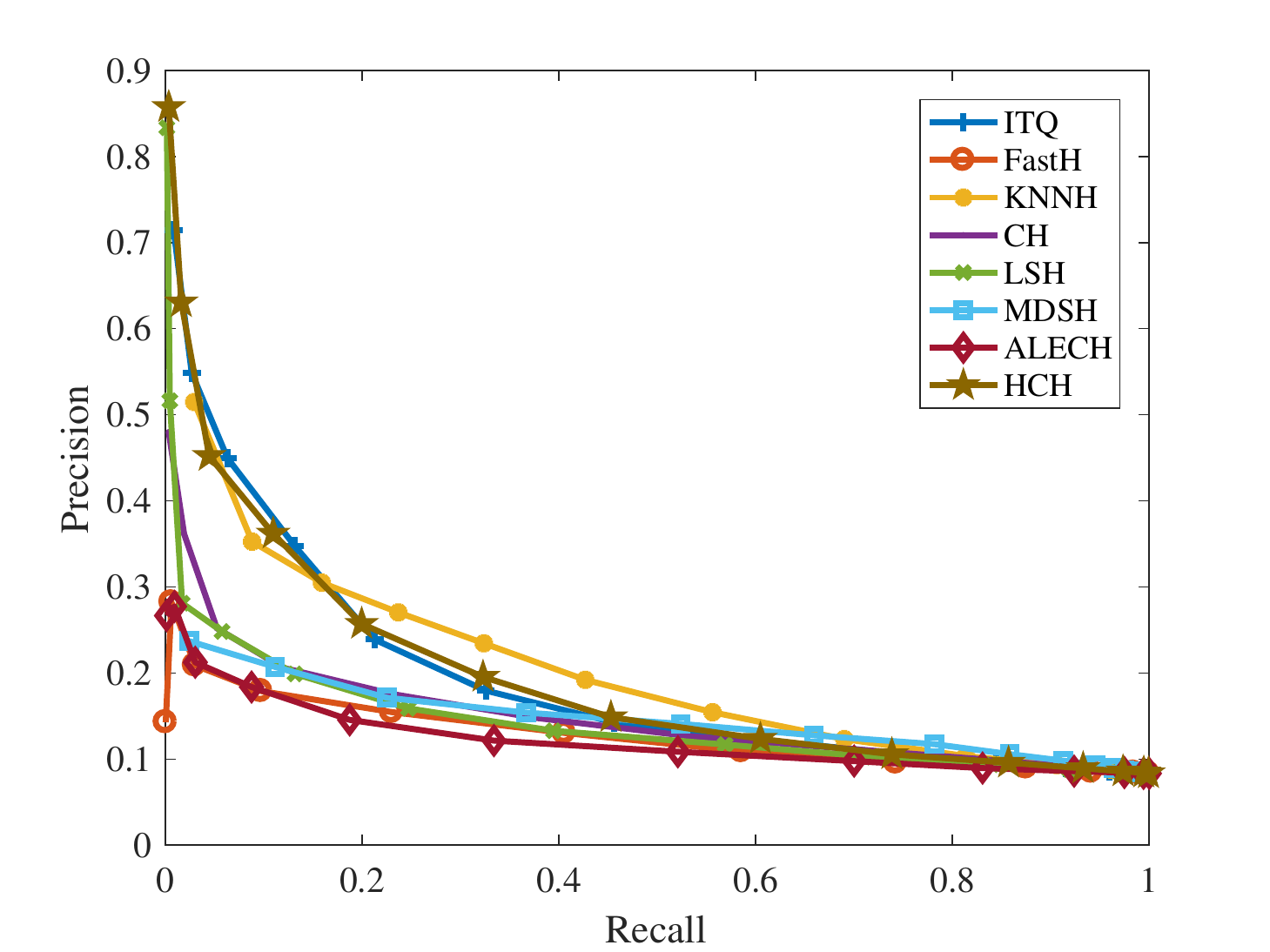}}}
		\vspace{-1mm}		
		\subfigure[Place]{\scalebox{0.3} {\includegraphics{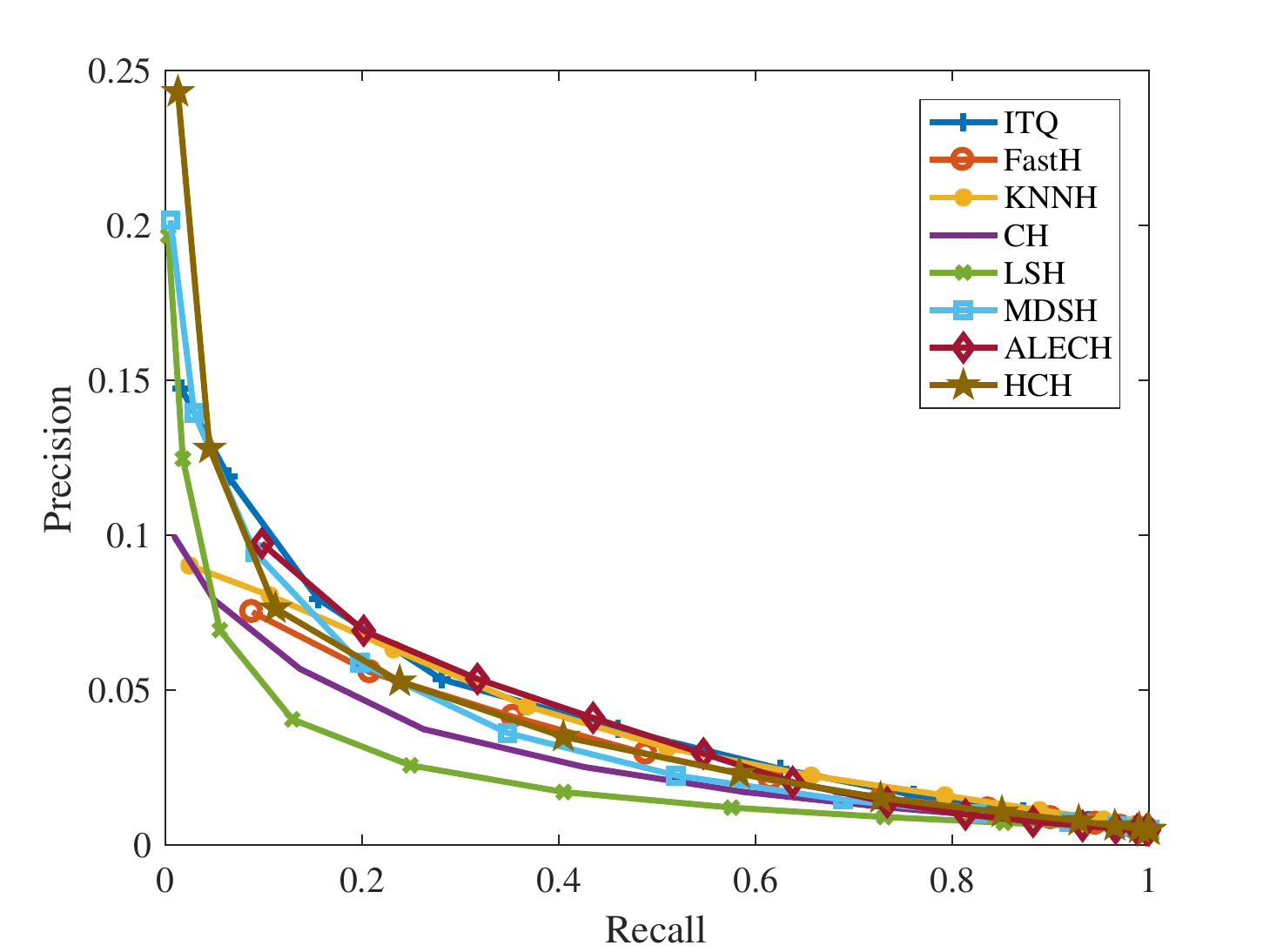}}}		
		\caption{Precision-recall curves of all hashing methods, on four data sets at 16 hash bits}
		\label{fig6}
	\end{center}
\end{figure*}

\begin{figure*}[!ht]
	\begin{center}
		\vspace{-1mm}
		\subfigure[Mnist]{\scalebox{0.3}{\includegraphics{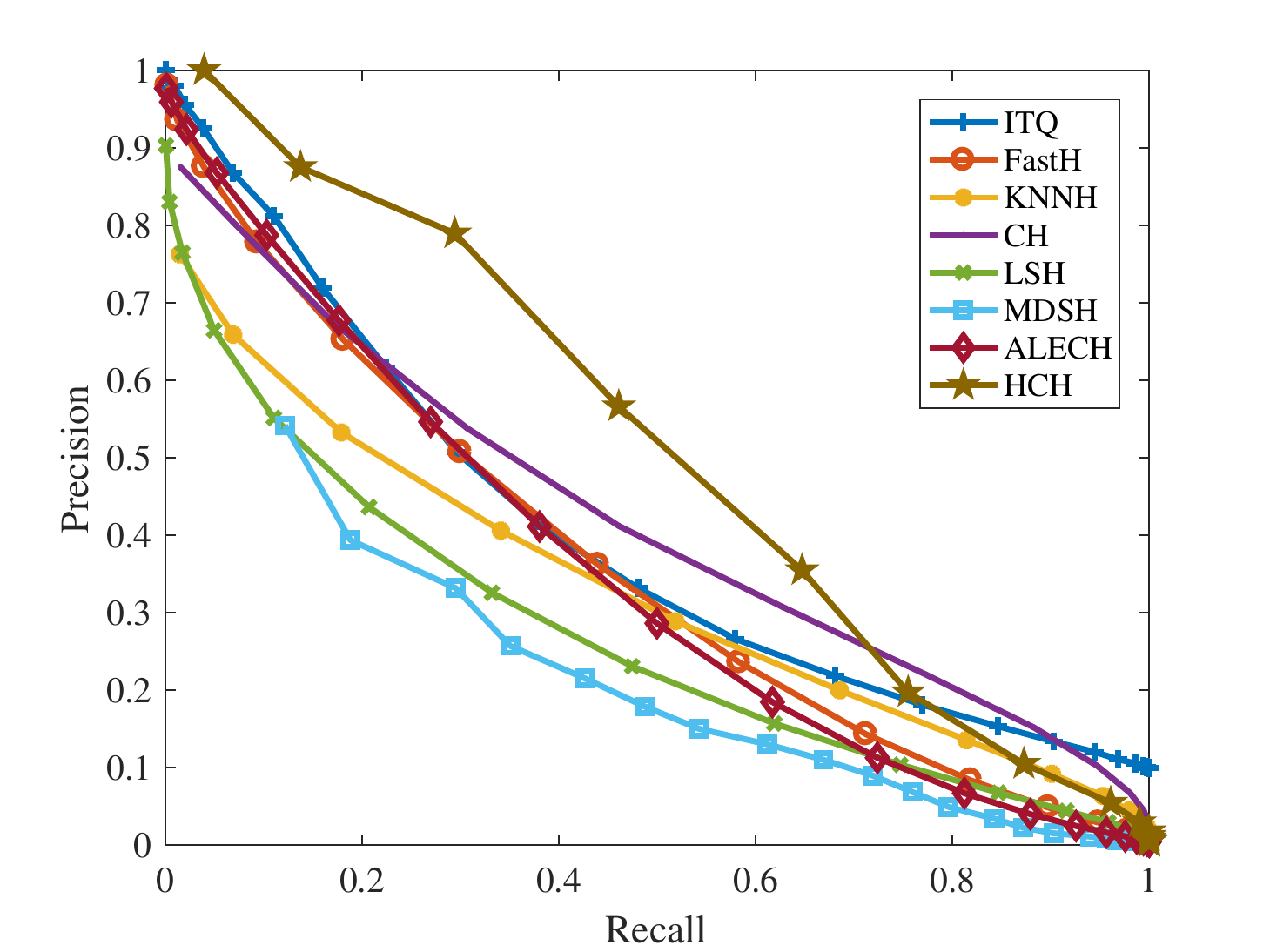}}}	
		\vspace{-1mm}
		\subfigure[Cifar]{\scalebox{0.3} {\includegraphics{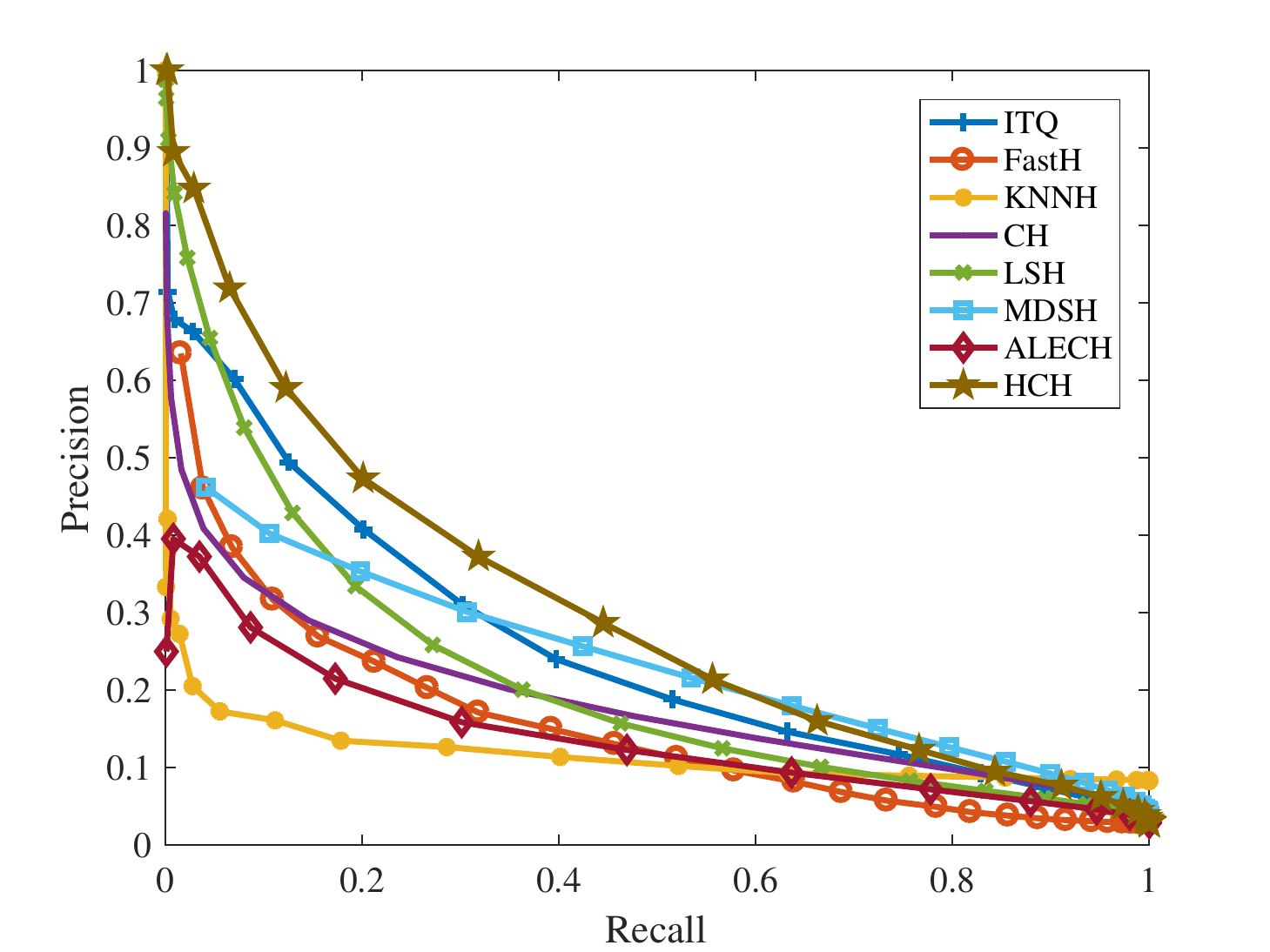}}}			
		\vspace{-1mm}
		\subfigure[Labelme]{\scalebox{0.3} {\includegraphics{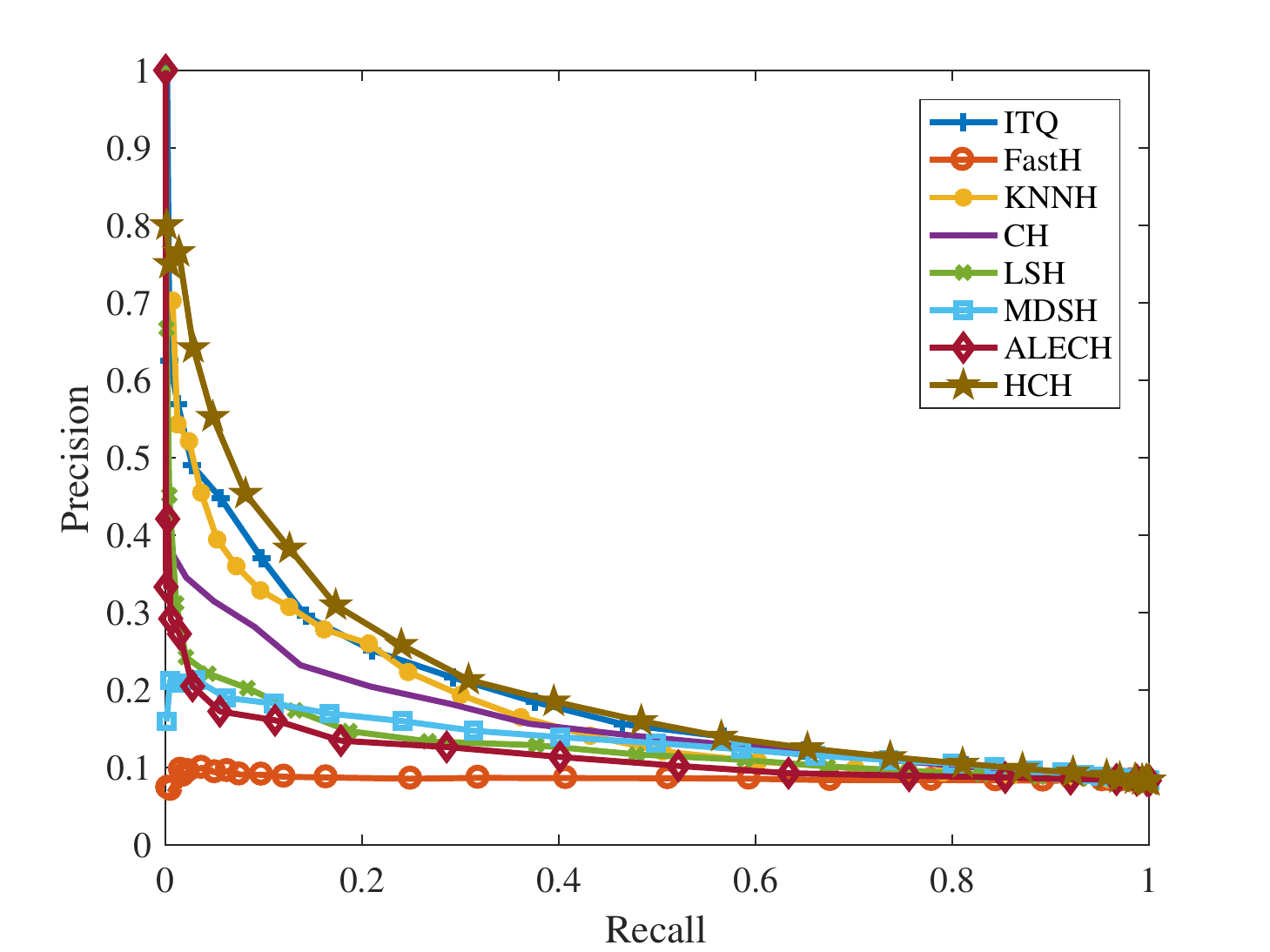}}}
		\vspace{-1mm}		
		\subfigure[Place]{\scalebox{0.3} {\includegraphics{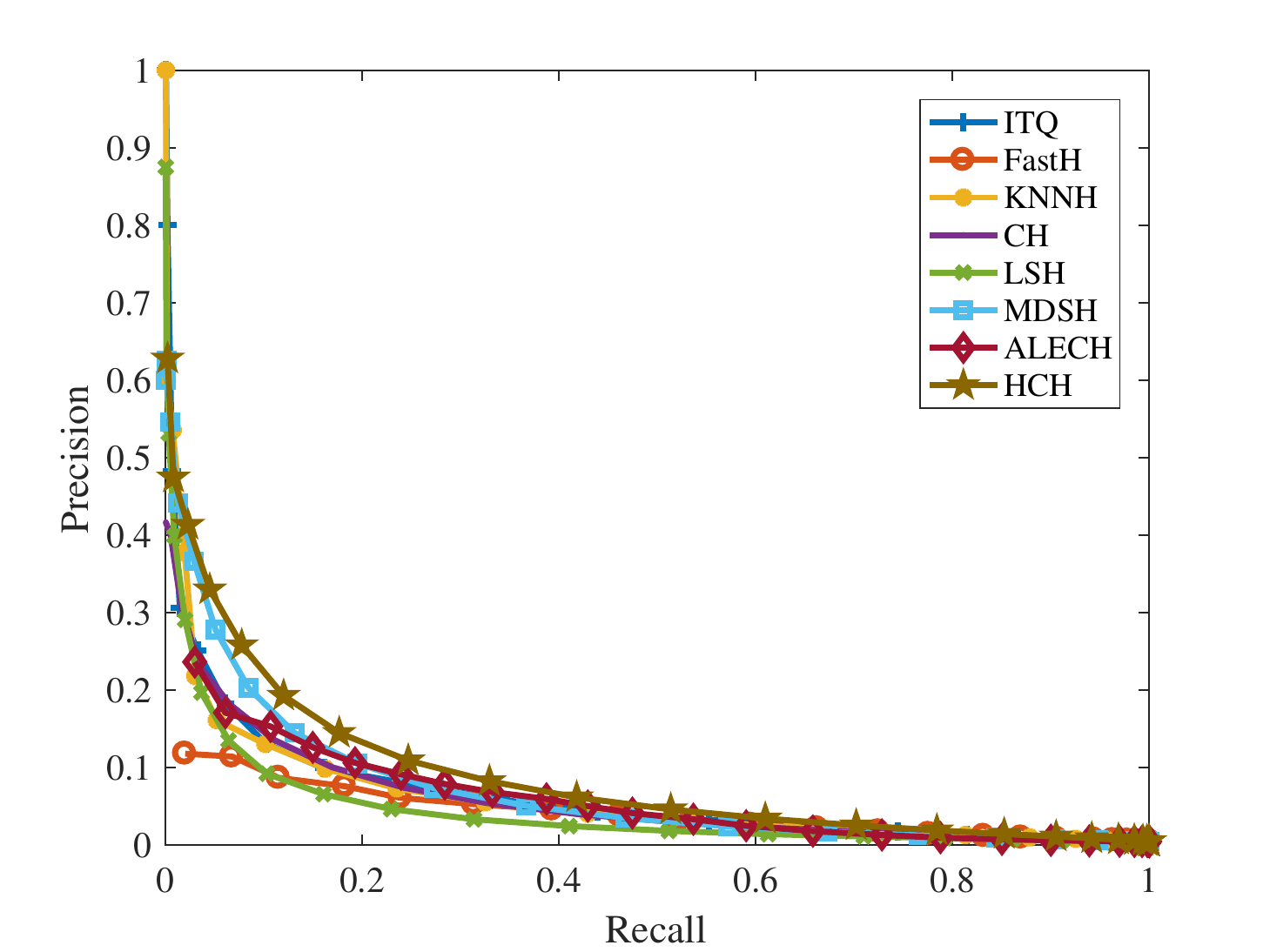}}}		
		\caption{Precision-recall curves of all hashing methods, on four data sets at 32 hash bits}
		\label{fig7}
	\end{center}
\end{figure*}

\begin{figure*}[!ht]
	\begin{center}
		\vspace{-1mm}
		\subfigure[Mnist]{\scalebox{0.3}{\includegraphics{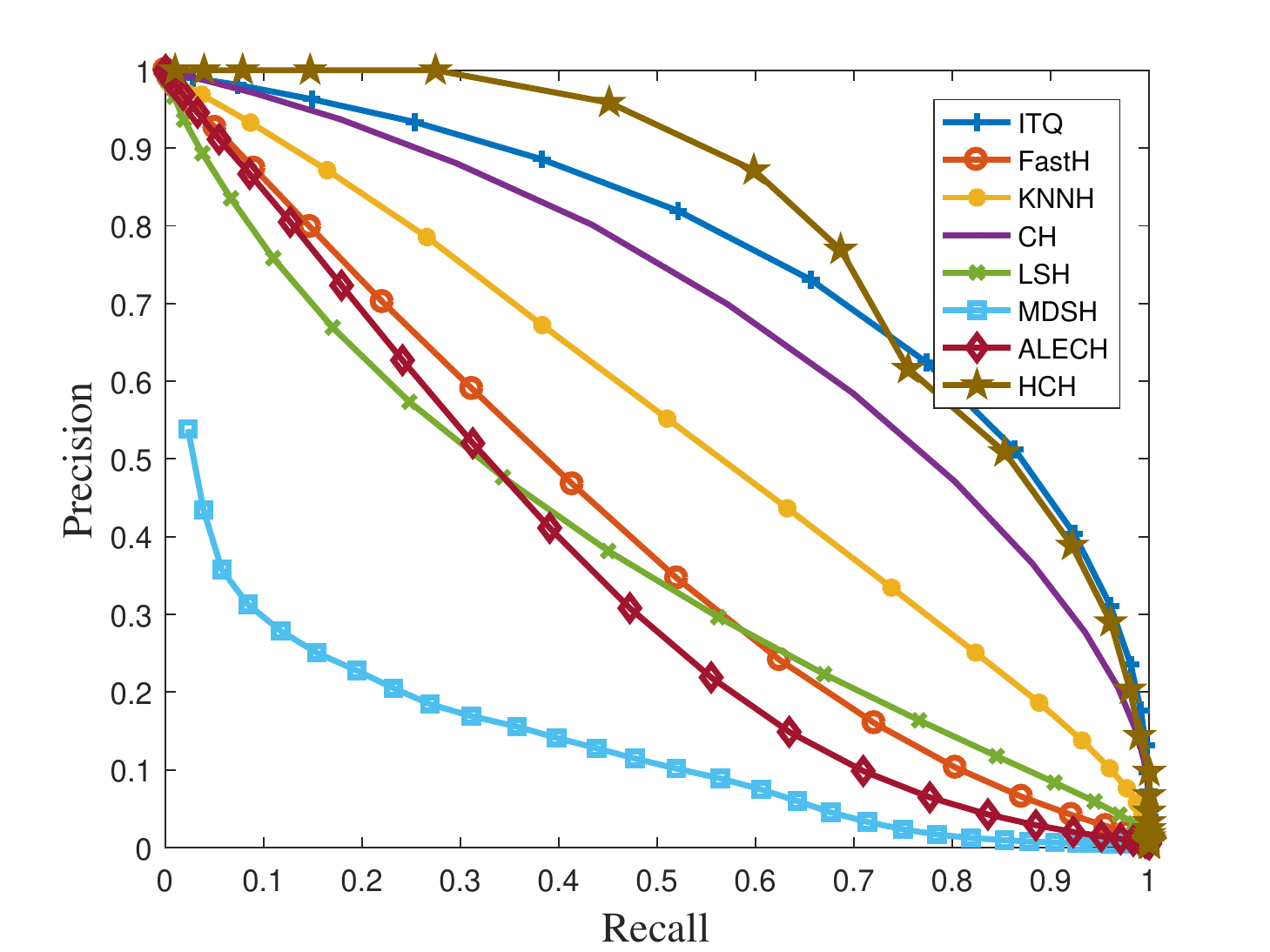}}}	
		\vspace{-1mm}
		\subfigure[Cifar]{\scalebox{0.3} {\includegraphics{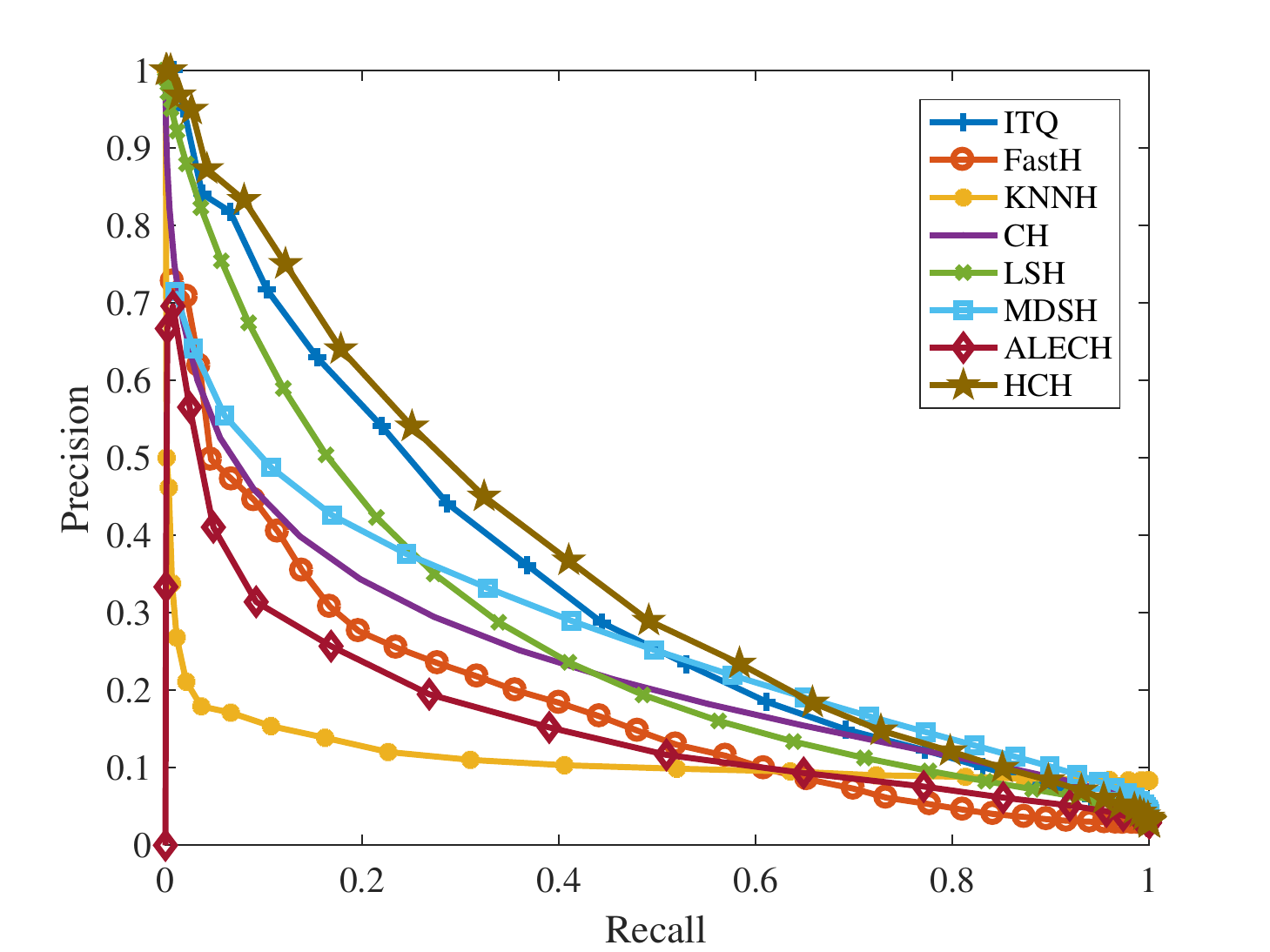}}}			
		\vspace{-1mm}
		\subfigure[Labelme]{\scalebox{0.3} {\includegraphics{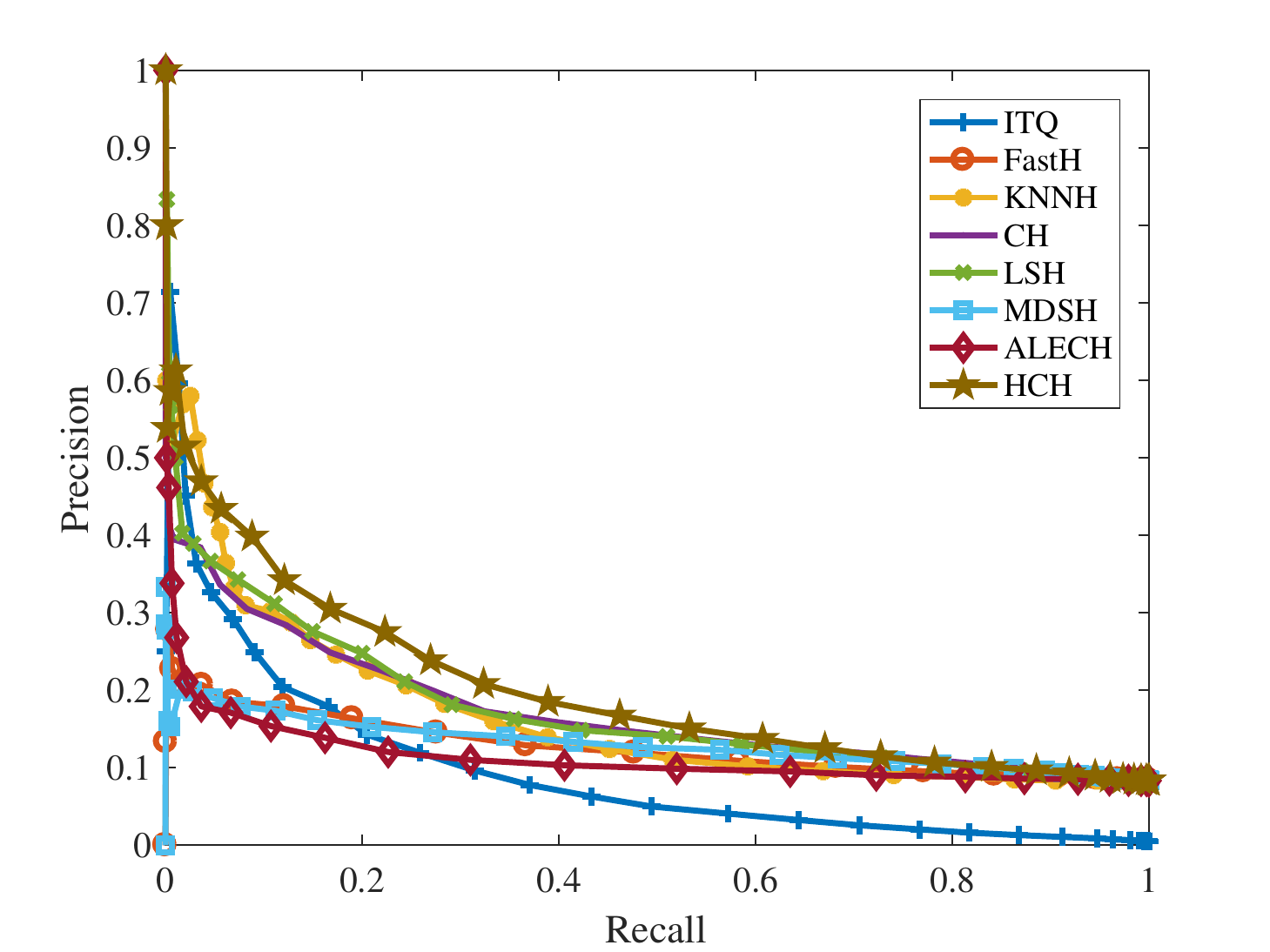}}}
		\vspace{-1mm}		
		\subfigure[Place]{\scalebox{0.3} {\includegraphics{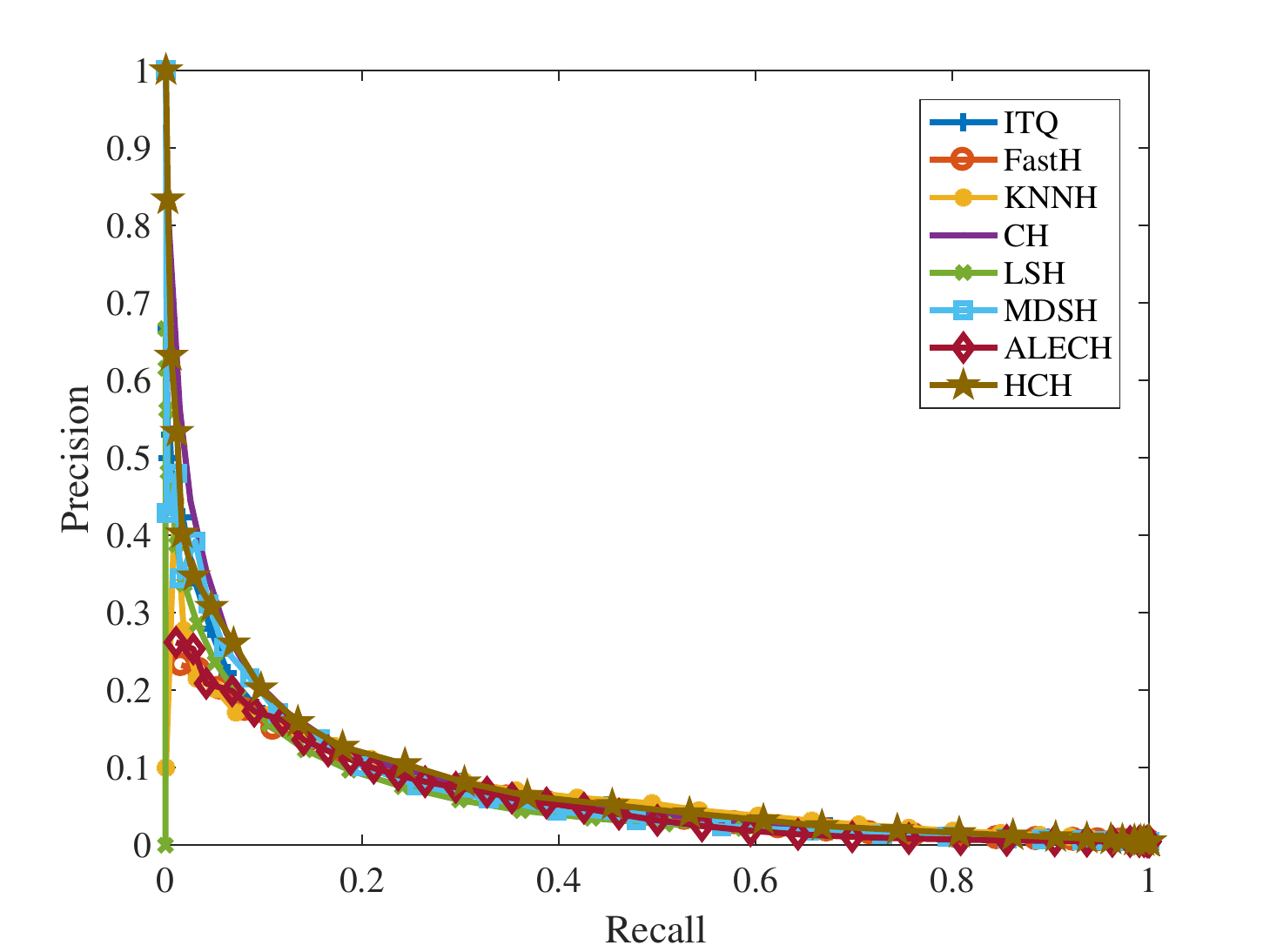}}}		
		\caption{Precision-recall curves of all hashing methods, on four data sets at 48 hash bits}
		\label{fig8}
	\end{center}
\end{figure*}

\begin{figure*}[!ht]
	\begin{center}
		\vspace{-1mm}
		\subfigure[Mnist]{\scalebox{0.3}{\includegraphics{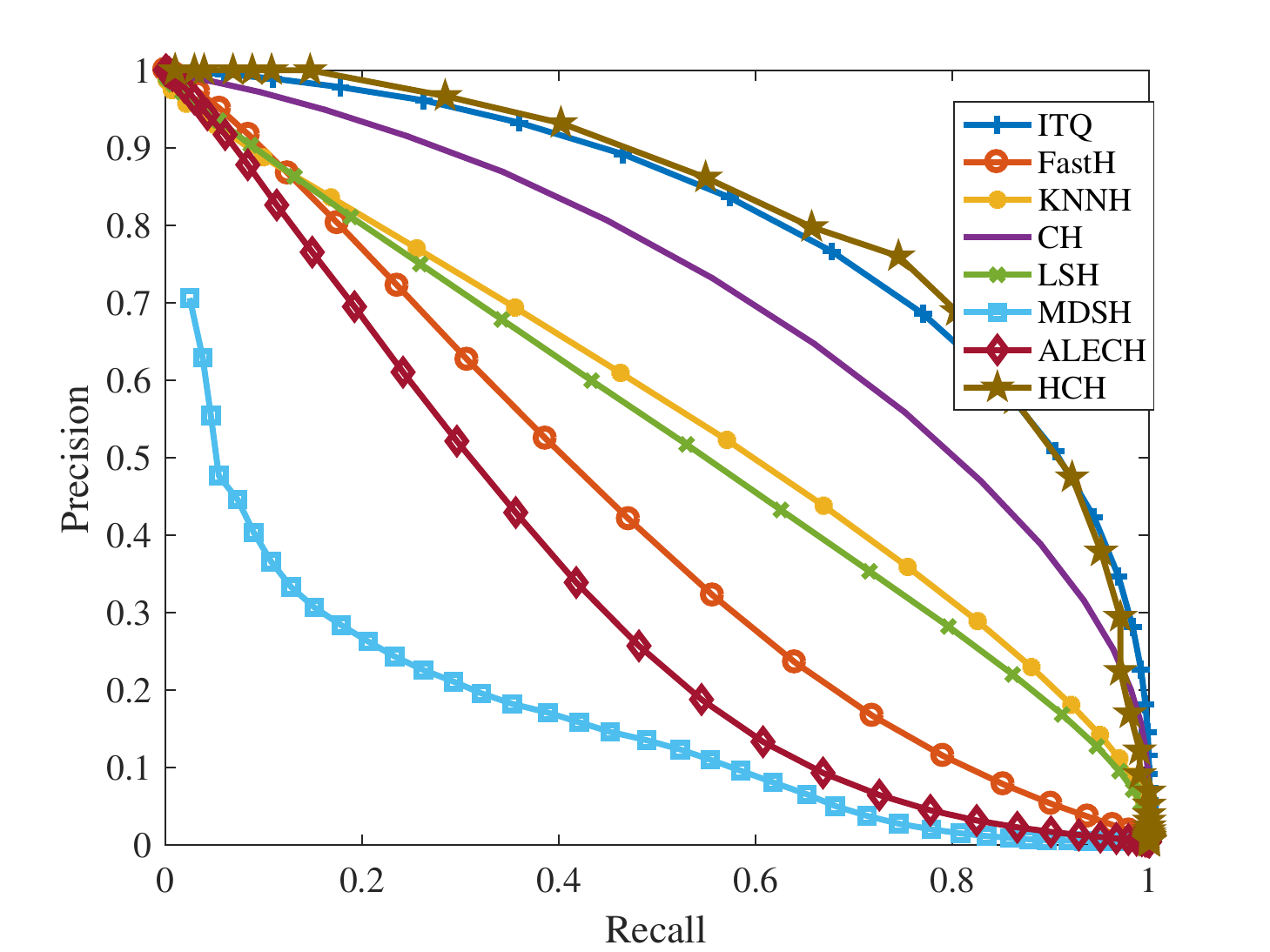}}}	
		\vspace{-1mm}
		\subfigure[Cifar]{\scalebox{0.3} {\includegraphics{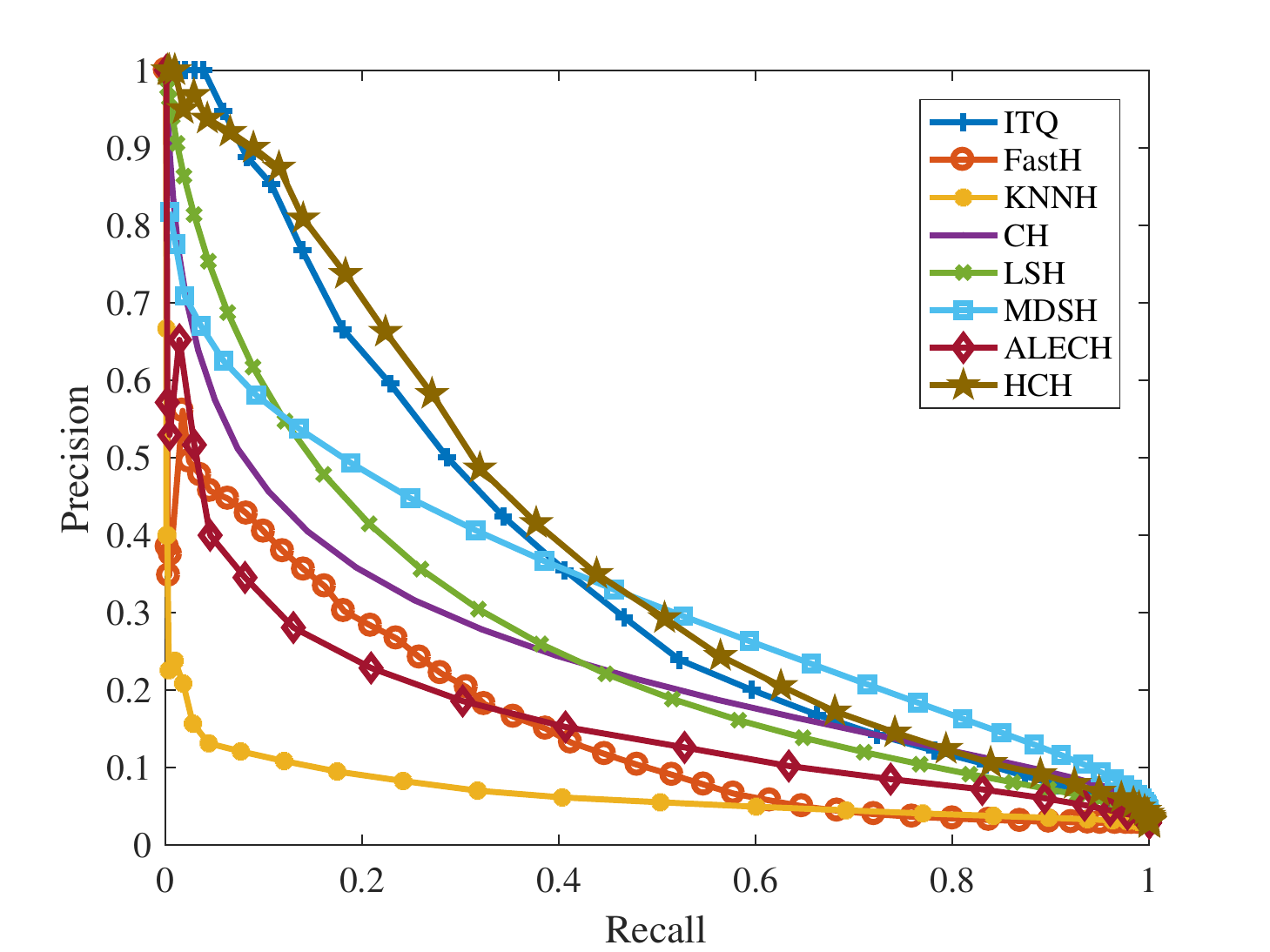}}}			
		\vspace{-1mm}
		\subfigure[Labelme]{\scalebox{0.3} {\includegraphics{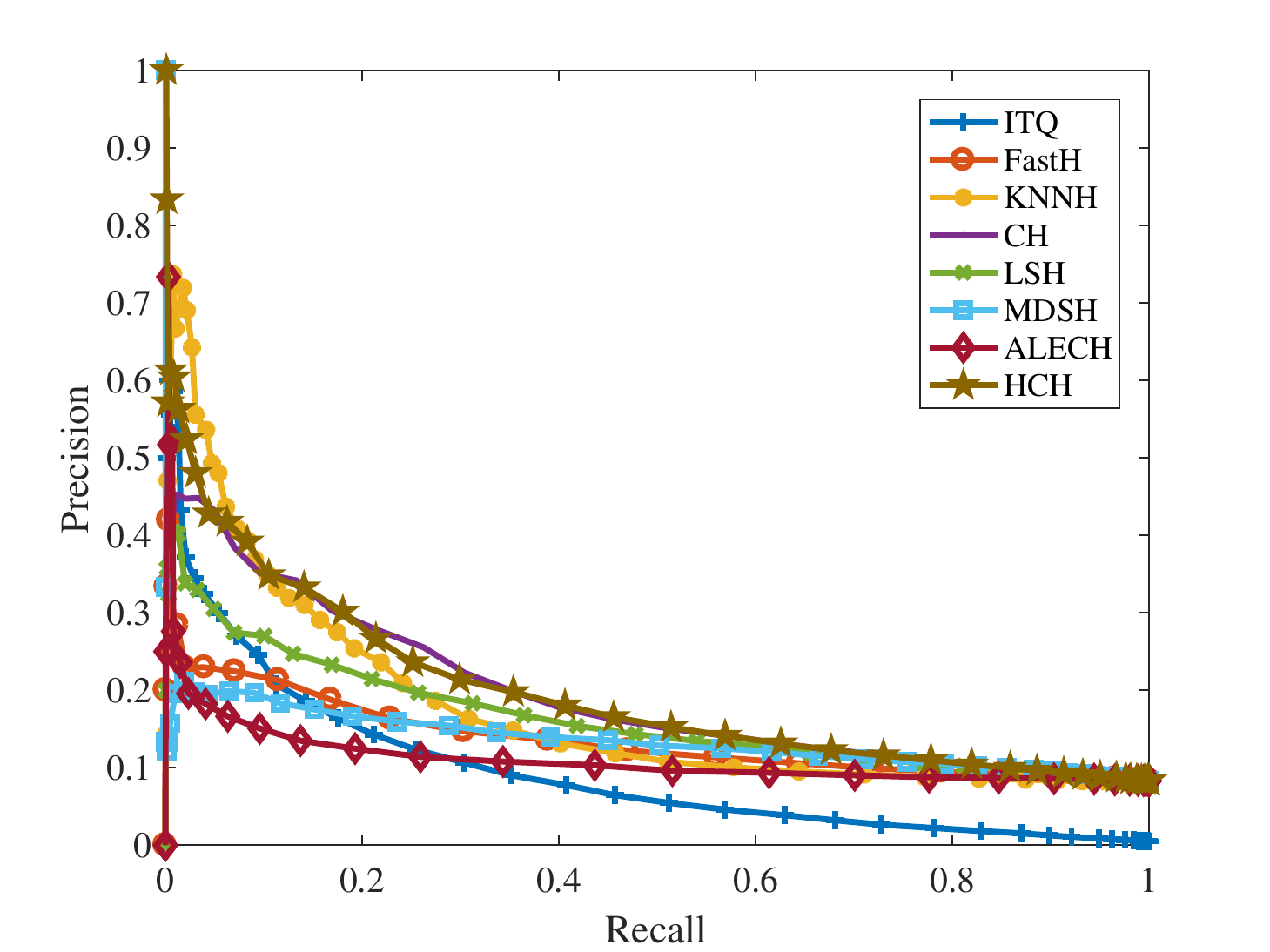}}}
		\vspace{-1mm}		
		\subfigure[Place]{\scalebox{0.3} {\includegraphics{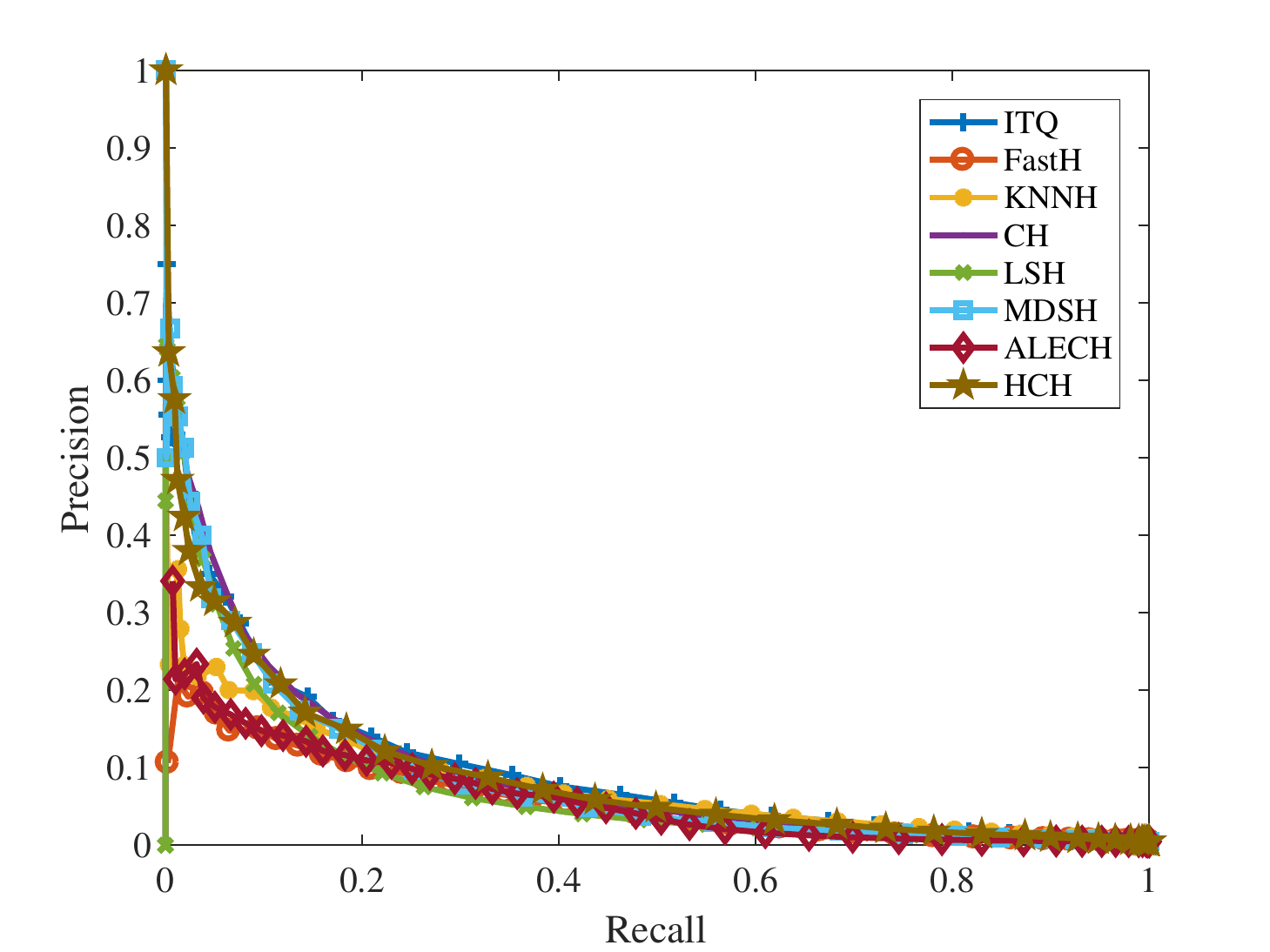}}}		
		\caption{Precision-recall curves of all hashing methods, on four data sets at 64 hash bits}
		\label{fig9}
	\end{center}
\end{figure*}

\begin{figure*}[!ht]
	\begin{center}
		\vspace{-1mm}
		\subfigure[Mnist]{\scalebox{0.6}{\includegraphics{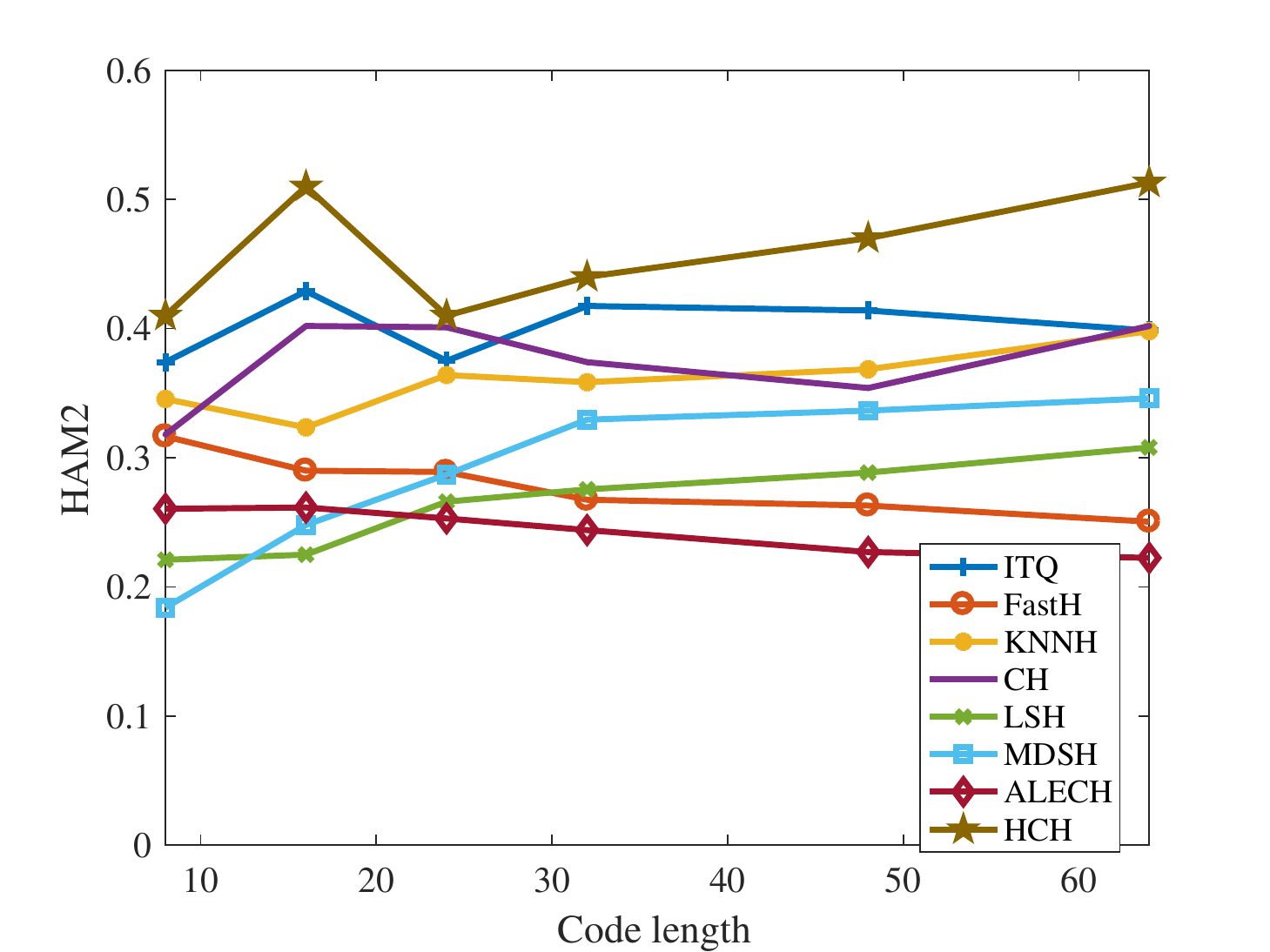}}}	
		\vspace{-1mm}
		\subfigure[Cifar]{\scalebox{0.6} {\includegraphics{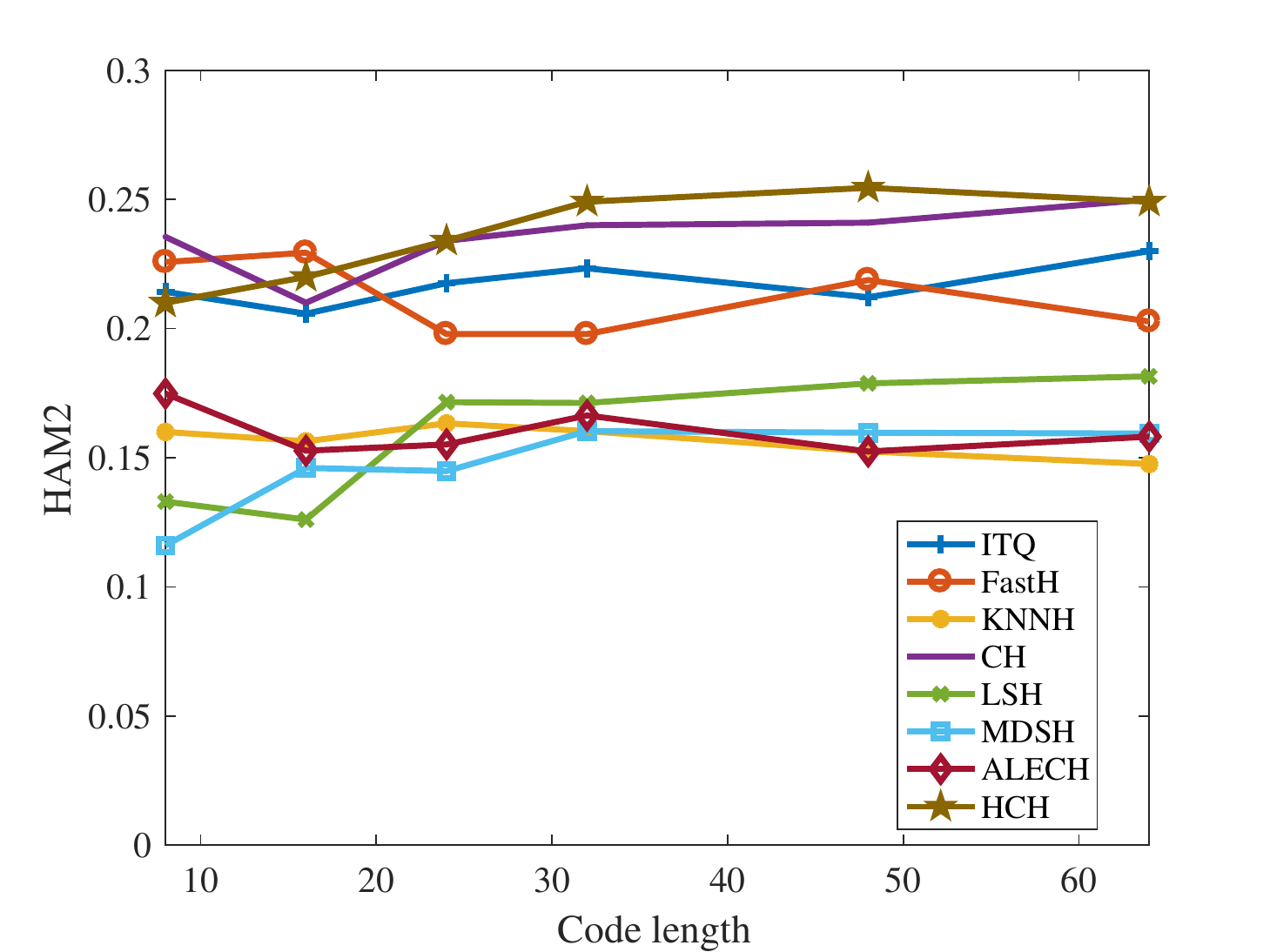}}}			
		\vspace{-1mm}
		\subfigure[Labelme]{\scalebox{0.6} {\includegraphics{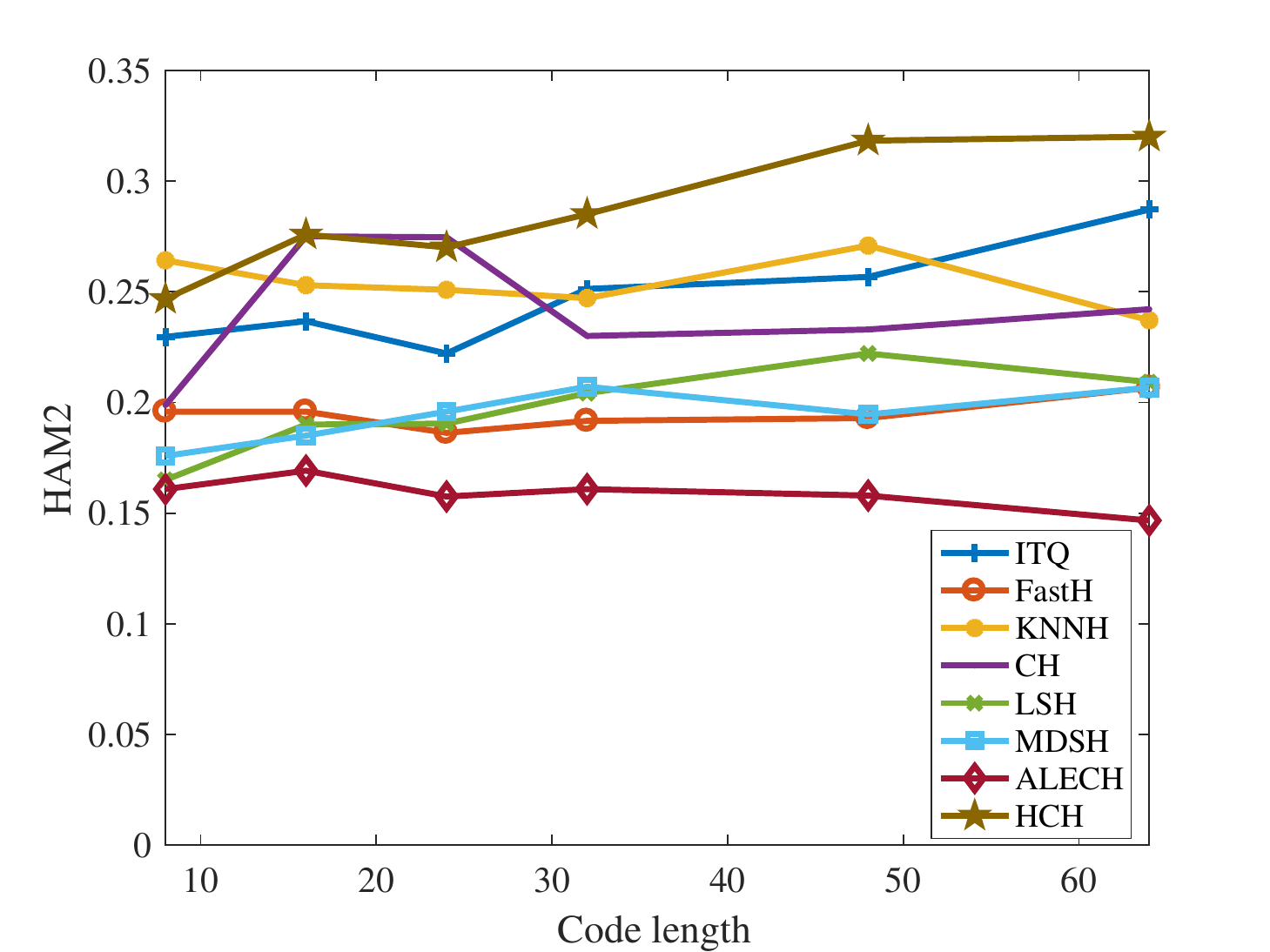}}}
		\vspace{-1mm}		
		\subfigure[Place]{\scalebox{0.6} {\includegraphics{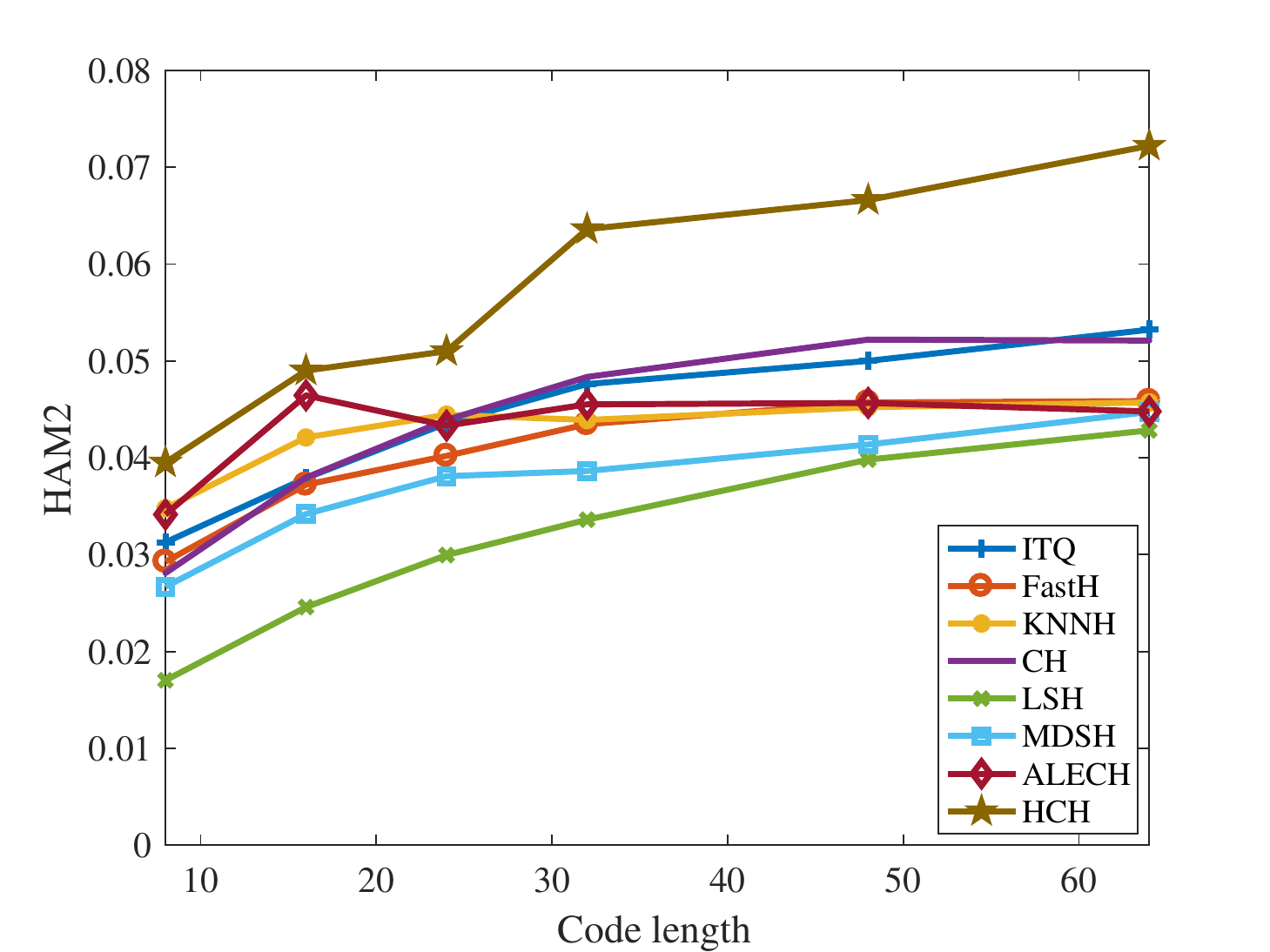}}}		
		\caption{HAM2 results of all hashing methods on four data sets at different number of hash bits, \ie $[8,16,24,32,48,64]$}
		\label{fig10}
	\end{center}
\end{figure*}

	\begin{table*}[!ht]
		\centering
		\caption{ Map results of all hash algorithms on Mnist and Cifar datasets.} 
		\centering
		\vspace{2mm}
		{
			\centering
			{
				\begin{tabular}
					{|p{1.3cm}|p{0.9cm}p{0.9cm}p{0.9cm}p{0.9cm}|p{0.9cm}p{0.9cm}p{0.9cm}p{0.9cm}|} \hline
					\multirow{1}{*}{Method}  &\multicolumn{4}{c|}{Mnist} &\multicolumn{4}{c|}{Cifar-10} \\ \hline
					   &16  &32    &48  &64  &16  &32  &48  &64             \\\hline						
					\multirow{1}{*}{ITQ} &0.3929 &0.4102 &0.4054 &0.3985   &0.1725 &0.1770 &0.1774 &0.1747   \\ \hline
	                 \multirow{1}{*}{FastH} &0.2745 &0.2574 &0.2538 &0.2442   &0.1318 &0.1366 &0.1316 &0.1309   \\ \hline
					\multirow{1}{*}{KNNH} &0.3092 &0.3046 &0.3049 & 0.3096  &0.1570 &0.1852 &0.1734 &0.1717   \\ \hline
					\multirow{1}{*}{CH} &0.4135 &0.3768 &0.3279 &0.3388   &0.1580 &0.1525 &0.1594 & 0.1482  \\ \hline	
					\multirow{1}{*}{LSH} &0.2103 &0.2710 &0.2885 &0.2974   &0.1239 &0.1470 &0.1462 & 0.1520  \\ \hline	
					\multirow{1}{*}{MDSH} &0.3270 &0.3953 &0.3483 &0.3050   &0.1685 &0.1445 &0.1508 & 0.1407  \\ \hline	
					\multirow{1}{*}{ALECH} &0.2419 &0.2225 &0.2133 &0.2067   &0.1374  &0.1273  &0.1246  &0.1202   \\ \hline
					\multirow{1}{*}{HCH} &0.3758 &0.4104 &0.4101 &0.4182   &0.1736 &0.1798 &0.1791 &0.1752   \\ \hline	
				\end{tabular}
			}
		}
		\label{tab3}
	\end{table*}

	\begin{table*}[!ht]
		\centering
		\caption{ Map results of all hash algorithms on Labelme and Place datasets.} 
		\centering
		\vspace{2mm}
		{
			\centering
			{
				\begin{tabular}
					{|p{1.3cm}|p{0.9cm}p{0.9cm}p{0.9cm}p{0.9cm}|p{0.9cm}p{0.9cm}p{0.9cm}p{0.9cm}|} \hline
					\multirow{1}{*}{Method}  &\multicolumn{4}{c|}{Labelme} &\multicolumn{4}{c|}{Place} \\ \hline
					&16  &32    &48  &64  &16  &32  &48  &64             \\\hline						
					\multirow{1}{*}{ITQ} &0.1971 &0.2121 &0.2162 &0.2377   &0.0706 &0.1145 &0.1482 &0.1654   \\ \hline
					\multirow{1}{*}{FastH} &0.1942 &0.1894 &0.1949 &0.1950   &0.0854 &0.1131 &0.1200 & 0.1400  \\ \hline
					\multirow{1}{*}{KNNH} &0.2125 &0.1957 &0.2048 &0.1831   &0.1029 &0.1256 &0.1376 &0.1385   \\ \hline
					\multirow{1}{*}{CH} &0.1904 &0.1745 &0.1775 &0.1731   &0.1022 &0.1211 &0.1381 &0.1312   \\ \hline	
					\multirow{1}{*}{LSH} &0.1582 &0.1659 &0.1732 &0.1819   &0.0528 &0.0860 &0.1135 &0.1218   \\ \hline	
					\multirow{1}{*}{MDSH} &0.1779 &0.1772 &0.1714 & 0.1816  &0.0713 &0.1037 &0.1226 & 0.1427  \\ \hline	
					\multirow{1}{*}{ALECH} &0.1389 &0.1275 &0.1251 & 0.1225  &0.0524  &0.0906  &0.1114  & 0.1285  \\ \hline
					\multirow{1}{*}{HCH} &0.1984 &0.2192 &0.2187 &0.2401   &0.1154 &0.1283 &0.1529 &0.1863   \\ \hline	
				\end{tabular}
			}
		}
		\label{tab4}
	\end{table*}

\subsection{Experimental setting}

In this paper, we mainly carry out two part experiments: 1 The convergence and parameter sensitivity of the proposed algorithm are analyzed. 2. The performance comparison between the proposed algorithm and other outstanding hash algorithms. In the second part of the experiment, we compared the precision recall performance of all algorithms on four data sets, The mean average accuracy ($MAP$) of different bits and the average precision of Hamming radius 2 corresponding to different bits. In addition, in the proposed algorithm, the value range of parameters $\alpha$, $\beta$, $\lambda$ and $\eta$ is: $[{10^{ - 3}},{10^{ - 2}},10,1,10,{10^2},{10^3}]$. $MAP$ is the mean value of the average precision ($AP$) returned by the retrieval of all query samples. The calculation formula of $AP$ is as follows:
\begin{eqnarray}
\label{eq47}
\begin{array}{l}
AP = \frac{1}{l}\sum\limits_{r = 1}^K {Precision(r)\sigma (r)} 
\end{array}
\end{eqnarray}
where $l$ is the number of samples of real relevant nearest neighbors, and $Precision(r)$ is the accuracy of the training data retrieved by top r. If the r-th instance is related to the sample of the query, then $\sigma (r) = 1$, otherwise $\sigma (r) = 0$. $MAP$ is defined as follows:
\begin{eqnarray}
\label{eq48}
\begin{array}{l}
MAP = \frac{1}{m}\sum\limits_{r = 1}^m {AP(i)} 
\end{array}
\end{eqnarray}
where $m$ represents the number of query samples, and $AP(i)$ is the average accuracy of the \emph i-th sample.

\subsection{Convergence and parameter sensitivity analysis of the algorithm}

As shown in Fig. \ref{fig3}, it shows the objective function value of each iteration of the proposed algorithm on four data sets. We can find that the proposed algorithm converges within 5 iterations on four data sets. This shows that the proposed algorithm has fast convergence speed and can greatly reduce the time cost of training. The proposed objective function has four parameters, namely, $\alpha$, $\beta$, $\lambda$ and $\eta$, as shown in Eq. (\ref{eq18}). We have done the HAM2 result experiment of the proposed method under different parameter settings, as shown in Fig. \ref{fig4} and Fig. \ref{fig5}. In Fig. \ref{fig4}, we set the values of $\lambda$ and $\eta$ to 1, and then adjust the values of $\alpha$ and $\beta$ to conduct the experiment. From Fig. \ref{fig4}, we can see that on the data sets Mnist and Cifar, when $\alpha = 10^2$ or $10^3$, the proposed algorithm achieves the best effect. On the dataset of Labelme and Place, when $\alpha =10^{-3}$ and $\beta = 10^3$, the proposed algorithm has the best effect. In addition, we can also find that different $\alpha$ and $\beta$ values will affect the effect of the proposed algorithm. Because $\alpha$ controls the similarity between hash codes in the established hyper-class, \ie $\alpha \sum\nolimits_i^{{n_1}} {\left\| {\mathbf s_1^{(i)}} \right\|_2^2} ,\alpha \sum\nolimits_i^{{n_2}} {\left\| {\mathbf s_2^{(i)}} \right\|_2^2} , \ldots ,\alpha \sum\nolimits_i^{{n_c}} {\left\| {\mathbf s_c^{(i)}} \right\|_2^2} $, and ${\mathbf s_1},{\mathbf s_2}, \ldots ,{\mathbf s_c}$ stores the similarity relationship between hash codes in each hyper-class. $\beta$ controls the similarity of hash codes between hyper-classes, \ie $\beta \sum\nolimits_{i = 1}^{{n_1}} {\left\| {{\mathbf I_{{}^\neg 1}}\mathbf H_1^{(i)} - {\mathbf V_{{}^\neg 1}}} \right\|_2^2} ,\beta \sum\nolimits_{i = 1}^{{n_2}} {\left\| {{\mathbf I_{{}^\neg 2}}\mathbf H_2^{(i)} - {\mathbf V_{{}^\neg 2}}} \right\|_2^2} , \ldots ,\\\beta \sum\nolimits_{i = 1}^{{n_c}} {\left\| {{\mathbf I_{{}^\neg c}}\mathbf H_c^{(i)} - {\mathbf V_{{}^\neg c}}} \right\|_2^2} $, which indicates the similarity between each hyper-class and other hyper-classes. Therefore, we need to carefully adjust the values of the parameters $\alpha$ and $\beta$. 

In Fig. \ref{fig5}, we set the values of $\alpha$ and $\beta$ to 1, and then adjust the values of $\lambda$ and $\eta$ to conduct the experiment. From Fig. \ref{fig5}, we can see that the optimal parameters are different on each data set. For example, on Mnist dataset, when $\lambda = 10^0$ and $\eta = 10^2$, the proposed algorithm achieves the best results. Similarly, when $\lambda  = [{10^3},{10^{ - 1}},{10^{ - 3}}]$ and  $\eta  = [{10^3},{10^2}]$, the proposed algorithm achieves the best results on datasets Cifar, Labelme and Place respectively. Because they affect the values of $\lambda \left\| {\mathbf H - {\mathbf X^T}{\mathbf U^T}} \right\|_F^2$ and $\eta \left\| \mathbf U \right\|_2^2$ respectively, these two terms control the value of the relationship matrix between the hash code and the sample. Therefore, we also need to carefully adjust the values of the parameters $\lambda$ and $\eta$.

\subsection{Performance comparison with other hash algorithms}

We show the precision recall curves of all algorithms on four data sets under different bit numbers (\ie 16, 32, 48, 64), as shown in Figs. \ref{fig6}-\ref{fig9}. In addition, we also show the HAM2 results of all algorithms under different bit numbers and $MAP$ results under different bits, as shown in Fig. \ref{fig10}, table \ref{tab3} and table \ref{tab4}.

From figs. \ref{fig6}-\ref{fig9}, we can find that the proposed algorithm achieves the best performance. The reason is that the proposed algorithm is a hash algorithm based on hyper-class representation, which can make the hash code similarity within each hyper-class as high as possible and the hash code similarity between hyper-classes as low as possible. So it can learn more suitable hash codes. Specifically, on Mnist and Cifar datasets, the proposed HCH algorithm achieves the best results on each number of bits, compared with other comparison algorithms. When the number of bits is only 16, the proposed algorithm does not achieve the best performance on the dataset Labelme and Place. The reason is that different datasets have different characteristics. In the datasets Labelme and Place, 16 bits are not enough to represent the similarity of data within the hyper-class and between hyper-classes. However, when the hash code is larger than 16 bits, \ie when the hash code is 32 bits, the proposed HCH algorithm still achieves the best results.

From Fig. \ref{fig10}, we can see that the proposed HCH algorithm achieves the best performance in the HAM2 results under different bit numbers. Specifically, on Mnist and Place datasets, when the bit number is 64, the effect of HCH algorithm is most obvious on HAM2. On data sets Cifar and Labelme, when the number of bits is 48, the effect of HCH algorithm is most obvious on HAM2. Therefore, the proposed HCH algorithm has the best hash learning ability compared with other comparison algorithms. 

From table \ref{tab3} and table \ref{tab4}, we can see that the proposed algorithm achieves the best results in the $MAP$ results. Specifically, on Mnist data, when the number of bits is 64, HCH is improved by 1.97\% compared with ITQ algorithm. Compared with CH algorithm, HCH algorithm is improved by 7.94\%. On the Place dataset, HCH is improved by 2.09\% compared with ITQ algorithm. Compared with CH algorithm, HCH algorithm is improved by 5.11\%. The reason is that the proposed HCH algorithm is different from the previous data independent hash algorithm (LSH) and data dependent hash algorithm (FastH, ITQ, KNNH, CH, etc.). HCH depends on the hyper-class representation of data. It considers the potential class information in data from the level of data characteristics, \ie hyper-class information. According to the hyper-class information, we can further construct a hash code that can better represent the relationship between data features, so as to improve the hash learning ability of the algorithm.

\section{Conclusion}  \label{conclusion}

In this paper, we have proposed a hash algorithm based on hyper-class representation. Specifically, we first calculate the relationship between each feature and all other features, and apply a weight to each feature, so as to select the feature that is most likely to be used as the decision feature. Then we use the selected decision features to construct the hyper-class representation of data. Finally, we propose a new hash algorithm based on the principle of ``the similarity of hash codes withinhyper-class is as high as possible, and the similarity between hash codes of data between hyper-classes is as low as possible". In the experiment, the proposed algorithm shows better performance on four data sets, compared with other comparison algorithms.

The proposed hash algorithm  is based on the hyper-class representation of data. Therefore, in the future work, we plan to apply the proposed representation of data to other fields. \ie different data representations are proposed for different data mining algorithms.

\section*{Acknowledgment}
This work has been supported in part by the Natural Science Foundation of China under grant 61836016.

\ifCLASSOPTIONcaptionsoff
  \newpage
\fi

	\bibliography{refs}{}
	\bibliographystyle{IEEEtran}

\begin{IEEEbiography}[{\includegraphics[width=1in,height=1.25in,clip,keepaspectratio]{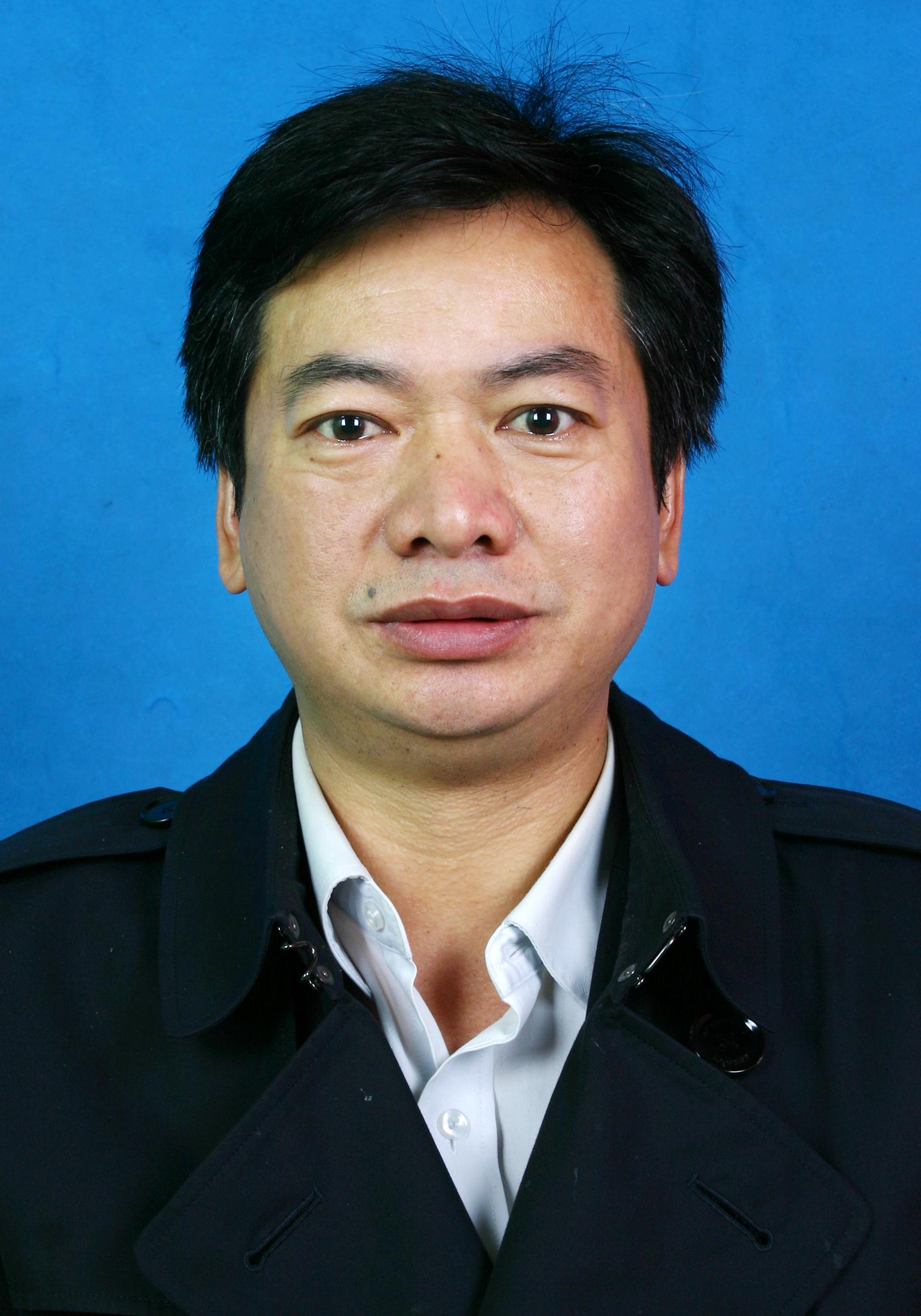}}]{Shichao Zhang}
is a China National Distinguished Professor with the Central South University, China. He holds a PhD degree from the Deakin University, Australia. His research interests include data mining and big data. He has published 90 international journal papers and over 70 international conference papers. He is a CI for 18 competitive national grants. He  serves/served as an associate editor for four journals.
\end{IEEEbiography}

\begin{IEEEbiography}[{\includegraphics[width=1in,height=1.25in,clip,keepaspectratio]{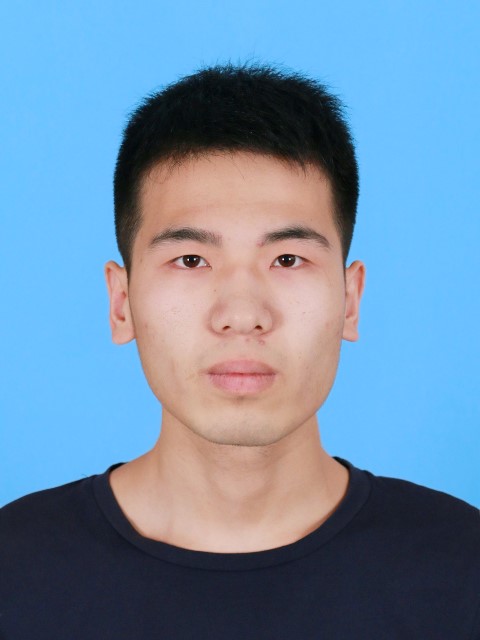}}]{Jiaye Li}
	is currently working toward the PhD degree at Central South University, China. His research interests include machine learning, data mining and deep learning.
\end{IEEEbiography}


\end{document}